\documentclass[10pt,twocolumn,letterpaper]{article}

\usepackage[pagenumbers]{cvpr} %

\usepackage{booktabs} %
\usepackage{diagbox}
\usepackage{multirow}
\usepackage{mathrsfs}
\usepackage{graphicx}
\usepackage{graphbox}
\usepackage{enumitem}
\usepackage{soul}
\usepackage{url}
\usepackage[normalem]{ulem}
\usepackage[skip=3pt]{caption}
\usepackage{subcaption}
\usepackage{rotating}
\usepackage{makecell}
\usepackage{tikz}

\newcommand{\mytilde}{\raise.17ex\hbox{$\scriptstyle\mathtt{\sim}$}}

\newcommand{\Epsilon}{\mathcal{E}}

\newcommand{\suppmat}[1]{{\color{black}#1}}

\definecolor{cvprblue}{rgb}{0.21,0.49,0.74}
\usepackage[pagebackref,breaklinks,colorlinks,allcolors=cvprblue]{hyperref}

\def\paperID{4816} %
\def\confName{CVPR}
\def\confYear{2026}

\newcommand{\methodName}{StableMaterials\xspace}

\title{\textbf{\methodName:} Enhancing Diversity in Material Generation via Semi-Supervised Learning}

\author{Giuseppe Vecchio\\
Adobe Research\\
{\tt\small gvecchio@adobe.com}
}

\let\oldtwocolumn\twocolumn
\renewcommand\twocolumn[1][]{%
    \oldtwocolumn[{#1}{
    \begin{center}
            \captionsetup{type=figure}
            \includegraphics[width=\textwidth]{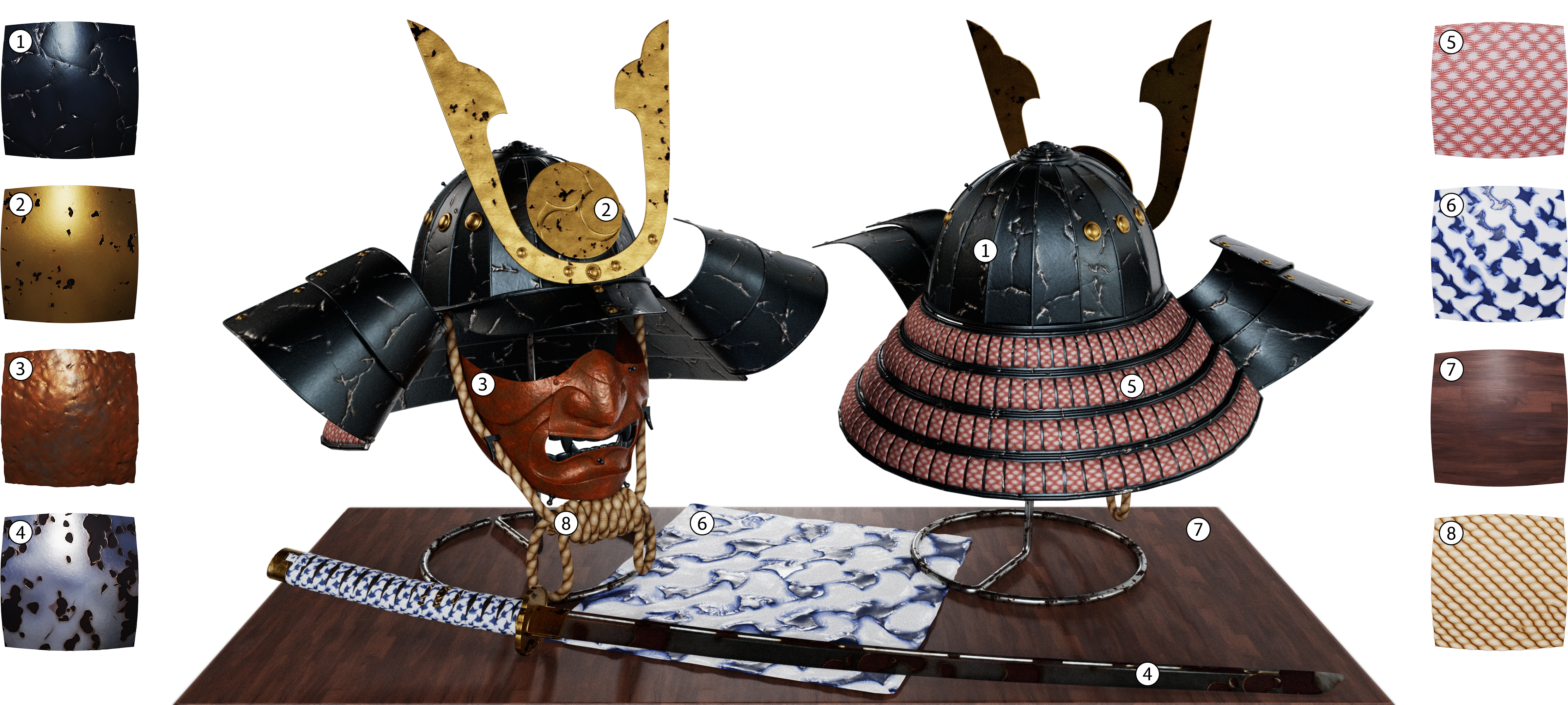}
            \captionof{figure}{We present \methodName, a diffusion--based model for materials generation through text or image prompting. Our approach enables high-resolution, tileable material maps, inferring both diffuse (Basecolor) and specular (Roughness, Metallic) properties, as well as the material mesostructure (Height, Normal).}
            \label{fig:thumbnail}
        \end{center}
    }]
}

\begin{document}
\maketitle
\begin{abstract}
We introduce \textbf{\methodName}, a novel approach for generating photorealistic physical-based rendering (PBR) materials that integrate semi-supervised learning with Latent Diffusion Models (LDMs). 
Our method employs adversarial training to distill knowledge from existing large-scale image generation models, minimizing the reliance on annotated data and enhancing the diversity in generation. This distillation approach aligns the distribution of the generated materials with that of image textures from an SDXL model, enabling the generation of novel materials that are not present in the initial training dataset.
Furthermore, we employ a diffusion-based refiner model to improve the visual quality of the samples and achieve high-resolution generation. Finally, we distill a latent consistency model for fast generation in just four steps and propose a new tileability technique that removes visual artifacts typically associated with fewer diffusion steps. 

We detail the architecture and training process of \methodName, the integration of semi-supervised training within existing LDM frameworks. %
Comparative evaluations with state-of-the-art methods show the effectiveness of \methodName, highlighting its potential applications in computer graphics and beyond.
\methodName is publicly available at \small{\url{https://gvecchio.com/stablematerials}}.
\end{abstract}
    
\section{Introduction}
\label{sec:introduction}

Authoring of materials has been a long-standing challenge in computer graphics, requiring very specialized skills and a high level of expertise. To simplify the creation of materials for 3D applications, such as videogames, architecture design, simulation, media, and more, recent methods have tried to leverage learning-based approaches to capture materials from input images~\cite{deschaintre2018single,deschaintre2019flexible,li2017modeling,li2018materials,martin2022materia,guo2021highlight,zhou2021adversarial,vecchio2021surfacenet,bi2020deep,gao2019deep,vecchio2024controlmat}, or generation from a set of conditions~\cite{guehl2020semi,guo2020materialgan,hu2022controlling,zhou2022tilegen,guo2023text2mat,vecchio2024controlmat,vecchio2023matfuse}. 
While these approaches have reduced technical barriers to material creation, their effectiveness depends on the quality and diversity of training data, which can limit their use in real-world applications.

Despite recent efforts to create large-scale materials datasets, such as~\citet{deschaintre2018single}, OpenSVBRDF~\cite{ma2023opensvbrdf}, and MatSynth~\cite{vecchio2023matsynth}, these datasets are limited in diversity~\cite{zhou2023PhotoMat}, not capturing the vast range observed in large-scale image datasets such as LAION~\cite{schuhmann2022laion}. 
These limitations can constrain the capabilities of learning-based approaches, potentially creating gaps in their generative capabilities and affecting realism and diversity.

Fine-tuning has become a common practice to reduce these gaps in training data by leveraging existing knowledge from large-scale pretrained models.
Techniques like Low-Rank Adaptation (LoRA)~\cite{hu2021lora} effectively fine-tune models while preventing catastrophic forgetting. Methods such as Diff-Instruct~\cite{luo2024diff}, on the other hand, employ distillation strategies to transfer knowledge from pretrained models. However, while fine-tuning or distillation within the same domain are straightforward, they pose significant challenges across different domains (e.g., image to material).

To overcome these limitations, we introduce \textbf{\methodName}, an approach that takes advantage of semi-supervised adversarial training to: (1) include unannotated (non-PBR) samples in training, and (2) distill knowledge from a large-scale pretrained SDXL~\cite{podell2023sdxl} model. In particular, we use a pretrained SDXL to generate unannotated texture samples from text prompts.
However, since \methodName is trained to produce SVBRDF maps, we cannot perform direct supervision using the generated textures. To include these textures in the training of our methods, we learn a common latent representation between textures and materials; then, we complement the traditional supervised loss, with an unsupervised adversarial loss, forcing the model to also generate realistic maps for unannotated samples and close the gap between the two data distributions.
In addition, drawing inspiration from the SDXL~\cite{podell2023sdxl} architecture, we use a diffusion-based refinement model to enhance the visual quality of the samples and achieve high-resolution generation. We initially generate materials at the model base resolution of 512x512, and subsequently apply our refiner using SDEdit~\cite{meng2021sdedit} and patched diffusion. This approach allows to achieve high resolution while constraining the patched generation, ensuring consistency and memory efficiency.
Subsequently, we distill a latent consistency model~\cite{song2023consistency} that allows fast generation by reducing the number of inference steps to four steps per stage. However, this comes at the cost of introducing visible seams when using approaches such as \textit{noise rolling}~\cite{vecchio2024controlmat} to achieve tileability. To solve this issue, we propose a novel \textit{features rolling} technique, which shifts the tensor rolling from the diffusion step to the U-Net architecture by directly shifting the feature maps within each convolutional and attention layer.

We evaluate our method qualitatively and quantitalively, and compare it with previous work, demonstrating the benefit of the semi-supervised training.
In summary, we introduce \methodName a novel solution combining supervised and adversarial training to generate highly realistic material in scenarios where annotated data are scarce.
The contributions of this work are as follows: 
\begin{itemize}
    \item \methodName, a new diffusion--based model for PBR material generation, leveraging semi-supervised learning to incorporate unannotated data during training.
    \item A novel adversarial distillation technique to bridge the gap with large-scale models. %
    \item A novel ``features rolling'' approach to tileability, minimizing the artifacts produced by fewer diffusion steps.
    \item State-of-the-art performance in materials generation.
\end{itemize}

\section{Related Work}
\label{sec:related}

\begin{figure*}
    \centering
    \includegraphics[width=1.0\textwidth]{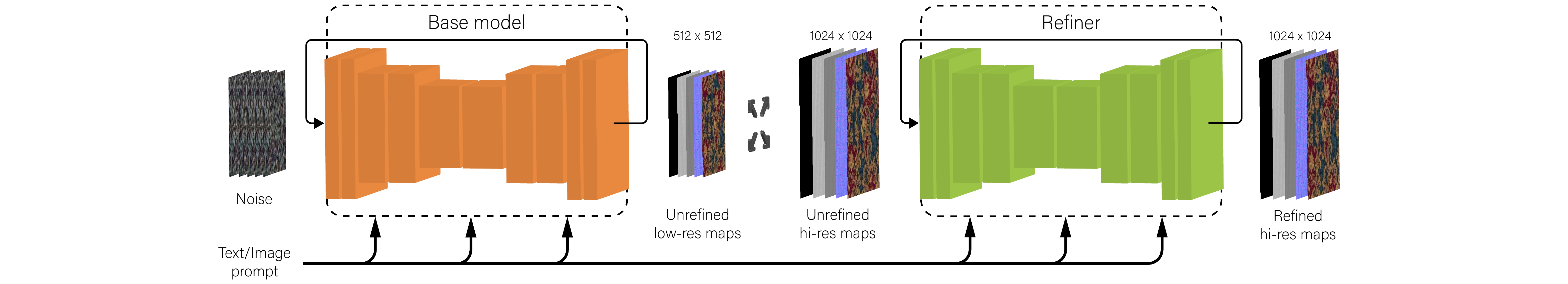}
    \caption{\textbf{Architecture of \methodName.} The \textit{base model} generates a low resolution materials of size 512x512. This generation is then upscaled and refined using SDEdit~\cite{meng2021sdedit} by the \textit{refiner model} using a patched approach to limit memory requirements.}
    \label{fig:architecture}
\end{figure*}

\paragraph{Materials Generation.}
Materials synthesis is an open challenge in computer graphics~\cite{guarnera2016brdf} with many recent data-driven approaches focusing on estimating SVBRDF maps from an image~\cite{deschaintre2018single,li2017modeling,li2018materials,martin2022materia,guo2021highlight,zhou2021adversarial,vecchio2021surfacenet,deschaintre2019flexible,bi2020deep,gao2019deep,vecchio2024controlmat}. 
Building on the success of generative models, several approaches to materials generation have emerged, including 
\citet{guehl2020semi} which combines procedural structure generation with data-driven color synthesis; MaterialGAN~\cite{guo2020materialgan}, a generative network which produces realistic SVBRDF parameter maps using the latent features learned from a StyleGAN2~\cite{stylegan2}; 
\citet{hu2022controlling} which generates new materials transferring the micro and meso-structure of a texture to a set of input material maps; and TileGen~\cite{zhou2022tilegen}, a generative model capable of producing tileable materials, conditioned by an input pattern but limited to class-specific training. 
Recent appraoches have focused on leveraging the generative capabilities of diffusion models for materials generation. In particular, \citet{vecchio2024controlmat} introduced ControlMat, which relies on the MatGen diffusion backbone, to capture materials from an input image. MatFuse~\cite{vecchio2023matfuse} extends generation control with multimodal conditioning, and enables editing of existing materials via `volumetric' inpainting. MaterialPalette~\cite{lopes2023material} extends the capture of materials to pictures of real-world scenes by finetuning a LoRA~\cite{hu2021lora} for each picture. Substance 3D Sampler~\cite{sampler} recently introduced a pipeline to generate materials by first synthesizing a texture via text conditioning.
However, these methods often lack diversity, struggling with complex material representation, or depend on image generation models, requiring additional steps to estimate the material parameters.
\methodName overcomes these limitations by including a wider variety of unannotated material samples via a semi-supervised training and improves on inference time by distilling a latent consistency model.

\paragraph{Generative models.} Image generation is a long-standing challenge in computer vision, due to the high dimensionality of images and complex data distributions. Generative Adversarial Networks (GAN)~\cite{goodfellow2014generative} enabled the creation of high-quality images~\cite{karras2017progressive,brock2018large,karras2020analyzing}, yet they suffer from unstable training~\cite{arjovsky2017wasserstein,gulrajani2017improved,mescheder2018convergence} and mode collapse behaviour.

Diffusion Models (DMs)~\cite{sohl2015deep,ho2020denoising}, particularly the more efficient Latent Diffusion (LDM) architecture~\cite{rombach2022high}, have emerged as an alternative to GANs, achieving state-of-the-art results in image generation tasks~\cite{dhariwal2021diffusion}, besides showing a more stable training.
Following the success of LDMs, research has focused on improving the generation quality~\cite{podell2023sdxl} and reducing the number of inference steps to speed up the generation process~\cite{song2023consistency,luo2023latent,sauer2023adversarial,sauer2024fast} through model distillation. Furthermore, due to the proliferation of large-scale pretrained models, several approaches were proposed to reuse their knowledge~\cite{hu2021lora,luo2024diff,ruiz2023dreambooth}. However, these approaches focus on fine-tuning within the same image domain, restricting their applicability to non-image domains.

\section{Method}
\label{sec:method}

\methodName builds on MatFuse~\cite{vecchio2023matfuse}, which adapts the LDM paradigm~\cite{rombach2022high} to synthesize high-quality, pixel-level reflectance properties for arbitrary materials (Fig.~\ref{fig:architecture}). We replace MatFuse's multi-encoder VAE architecture (used to learn a disentangled latent representation of material maps) with a more resource-efficient single-encoder model. %
We introduce a semi-supervised training strategy that distills knowledge from a large-scale SDXL model to increase generation diversity, and leverage latent consistency distillation~\cite{song2023consistency} and a novel \emph{feature rolling} technique for fast, tileable generation. A dedicated \emph{refiner model} enables high-resolution outputs while preserving global consistency.

\subsection{Material Representation}
\label{sec:matrep}
\methodName generates SVBRDF texture maps, representing a spatially varying Cook-Torrance microfacet model~\cite{cook1982reflectance,karis2013real}, using a GGX~\cite{walter2007microfacet} distribution function, as well as the mesostructure of the material.
Specifically, the generated maps include \emph{base color}, \emph{normal}, \emph{height}, \emph{roughness}, and \emph{metalness}, where roughness specifies the specular lobe width and metalness indicates conductor regions. 

\subsection{Material Generation}
\label{sec:model}
The generative model consists of a compression VAE~\cite{kingma2013auto} $\mathcal{E}$, encoding the material maps into a latent space, and a diffusion model~\cite{rombach2022high} $\epsilon_{\theta}$, modeling the distribution of these latent features.

\paragraph{Map Compression}
\label{sec:model_vae}

We first train a multi-encoder VAE with encoders $\Epsilon = {\mathcal{E}_1, \dots, \mathcal{E}_N}$ and a decoder $\mathcal{D}$. Each encoder $\mathcal{E}_i$ encodes the $i^{th}$ map $\textbf{M}_i$ into a latent vector $z_i$, and the concatenated tensor $z = concat(z_1, \dots, z_N)$ is decoded to reconstruct the maps $\hat{M} = \mathcal{D}(z)$. Following \cite{vecchio2023matfuse}, training uses pixel-space $L_2$ loss, perceptual LPIPS~\cite{zhang2021designing} loss, a patch-based adversarial loss~\cite{isola2017image,dosovitskiy2016generating}, and a rendering loss~\cite{deschaintre2018single}, with a Kullback-Leibler penalty~\cite{kingma2013auto,rezende2014stochastic} to regularize the latent space.

Afterward, we fine-tune a single-encoder model. We freeze the original decoder and train only the encoder $\Epsilon'$ to compress concatenated material maps into the same latent tensor $z$. This preserves the disentangled latent representation and maintains compression efficiency with fewer parameters, as shown in %
Sec.~\ref{sec:ablation_encoder}.

Lastly, we create a shared latent space for texture and materials. This allows us to have a common space that we can use to train our diffusion model. Specifically, we fine-tune an autoencoder that compresses a single texture (e.g., the base color) into $z$, again keeping the decoder frozen. This additional network is used for the semi-supervised training of the diffusion model, as described in Sec.~\ref{sec:training}.

\paragraph{Diffusion Model}
We train a diffusion model on the compressed latent representation $z$ of the material. This model, based on \citet{rombach2022high}, uses a time-conditional U-Net~\cite{ronneberger2015u} to denoise the latent vectors $z$. 
During training, we generate noised latent vectors using a deterministic forward diffusion process $q \left( z_t | z_{t-1} \right)$, as defined in~\citet{ho2020denoising}, transforming them into an isotropic Gaussian distribution. The diffusion network $\epsilon_{\theta}$ learns the backward diffusion $q(z_{t-1} | z_t)$ to denoise and reconstruct the original latent vector.
The model training is described in Sec.~\ref{sec:training}

\begin{figure*}
    \centering
    \includegraphics[width=1.0\linewidth]{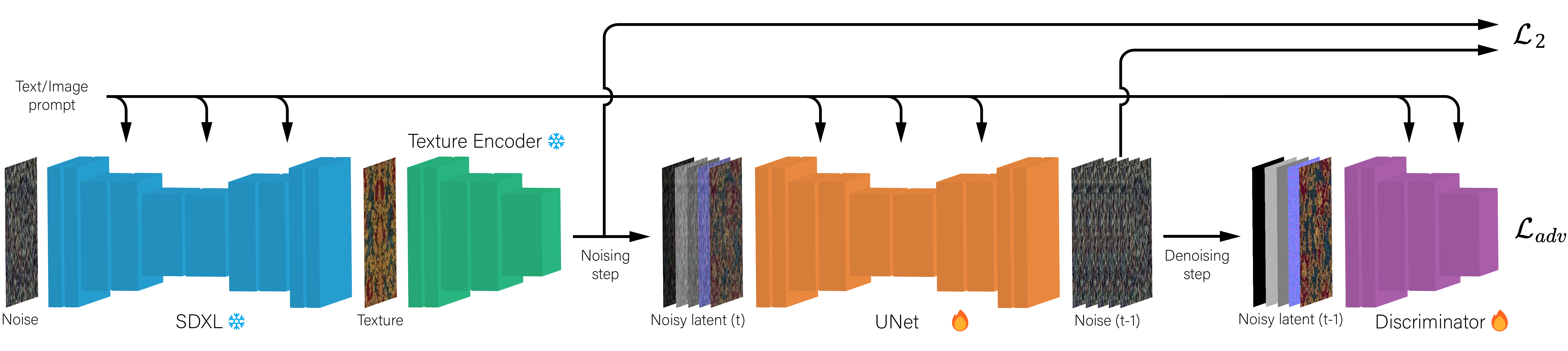}
    \caption{\textbf{Semi-Supervised training.} Both the SDXL model and \methodName are prompted to generate the same material. The supervised $\mathcal{L}_2$ loss, between the estimated noise and the added noise, is complemented by an adversarial loss $\mathcal{L}_{\text{adv}}$ computed on the denoised latent from \methodName.}
    \label{fig:method}
\end{figure*}

\paragraph{Conditioning}
\methodName allows flexible generation control via text or image prompts describing high-level appearance. We encode the text or image condition using a pretrained CLIP model~\cite{clip}, which outputs a single feature vector. To make training robust, we alternate between image and text prompts. Specifically, we use (i) an ambient-lit rendering of the material as an image condition and (ii) a text caption for text prompts; when no caption is available, we generate short descriptive tags as an alternative. In each training batch, one modality is randomly dropped—text with 75\% probability, image with 25\%—to balance the two conditions.

\subsection{Semi-Supervised Adversarial Distillation}
\label{sec:training}

To bridge the gap with image generation methods trained on large-scale datasets, we propose to distill knowledge from an SDXL model.
However, direct distillation is impractical due to domain differences between textures, represented from a single image, and materials, represented by multiple maps; therefore, we propose a semi-supervised approach to include unannotated samples (i.e.: textures without explicit material properties) during training.
Our method combines a \textit{supervised} loss on annotated materials with an \textit{unsupervised} adversarial loss that aligns the latent distribution of SDXL-generated textures with that of real materials. 

The \textbf{supervised Loss} ($\mathcal{L}{\text{sup}}$) ensures correct reconstruction of ground-truth material properties, maintaining physical plausibility.
The \textbf{adversarial Loss} ($\mathcal{L}{\text{adv}}$) guides the generator to map both materials and textures to a shared feature distribution, enabling diversity while maintaining realism. We introduce a latent discriminator ($LD$) that distinguishes between real material features and generated ones, effectively forcing the generator to produce material-like features even from unannotated textures.
This strategy avoids mode collapse being primarely supervised, with the adversarial loss working as a distillation strategy.

\paragraph{Supervised Loss} The supervised objective compares the denoiser’s prediction $\epsilon_\theta$ with the true noise $\epsilon_t$ introduced at each diffusion step $t$: 
\begin{align}
\mathcal{L}_{\text{sup}}
&= \mathbb{E}_{t, z_0, \epsilon}
   \left[
   \lVert \epsilon_t - \epsilon_\theta(z_{t,\text{mat}}, t) \rVert^2
   \right] \\
&\quad
+ \alpha\, \mathbb{E}_{t,z_0,\epsilon}
   \left[
   \lVert \epsilon_t - \epsilon_\theta(z_{t,\text{tex}}, t) \rVert^2
   \right].
\label{eq:loss_sup}
\end{align}
Here, $z_{t,\text{mat}}$ and $z_{t,\text{tex}}$ are noisy latents at step $t$ for materials and textures, respectively. The hyperparameter $\alpha$ (set to $0.15$) controls the relative importance of unannotated texture samples.

\paragraph{Adversarial Loss} In parallel, $\mathcal{L}{\text{adv}}$ encourages the model to treat textures and materials similarly, guiding the generator to produce material-like outputs even when starting from unannotated textures. We compute this loss on the denoised latents $z_{t-1}$: 
\begin{equation} 
    \mathcal{L}{\text{adv}}=-\mathbb{E}{\mathbf{z} \sim p(\mathbf{z})} \Bigl[LD\bigl(z_{t-1}\bigr)\Bigr], 
    \label{eq:loss_adv} 
\end{equation} 
where $z_{t-1} = \text{concat}\Bigl(z_{t-1,\text{mat}}, z_{t-1,\text{tex}}\Bigr)$ is the concatenation of denoised latents for materials and textures, and $LD(\cdot)$ is the output of our latent discriminator.
By using the discriminator to align the distributions of material and texture latents, this loss effectively distills knowledge from SDXL-generated textures while ensuring they conform to the features of real materials.

\paragraph{Latent Discriminator} We follow \citet{sauer2021projected, sauer2023stylegan} and train $LD$ with a hinge loss~\cite{lim2017geometric}, comparing \emph{real} latent embeddings $z_{t-1,\text{real}}$ (encoded from the VAE) against \emph{fake} latent embeddings $z_{t-1,\text{fake}}$ (denoised by the generator). The discriminator operates in a time-conditional fashion: it receives the same timestep $t$ and the CLIP embedding as the generator. Architecturally, $LD$ mirrors the U-Net encoder of the diffusion model and is initialized with the same weights, leveraging its understanding of the latent space to effectively guide the generator. Only ground-truth material latents are treated as real samples by the discriminator.

Unlike previous works that use adversarial distillation for fast generation~\cite{sauer2023adversarial, sauer2024fast}, our approach bridges the domain gap between materials and textures. By training primarily with supervised terms while using adversarial guidance, the generator learns plausible material features even from unannotated textures. The adversarial component ensures SDXL-generated textures map to realistic material latents, effectively countering any shading artifacts and enriching the diversity of generated materials.

\subsection{Fast High-Resolution Generation}
\label{sec:high}

\paragraph{Few steps generation}
To improve generation speed, we fine-tune a Latent Consistency Model (LCM)~\cite{luo2023latent}. LCM performs a one-stage guided distillation of an augmented Probability Flow ODE (PF-ODE) and directly predicts the solution at $t=0$ through a consistency function $f_{\theta}(z_t, c, t) \mapsto z_0$. Unlike two-stage methods~\cite{meng2023distillation}, LCM integrates the guidance scale $\omega$ directly into the PF-ODE and uses a skip-step strategy, ensuring consistency between time steps $t_{n+k}$ and $t_n$. This design avoids alignment issues present in typical two-stage approaches. As a result, LCM enables generation in only a few steps, resulting in faster material synthesis.

\paragraph{Features rolling}
Noise rolling~\cite{vecchio2024controlmat} has been used to achieve tileability through iterative diffusion but becomes less effective with fewer steps. 
We address this limitation proposing the \emph{features rolling}, which shifts rolling from the noisy inputs to the U-Net features. Within each convolution and attention layer, we randomly roll and then reverse the feature maps, thus preserving edge continuity while requiring fewer diffusion steps. For highly structured materials, we enable features rolling only after the first diffusion step to retain their global layout. \suppmat{We compare \textit{features rolling} with other tileability methods in Supplemental Materials.}

\paragraph{Latent Upscaling \& Refinement}
Combining features rolling with patch-based diffusion provides efficient, high-resolution synthesis, but patch-wise generation alone can introduce inconsistencies across tiles. We address this with a two-stage pipeline, similar to SDXL~\cite{podell2023sdxl}, using SDEdit~\cite{meng2021sdedit} for refinement. Specifically, we first generate at $512\times512$ resolution, then refine the output to the desired resolution with a strength of $0.5$, balancing new high-frequency details against global consistency. The refiner model is trained on full-resolution $512\times512$ crops from 4K materials (no downsampling), ensuring it captures fine surface details. We demonstrate the effectiveness of this approach in Sec.~\ref{sec:two_stage}, with additional samples provided in \suppmat{Supplemental Materials.}

\section{Implementation \& Results}
\label{sec:results}

In this section, we first introduce the datasets used in our work, we then assess the generation capabilities of \methodName and compare it with recent state-of-the-art methods.
We evaluate our design choices in the ablation studies.

\subsection{Dataset}
\label{sec:dataset}

We train our model on the combined MatSynth~\cite{vecchio2023matsynth} and~\citet{deschaintre2018single} datasets, for a total of $6,198$ unique PBR materials, using the original training/test splits. Our dataset includes 5 material maps (Basecolor, Normal, Height, Roughness, Metallic) and their renderings under different environmental illuminations. 
We complement the dataset with $4,000$ texture-text pairs from SDXL~\cite{podell2023sdxl}, using $200$ prompts. We query a ChatGPT~\cite{chatgpt} model to suggest relevant material prompts.
\suppmat{The full list of prompts and the SDXL textures, are provided in Supplemental Material.}

\subsection{Technical details}
\label{sec:procedure}

We train all models on a single NVIDIA RTX4090 GPU with 24GB of VRAM, employing gradient accumulation to achieve a larger batch.

\noindent\textbf{Autoencoder} is trained with mini-batch gradient descent, using the Adam optimizer~\cite{adam} and a batch size of $8$. We train the model for $1,000,000$ iterations with a learning rate of $4.5\cdot10^{-4}$ and enable the $\mathcal{L}_\text{adv}$ after $300,000$ iterations as in \citet{esser2021taming}.
We first train a multi-encoder VAE~\cite{vecchio2023matfuse}; then we fine-tune a single-encoder model to compress the concatenated maps into the same disentangled latent space, reducing the total of parameters from 271M to 101M, while keeping similar reconstruction performance (Tab.~\ref{tab:ablation_vae}). We fine-tune both the single encoder model and the texture encoder for $100,000$ steps while keeping the decoder frozen.

\begin{figure*}
    \centering
    \setlength{\tabcolsep}{.5pt}
    \begin{tabular}{cccccccccccccc}

        \small{Prompt} & \small{Basecolor} & \small{Normal} & \small{Height} & \small{Rough.} & \small{Metallic} & \small{Render} &
        \hspace{1mm}\small{Prompt} & \small{Basecolor} & \small{Normal} & \small{Height} & \small{Rough.} & \small{Metallic} & \small{Render} \\
        
        \vspace{-1mm}\includegraphics[width=0.069\linewidth]{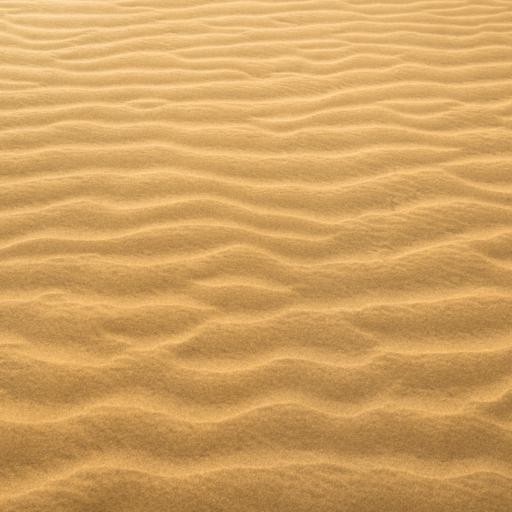} &
        \includegraphics[width=0.069\linewidth]{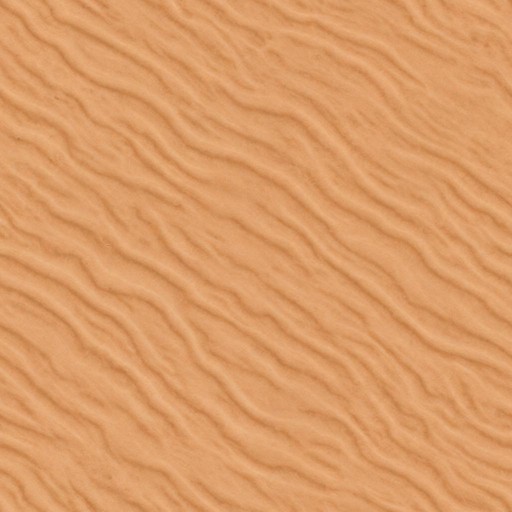} &
        \includegraphics[width=0.069\linewidth]{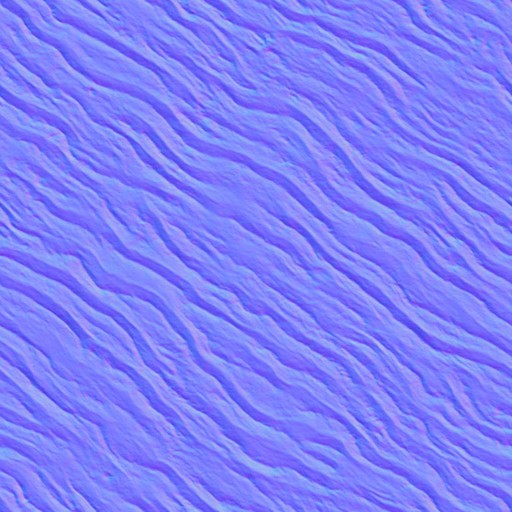} &
        \includegraphics[width=0.069\linewidth]{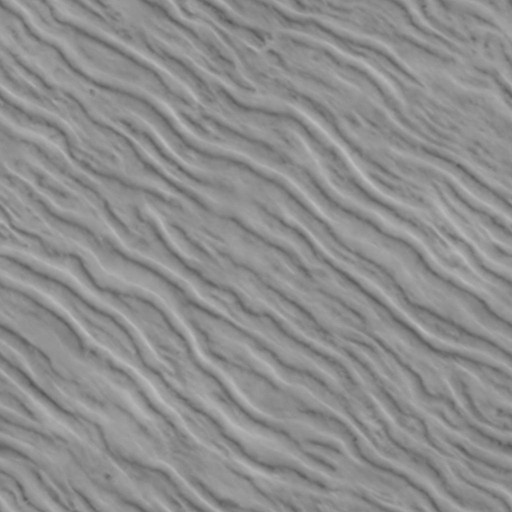} &
        \includegraphics[width=0.069\linewidth]{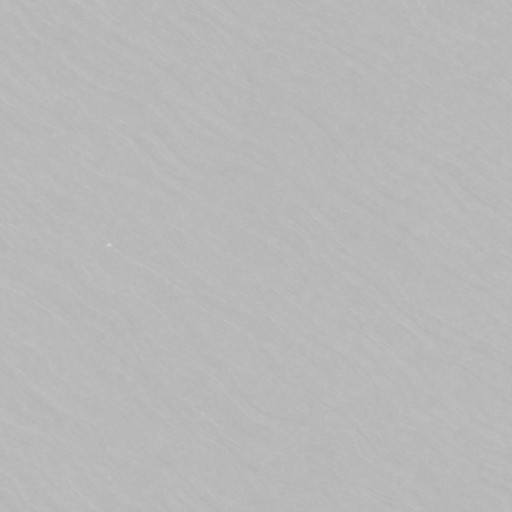} &
        \includegraphics[width=0.069\linewidth]{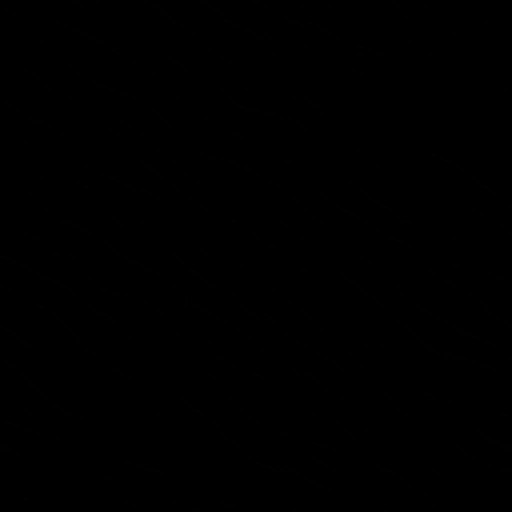} &
        \includegraphics[width=0.069\linewidth]{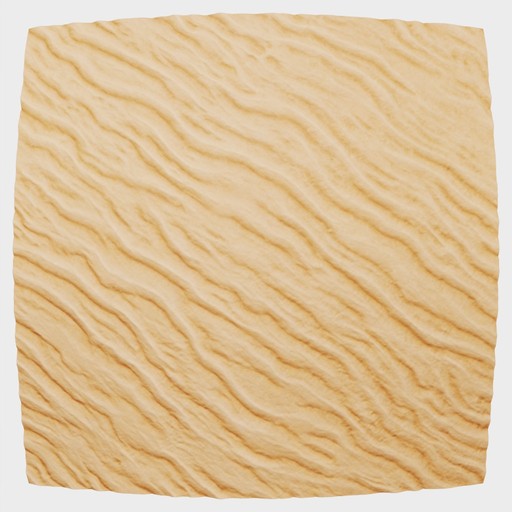} &
        \hspace{1mm}\includegraphics[width=0.069\linewidth]{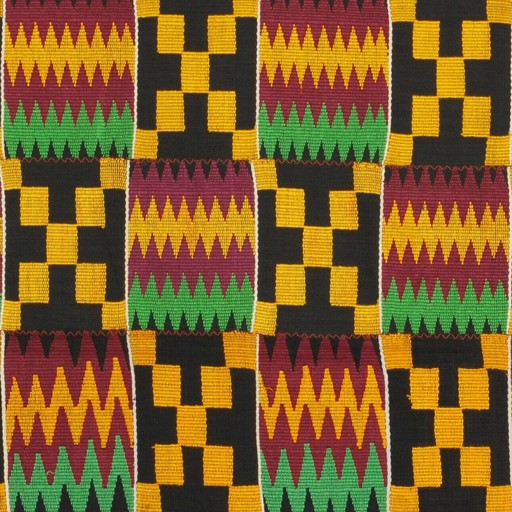} &
        \includegraphics[width=0.069\linewidth]{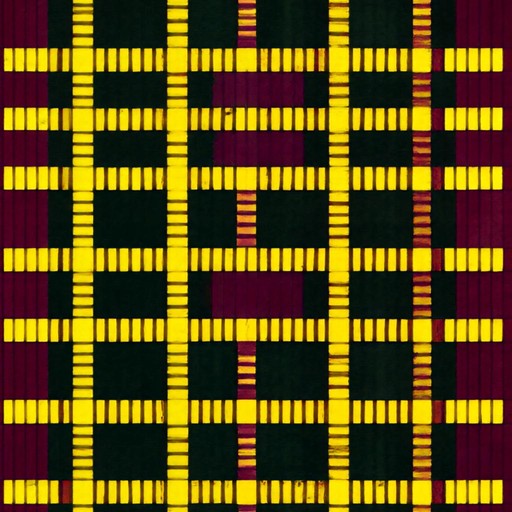} &
        \includegraphics[width=0.069\linewidth]{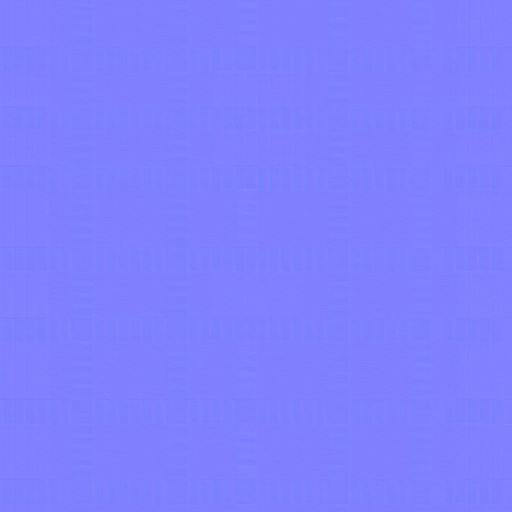} &
        \includegraphics[width=0.069\linewidth]{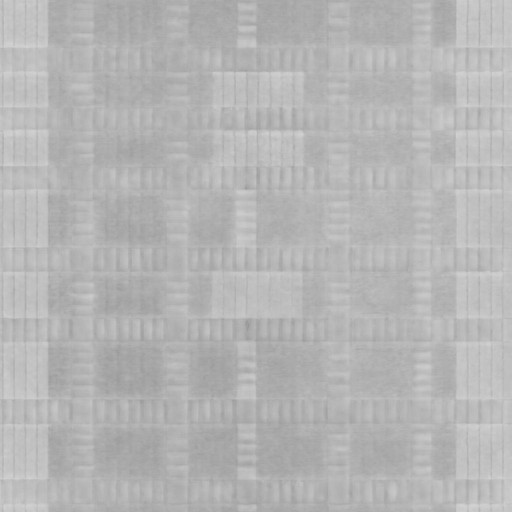} &
        \includegraphics[width=0.069\linewidth]{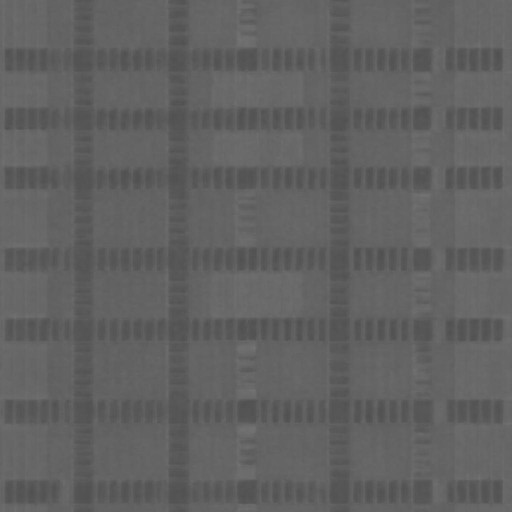} &
        \includegraphics[width=0.069\linewidth]{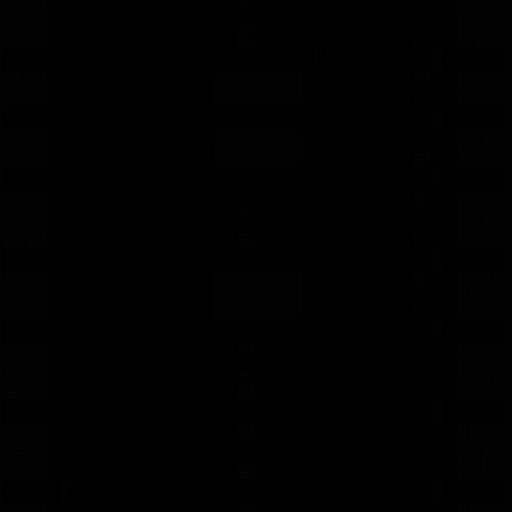} &
        \includegraphics[width=0.069\linewidth]{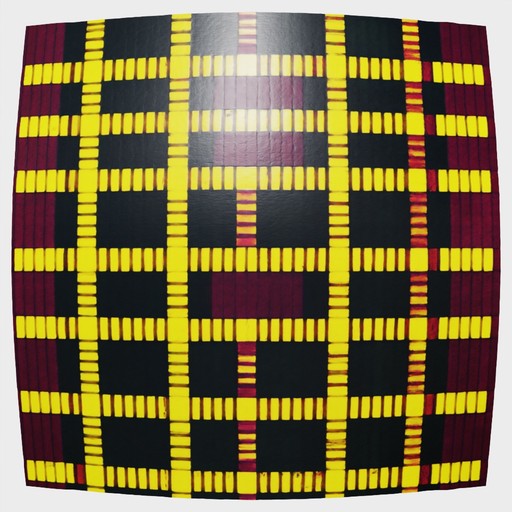} \\

        \vspace{-1mm}\includegraphics[width=0.069\linewidth]{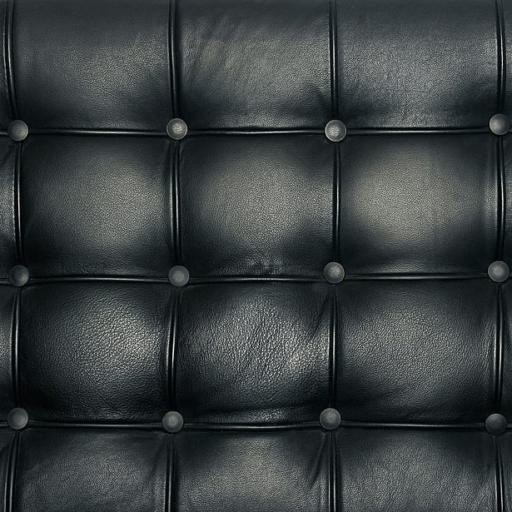} &
        \includegraphics[width=0.069\linewidth]{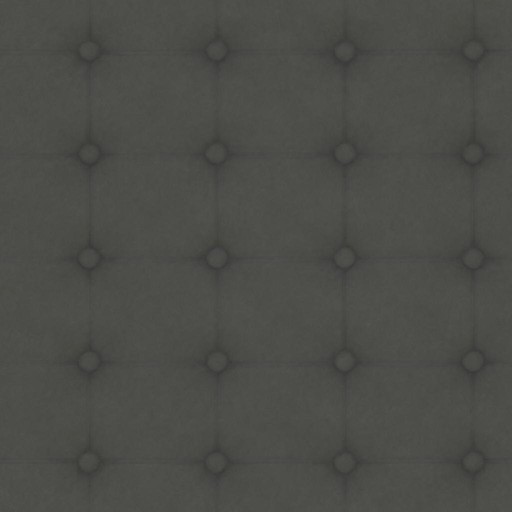} &
        \includegraphics[width=0.069\linewidth]{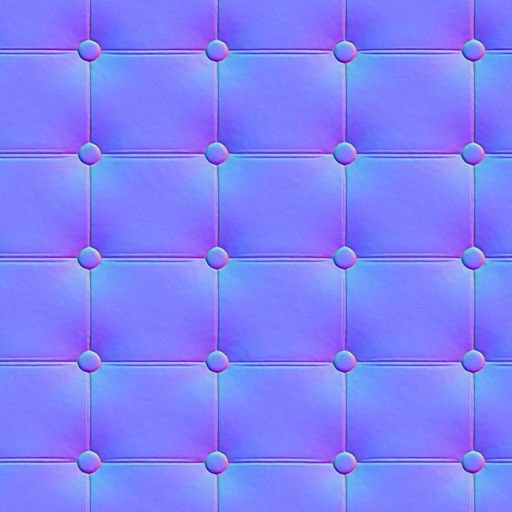} &
        \includegraphics[width=0.069\linewidth]{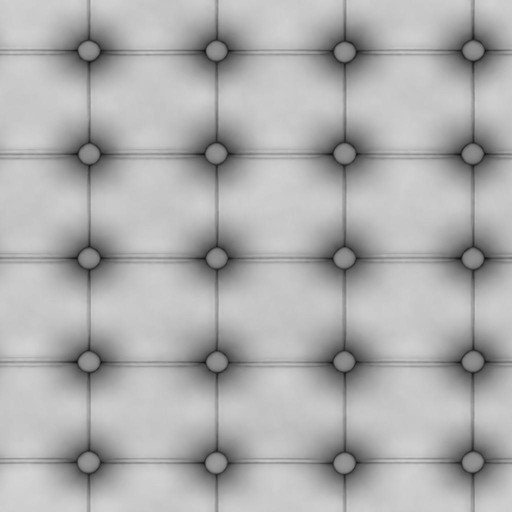} &
        \includegraphics[width=0.069\linewidth]{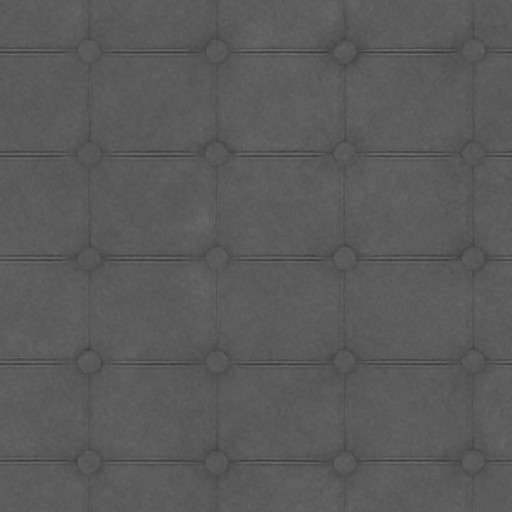} &
        \includegraphics[width=0.069\linewidth]{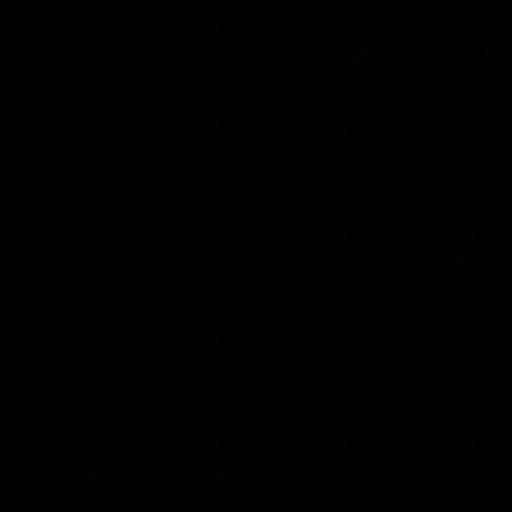} &
        \includegraphics[width=0.069\linewidth]{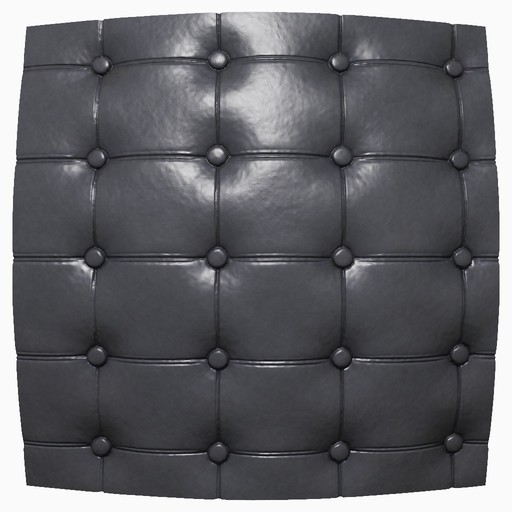} &
        \hspace{1mm}\includegraphics[width=0.069\linewidth]{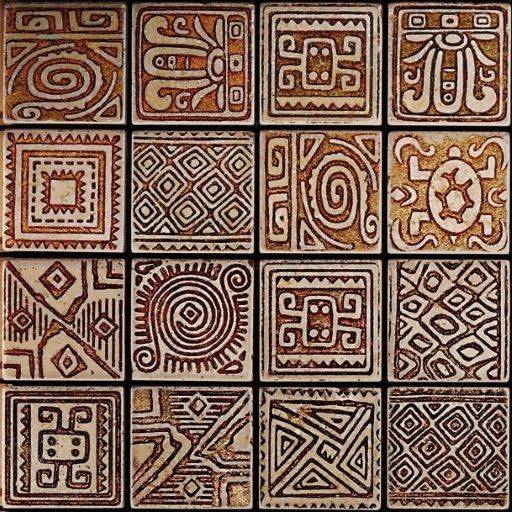} &
        \includegraphics[width=0.069\linewidth]{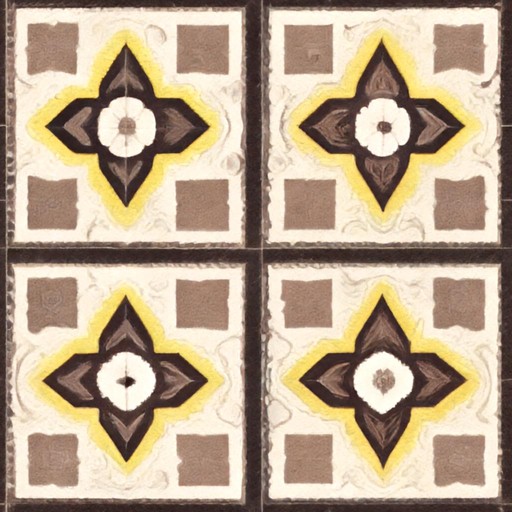} &
        \includegraphics[width=0.069\linewidth]{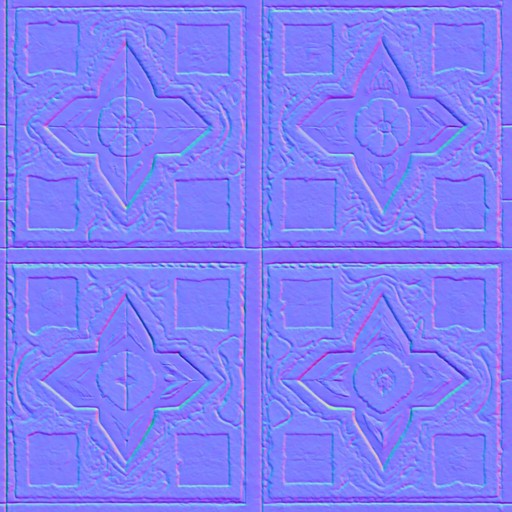} &
        \includegraphics[width=0.069\linewidth]{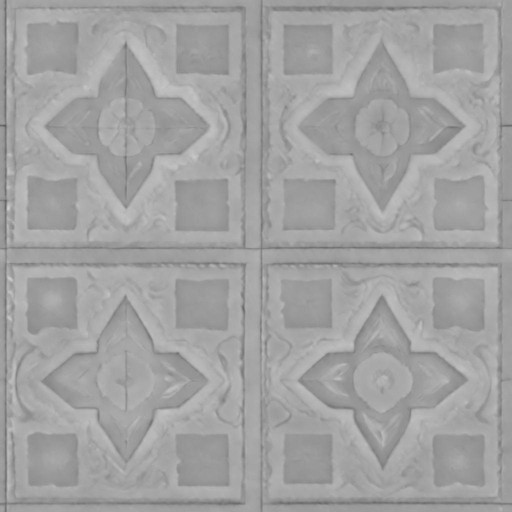} &
        \includegraphics[width=0.069\linewidth]{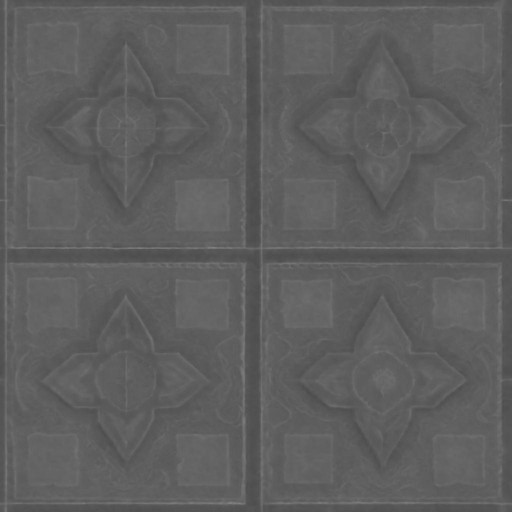} &
        \includegraphics[width=0.069\linewidth]{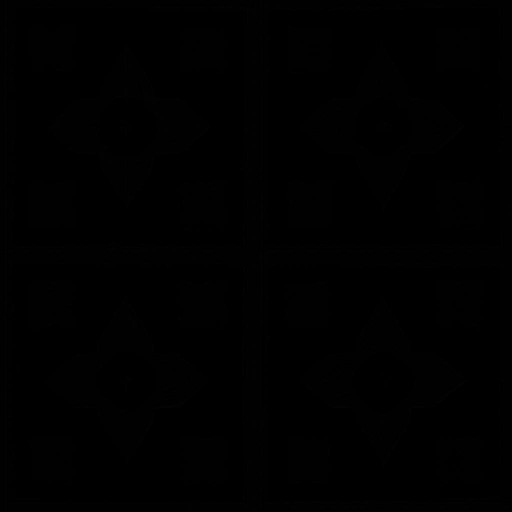} &
        \includegraphics[width=0.069\linewidth]{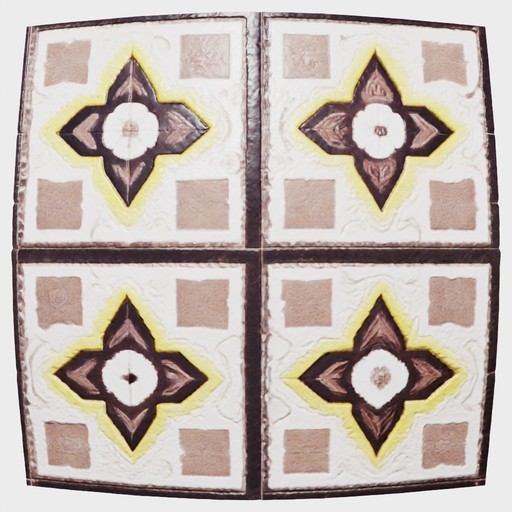} \\
    
    \end{tabular}
    
    \caption{\textbf{Image-prompting}. \methodName accurately captures the visual appearance of the input image, producing realistic materials for both in-domain (on the left) and out-domain (on the right) prompts. The render highlights the model's ability to handle diverse and complex surfaces.}
    \label{fig:generation_img}
\end{figure*}

\begin{figure*}
    \centering
    \setlength{\tabcolsep}{.5pt}
    \begin{tabular}{cccccccccccc}
        \small{Basecolor} & \small{Normal} & \small{Height} & \small{Rough.} & \small{Metallic} & \small{Render} &
        \hspace{1mm}\small{Basecolor} & \small{Normal} & \small{Height} & \small{Rough.} & \small{Metallic} & \small{Render} \\
        
        \vspace{-1mm}\includegraphics[width=0.081\linewidth]{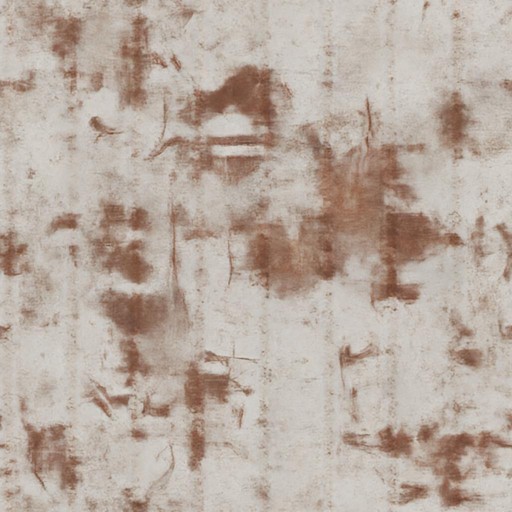} &
        \includegraphics[width=0.081\linewidth]{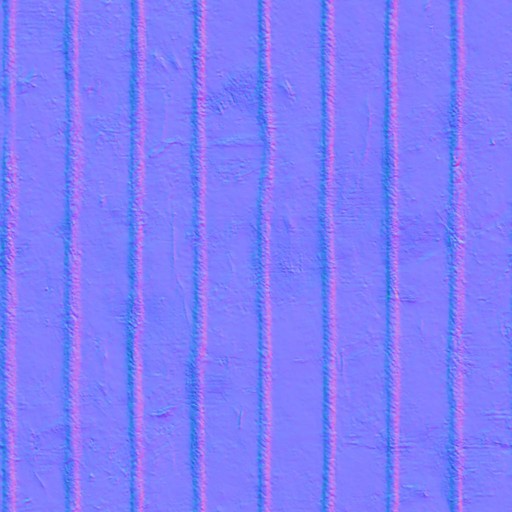} &
        \includegraphics[width=0.081\linewidth]{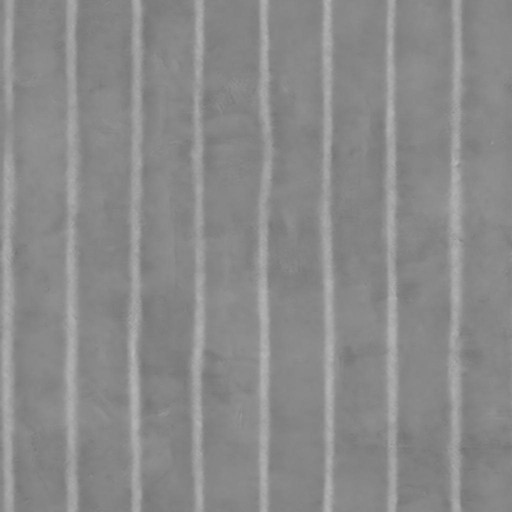} &
        \includegraphics[width=0.081\linewidth]{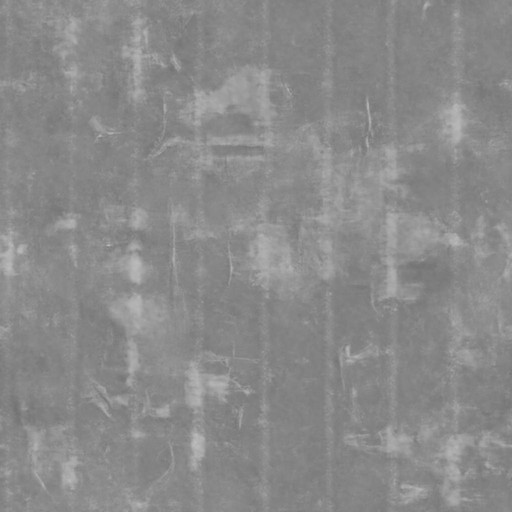} &
        \includegraphics[width=0.081\linewidth]{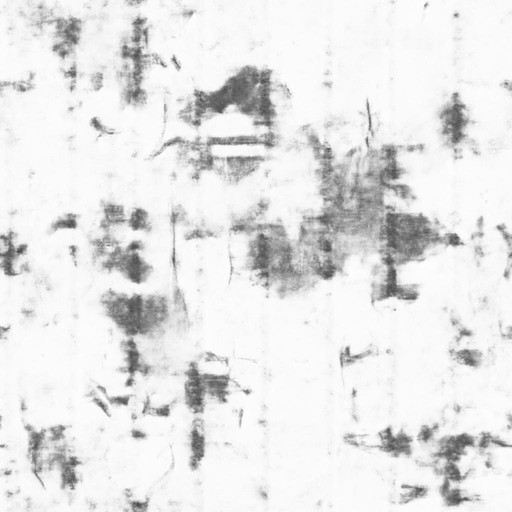} &
        \includegraphics[width=0.081\linewidth]{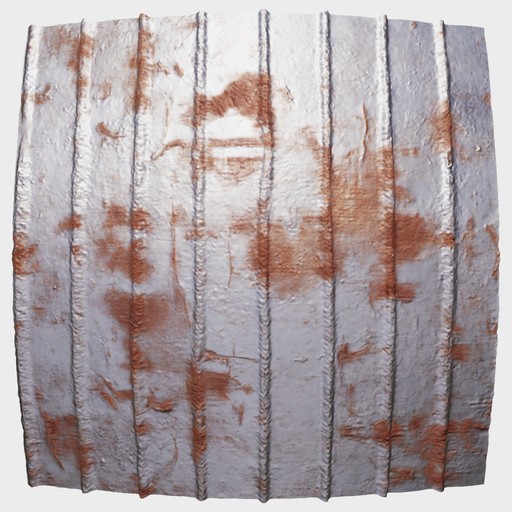} &
        \hspace{1mm}\includegraphics[width=0.081\linewidth]{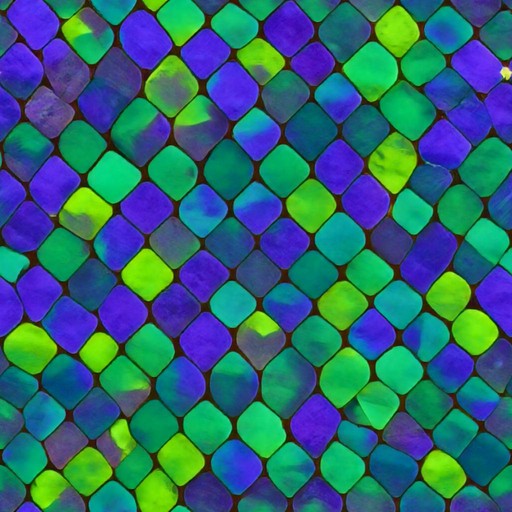} &
        \includegraphics[width=0.081\linewidth]{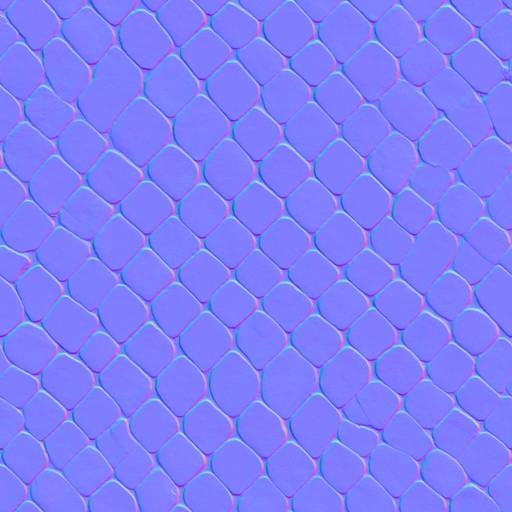} &
        \includegraphics[width=0.081\linewidth]{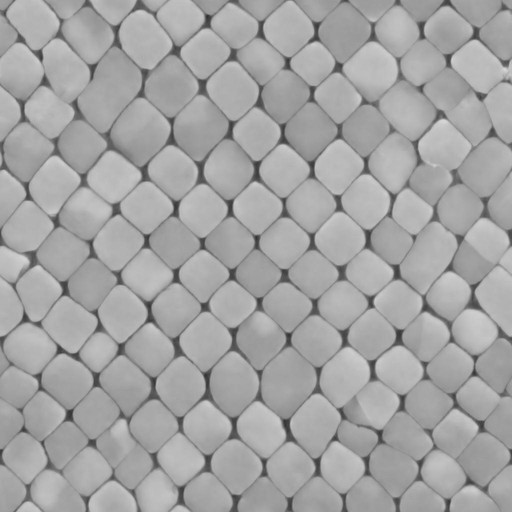} &
        \includegraphics[width=0.081\linewidth]{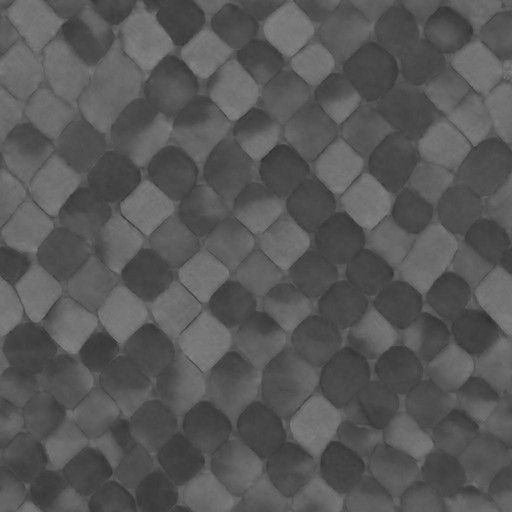} &
        \includegraphics[width=0.081\linewidth]{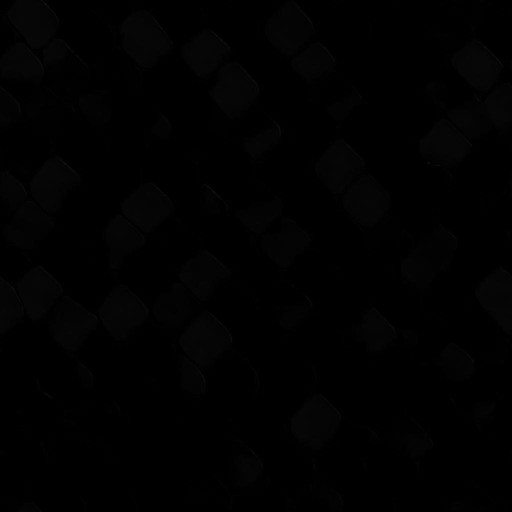} &
        \includegraphics[width=0.081\linewidth]{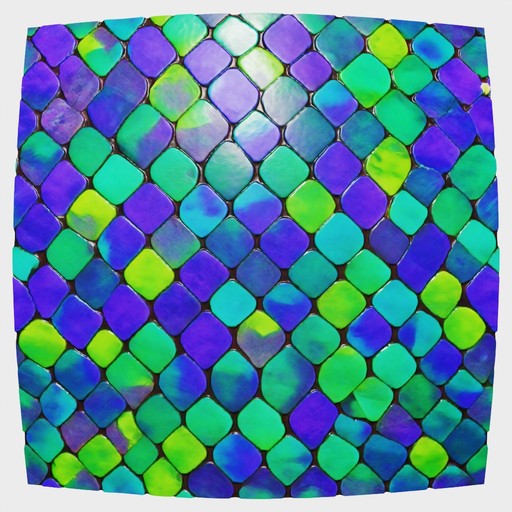} \\
        \multicolumn{6}{c}{\footnotesize{`Old rusty metal bars.'}} &
        \multicolumn{6}{c}{\footnotesize{`Fish scales reflecting a multitude of colors .'}} \\
    
        \vspace{-1mm}\includegraphics[width=0.081\linewidth]{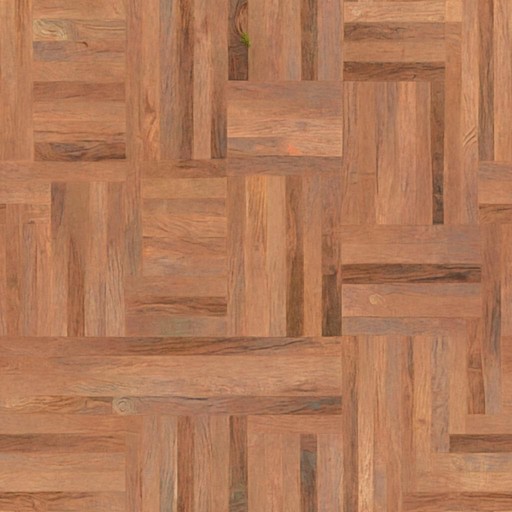} &
        \includegraphics[width=0.081\linewidth]{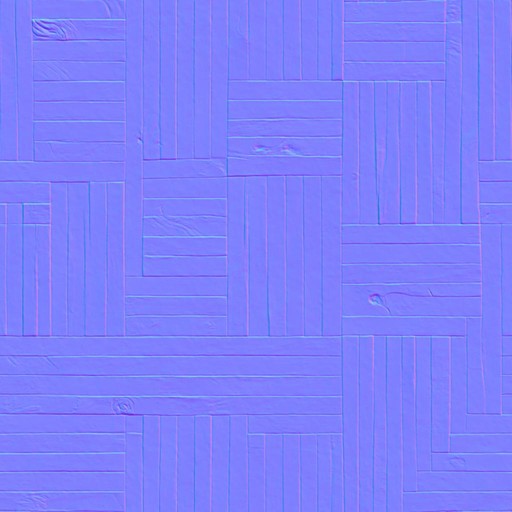} &
        \includegraphics[width=0.081\linewidth]{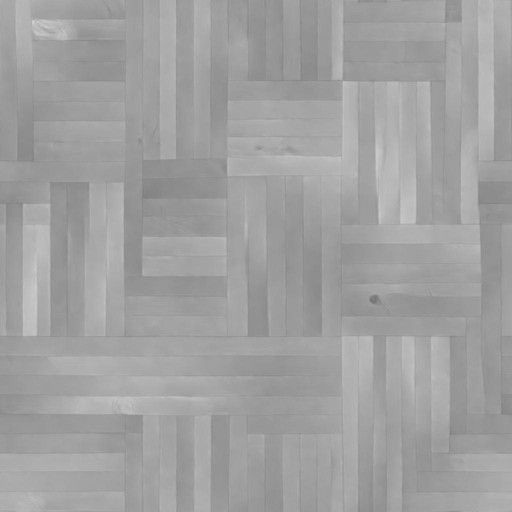} &
        \includegraphics[width=0.081\linewidth]{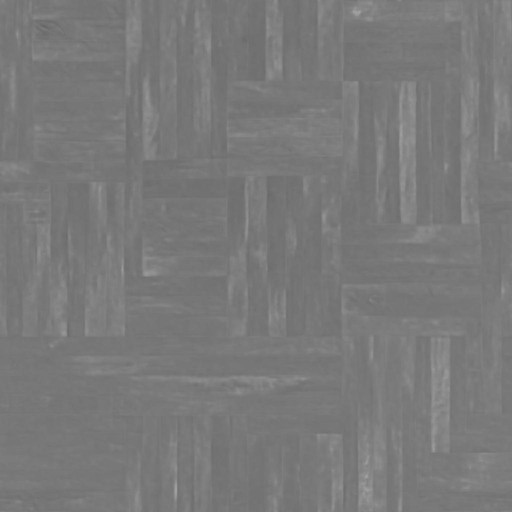} &
        \includegraphics[width=0.081\linewidth]{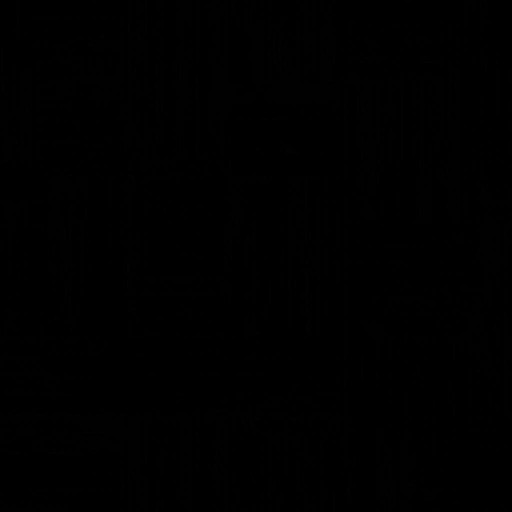} &
        \includegraphics[width=0.081\linewidth]{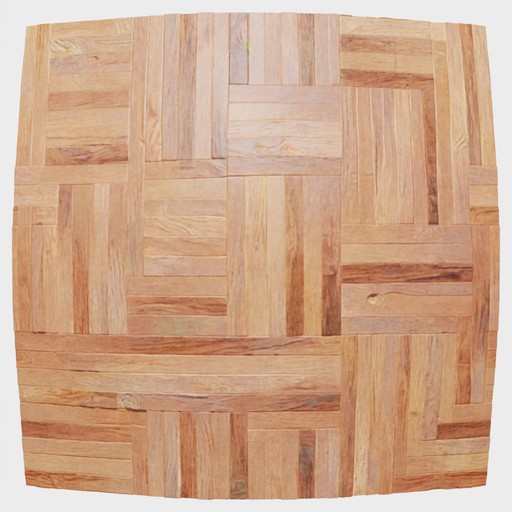} &
        \hspace{1mm}\includegraphics[width=0.081\linewidth]{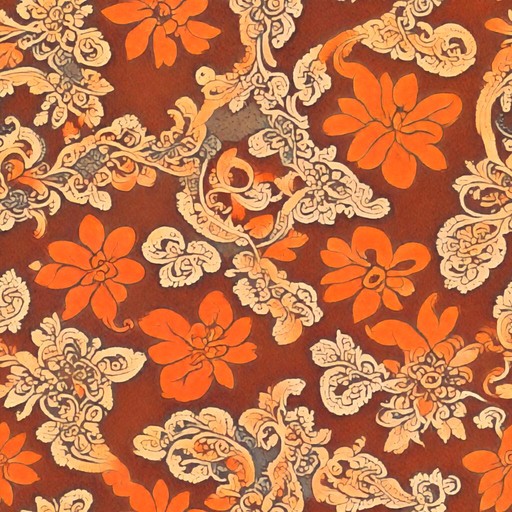} &
        \includegraphics[width=0.081\linewidth]{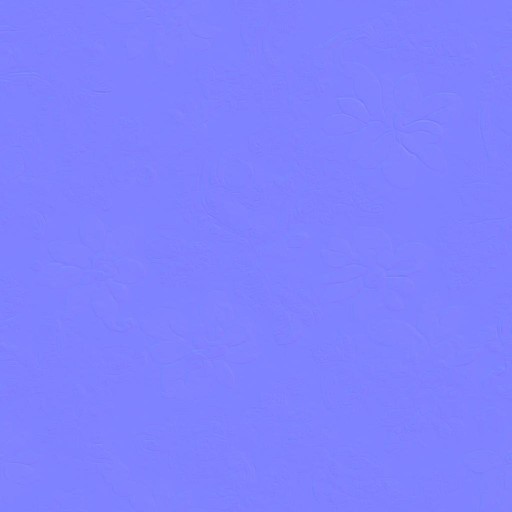} &
        \includegraphics[width=0.081\linewidth]{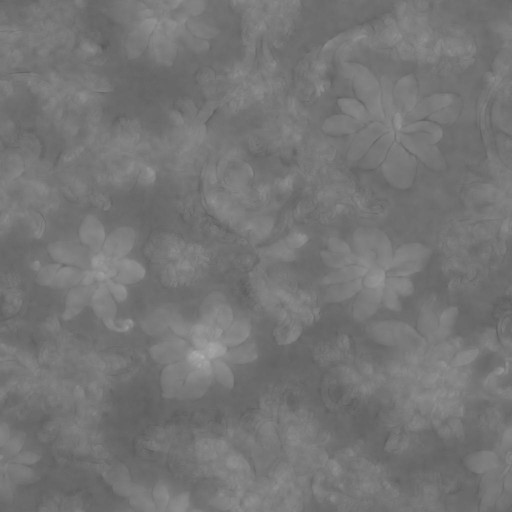} &
        \includegraphics[width=0.081\linewidth]{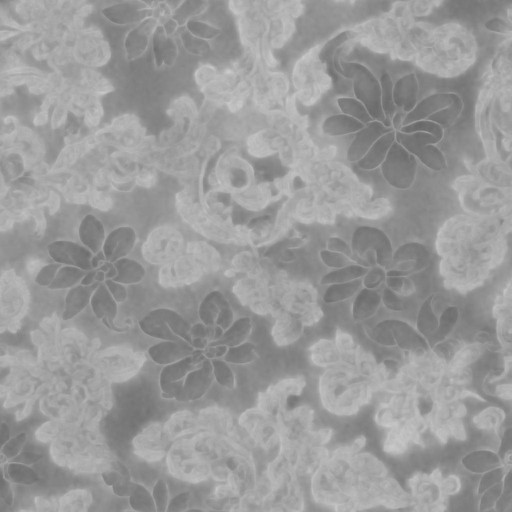} &
        \includegraphics[width=0.081\linewidth]{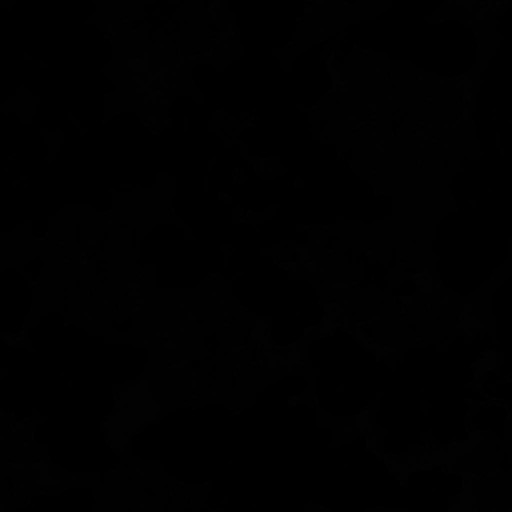} &
        \includegraphics[width=0.081\linewidth]{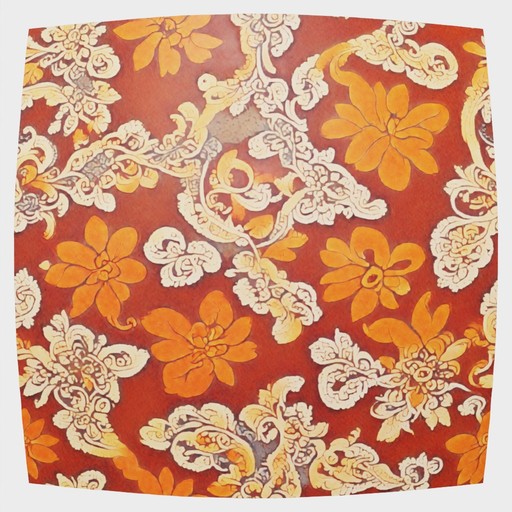} \\
        \multicolumn{6}{c}{\footnotesize{`Old wooden parquet floor.'}} &
        \multicolumn{6}{c}{\footnotesize{`Batik fabric with Indonesian patterns.'}} \\
    \end{tabular}
    
    \caption{\textbf{Text-prompting}. \methodName closely follow the input prompt, producing realistic materials for both in-domain (on the left) and out-domain (on the right) samples. The render highlights the model's ability to generate accurate properties for different types of materials.}
    \label{fig:generation_text}
\end{figure*}

\begin{figure*}
    \centering
    \setlength{\tabcolsep}{.5pt}
    \begin{tabular}{cccccccccccc}
        \small{Prompt} & \small{MatFuse} & \small{MatGen} & \small{\makecell{Material \\ Palette}} & \small{\makecell{\textbf{Stable} \\ \textbf{Materials}}} &
        \small{Prompt} & \small{MatFuse} & \small{MatGen} & \small{\makecell{Material \\ Palette}} & \small{\makecell{\textbf{Stable} \\ \textbf{Materials}}} \\
    
        \vspace{-1mm}\includegraphics[width=0.0975\linewidth]{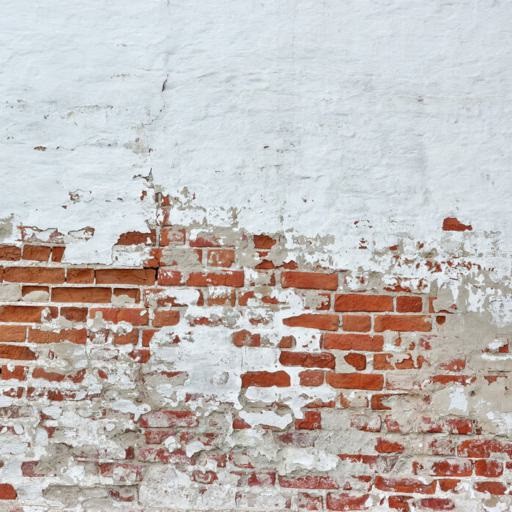} &
        \includegraphics[width=0.0975\linewidth]{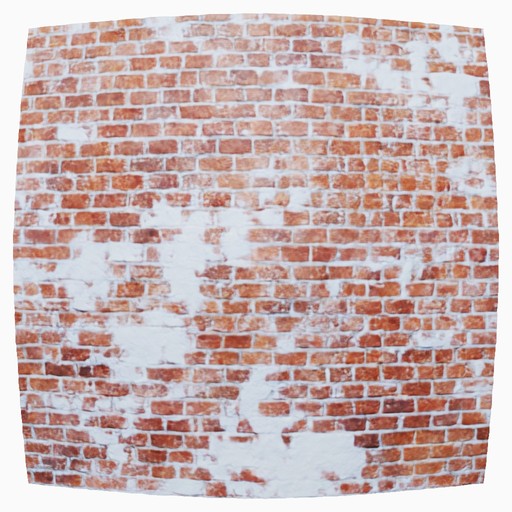} &
        \includegraphics[width=0.0975\linewidth]{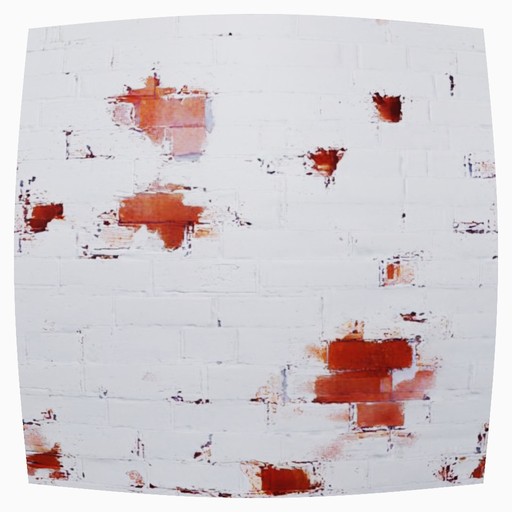} &
        \includegraphics[width=0.0975\linewidth]{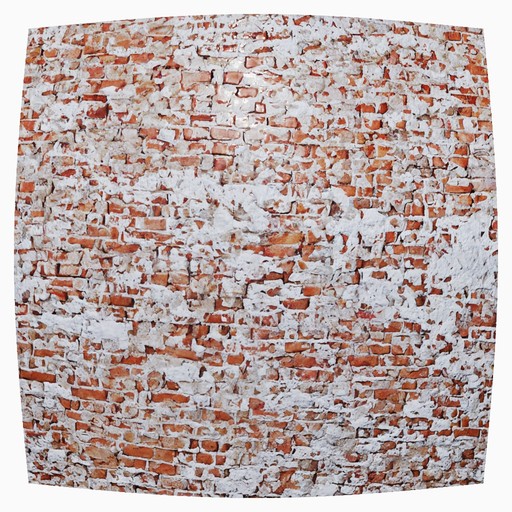} &
        \includegraphics[width=0.0975\linewidth]{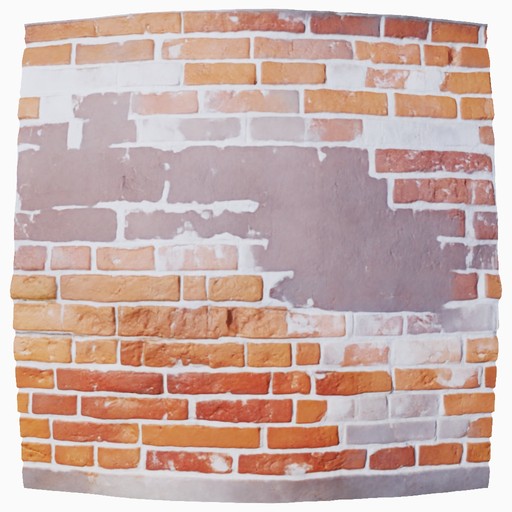} &
        \hspace{1mm}\includegraphics[width=0.0975\linewidth]{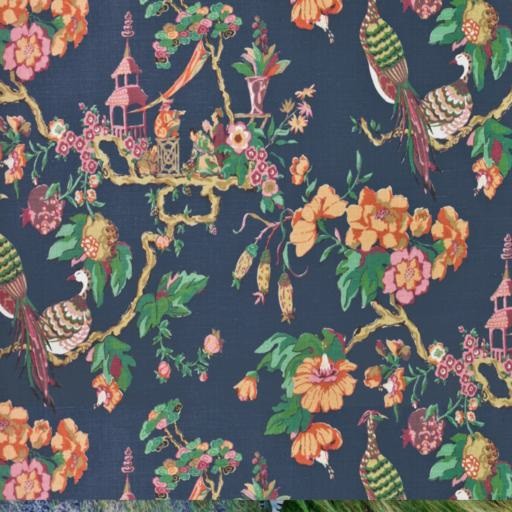} &
        \includegraphics[width=0.0975\linewidth]{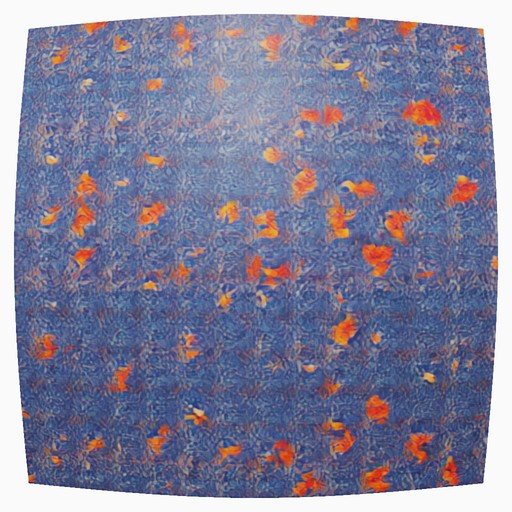} &
        \includegraphics[width=0.0975\linewidth]{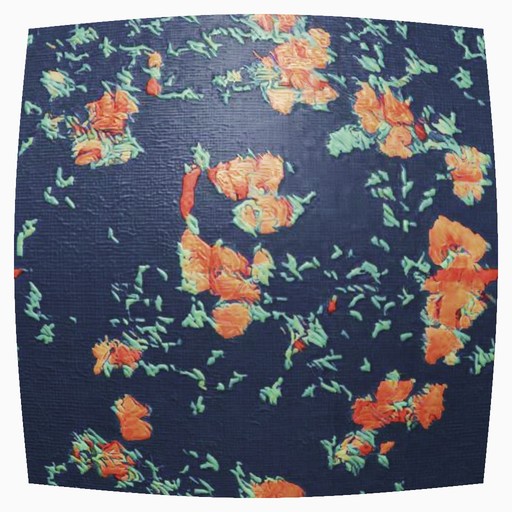} &
        \includegraphics[width=0.0975\linewidth]{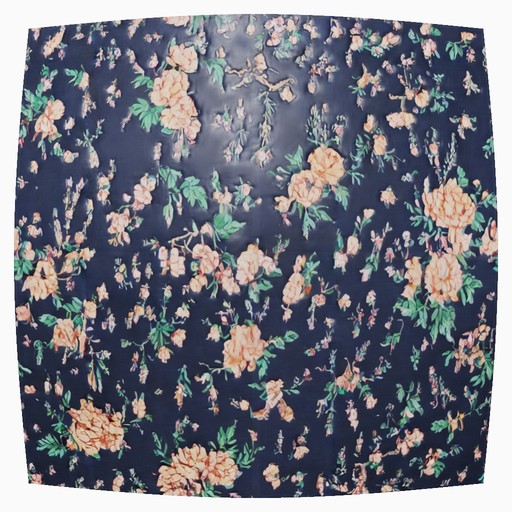} &
        \includegraphics[width=0.0975\linewidth]{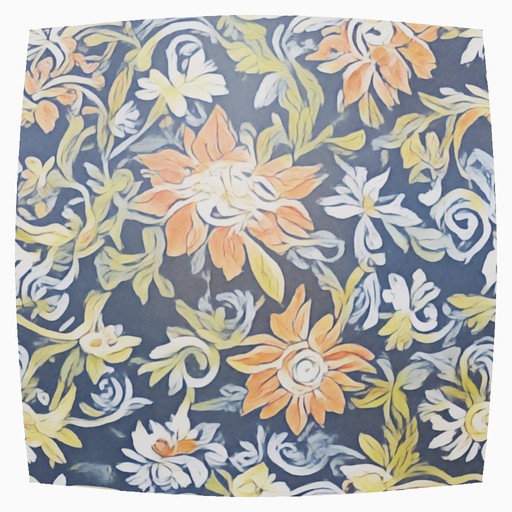} \\

        \vspace{-1mm}\includegraphics[width=0.0975\linewidth]{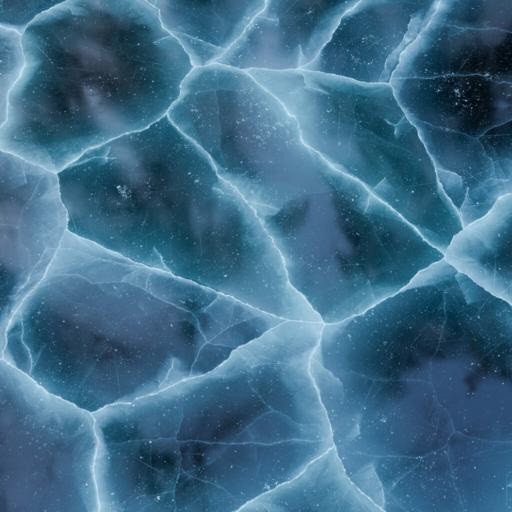} &
        \includegraphics[width=0.0975\linewidth]{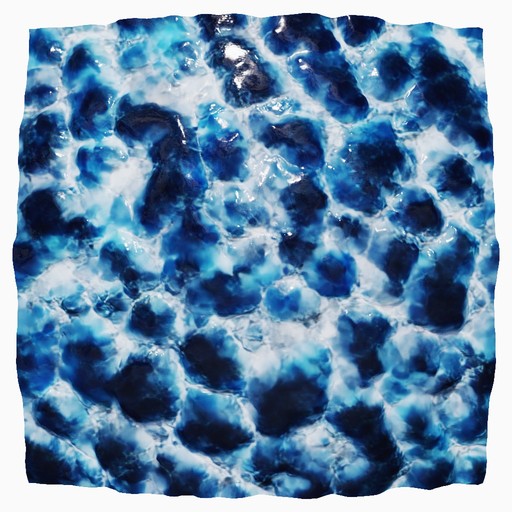} &
        \includegraphics[width=0.0975\linewidth]{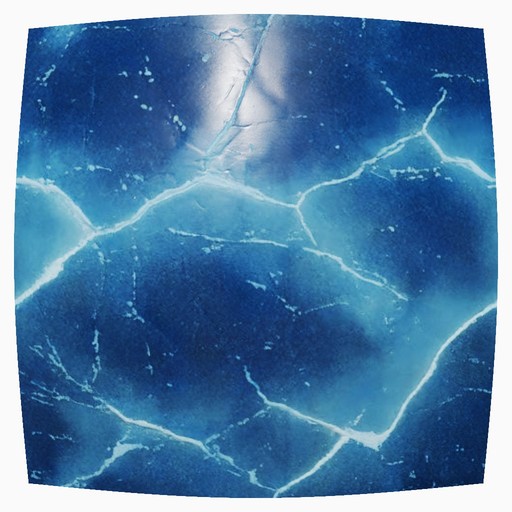} &
        \includegraphics[width=0.0975\linewidth]{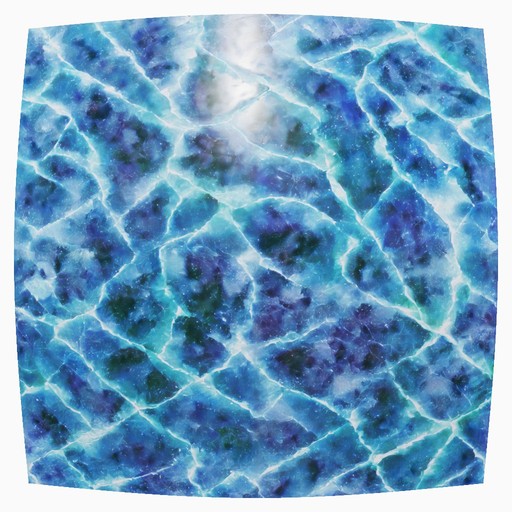} &
        \includegraphics[width=0.0975\linewidth]{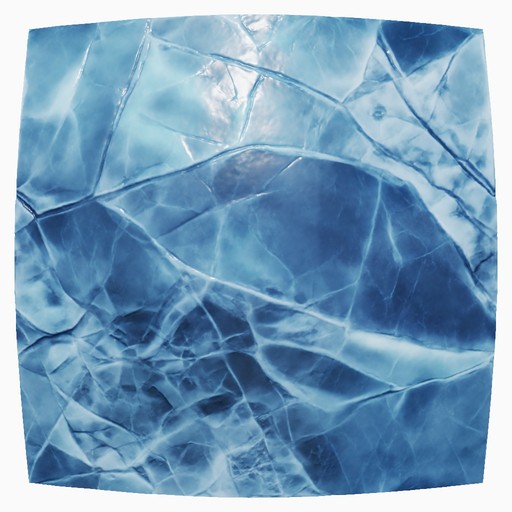} &
        \hspace{1mm}\includegraphics[width=0.0975\linewidth]{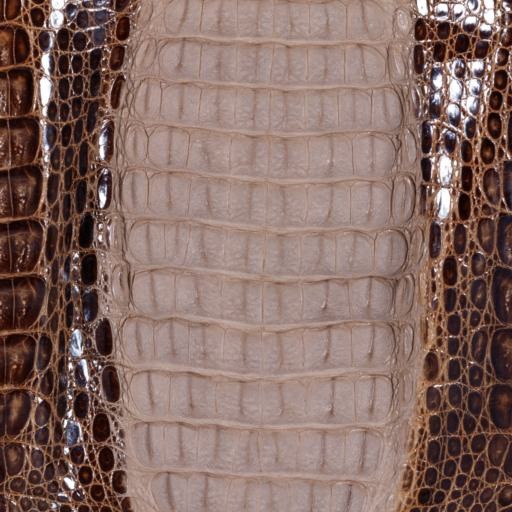} &
        \includegraphics[width=0.0975\linewidth]{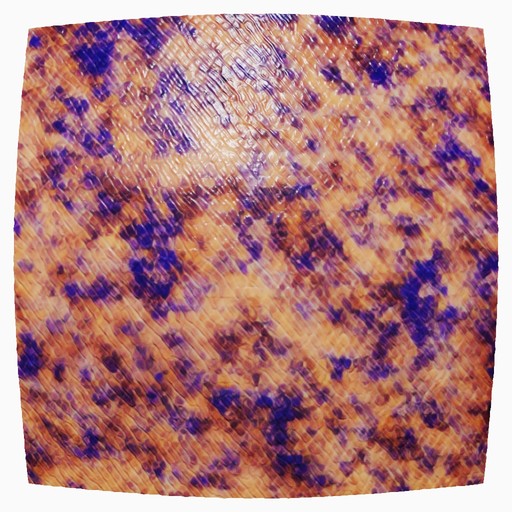} &
        \includegraphics[width=0.0975\linewidth]{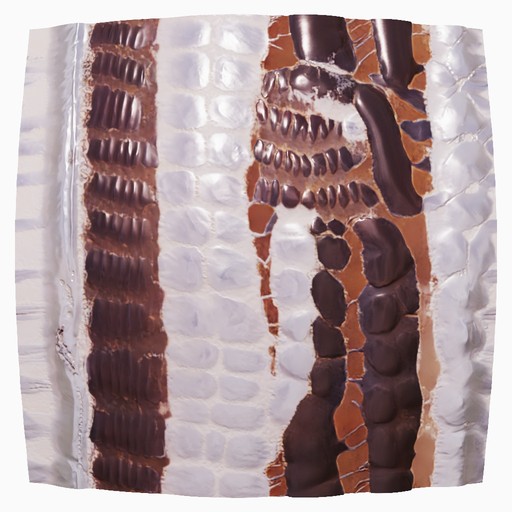} &
        \includegraphics[width=0.0975\linewidth]{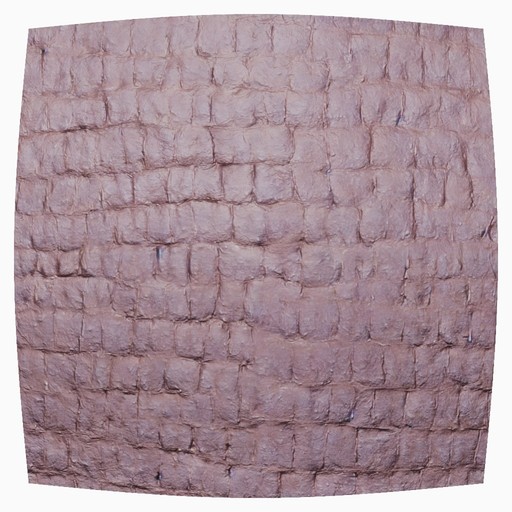} &
        \includegraphics[width=0.0975\linewidth]{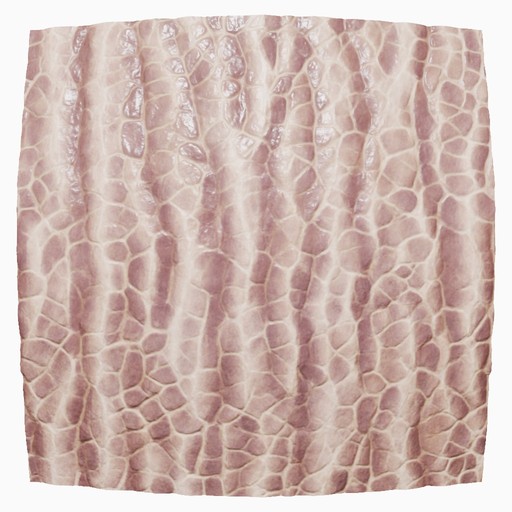} \\
        
    \end{tabular}
    
    \caption{\textbf{Comparison for image-prompting}. We compare \methodName with MatFuse, MatGen, and MaterialPalette in image-prompted generation, showing two in-domain (left column) and two out-domain (right column) renderings per model. \methodName improves over previous methods quality and ability to  captures the visual appearance of the input image.}
    \label{fig:comparison_img}
    \vspace{-1mm}
\end{figure*}

\begin{figure*}
    \centering
    \setlength{\tabcolsep}{.5pt}
    \begin{tabular}{cccccccc}
        \small{MatFuse} & \small{MatGen} & \small{\makecell{Substance \\ 3D Sampler}} & \small{\makecell{\textbf{Stable} \\ \textbf{Materials}}} &
        \hspace{0.25mm}\small{MatFuse} & \small{MatGen}  & \small{\makecell{Substance \\ 3D Sampler}} & \small{\makecell{\textbf{Stable} \\ \textbf{Materials}}} \\
        
        \vspace{-1mm} \includegraphics[width=0.1225\linewidth]{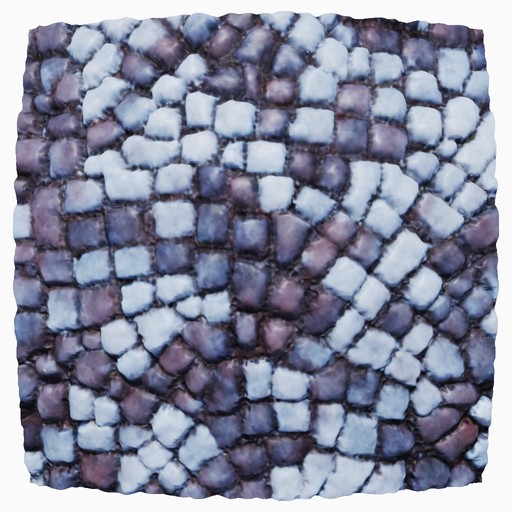} &
        \includegraphics[width=0.1225\linewidth]{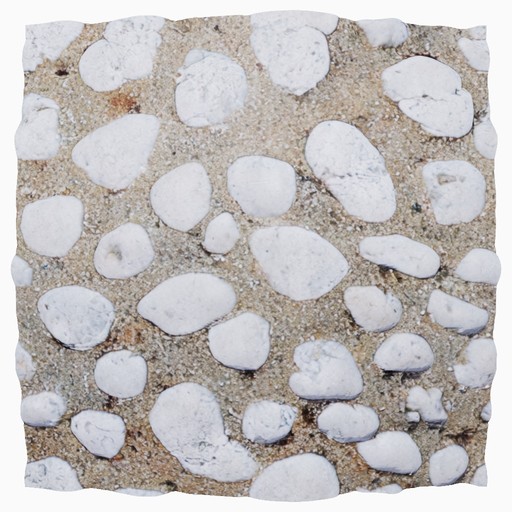} &
        \includegraphics[width=0.1225\linewidth]{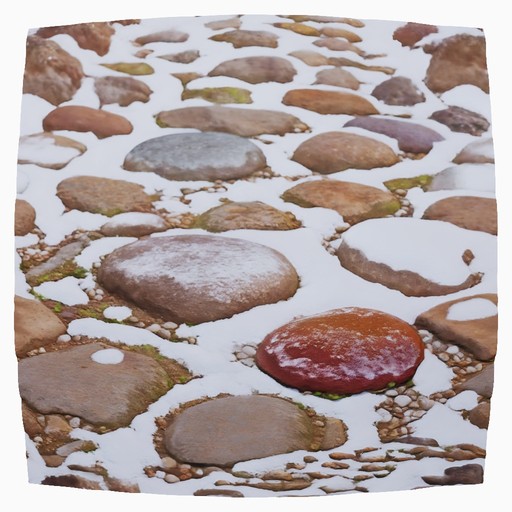} &
        \includegraphics[width=0.1225\linewidth]{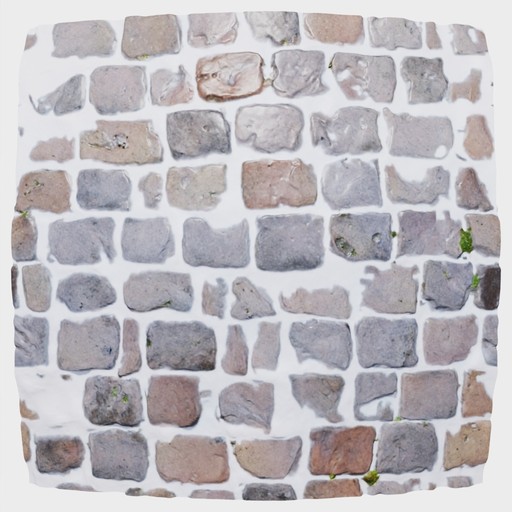} &
        \hspace{0.25mm} \includegraphics[width=0.1225\linewidth]{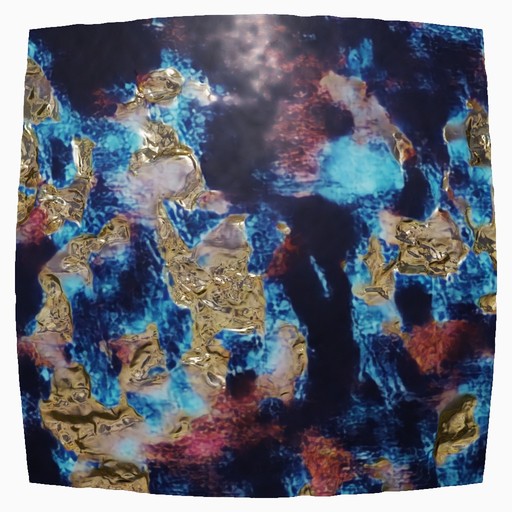} &
        \includegraphics[width=0.1225\linewidth]{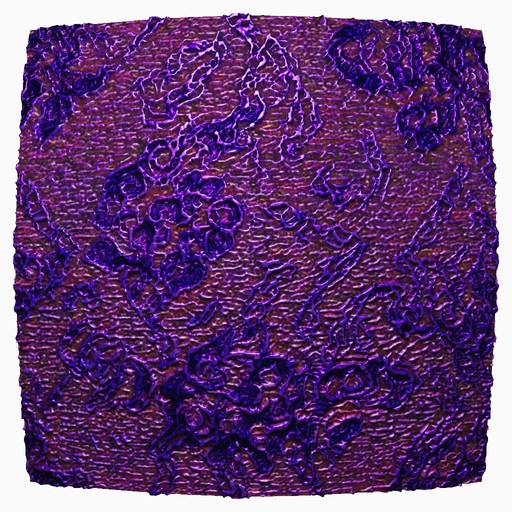} &
        \includegraphics[width=0.1225\linewidth]{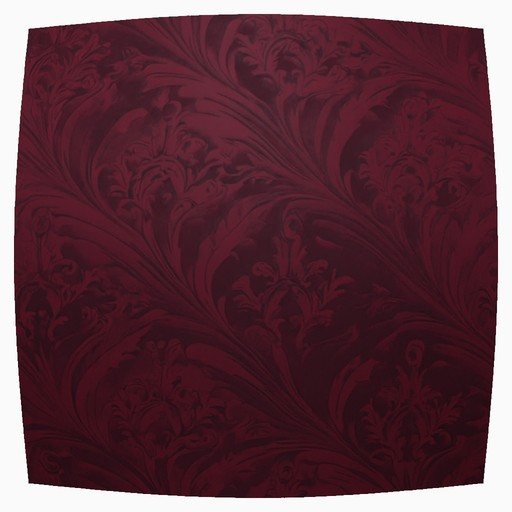} &
        \includegraphics[width=0.1225\linewidth]{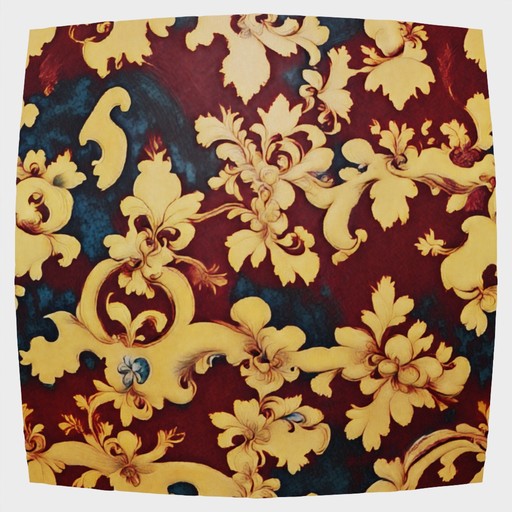} \\
        \multicolumn{4}{c}{\footnotesize{\makecell{`Cobblestone walkway covered in snow. Stones in different sizes and colors.'}}} &
        \multicolumn{4}{c}{\footnotesize{`Italian velvet fabric with Renaissance art inspired prints.'}} \\

        \vspace{-1mm} \includegraphics[width=0.1225\linewidth]{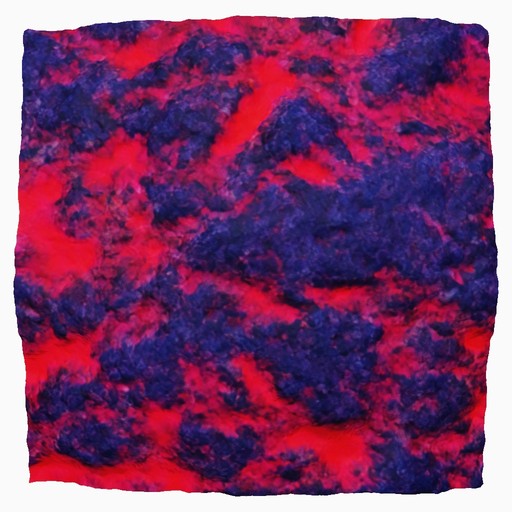} &
        \includegraphics[width=0.1225\linewidth]{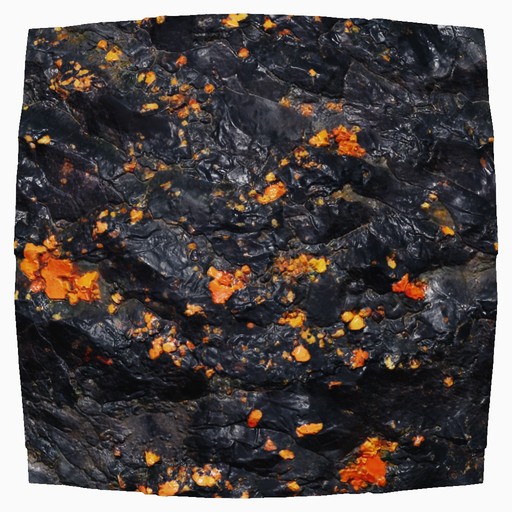} &
        \includegraphics[width=0.1225\linewidth]{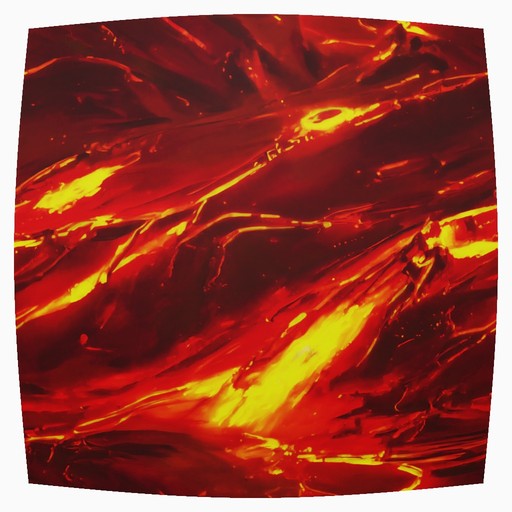} &
        \includegraphics[width=0.1225\linewidth]{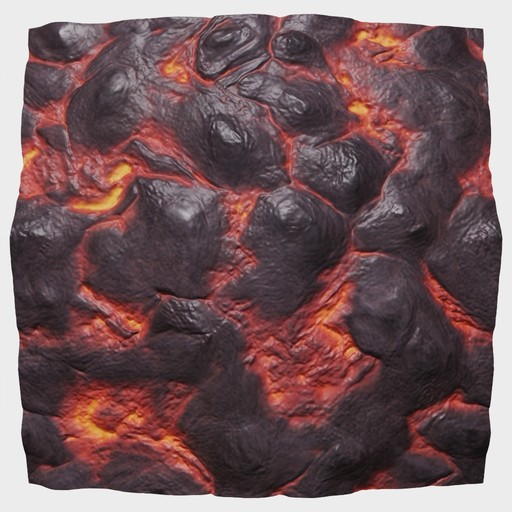} &
        \hspace{0.25mm} \includegraphics[width=0.1225\linewidth]{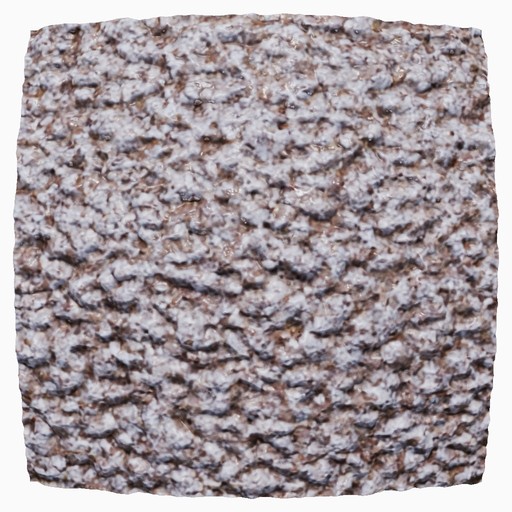} &
        \includegraphics[width=0.1225\linewidth]{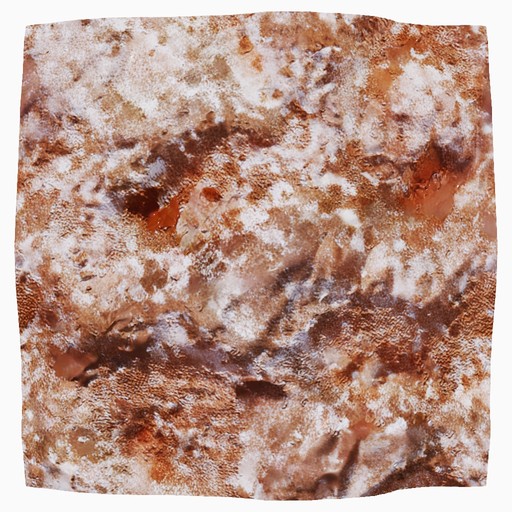} &
        \includegraphics[width=0.1225\linewidth]{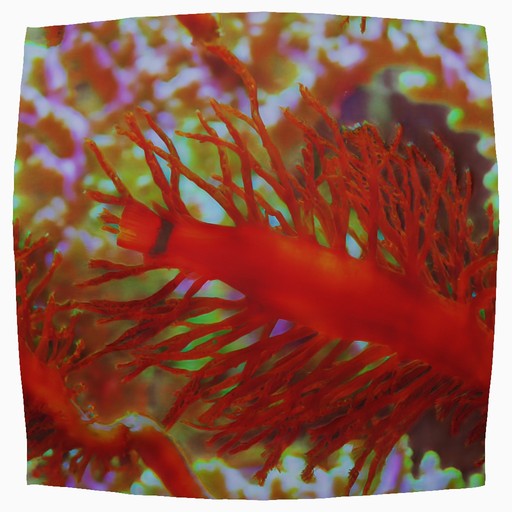} &
        \includegraphics[width=0.1225\linewidth]{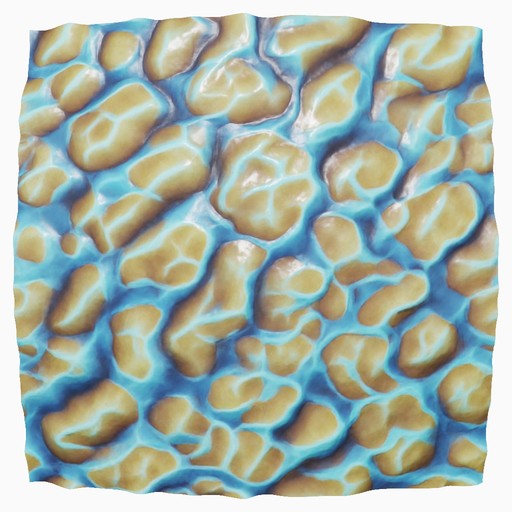} \\
        \multicolumn{4}{c}{\footnotesize{`Volcanic lava.'}} &
        \multicolumn{4}{c}{\footnotesize{`Coral pavona with a bumpy and porous texture.'}} \\
        
    \end{tabular}
    
    \caption{\textbf{Comparison for text prompting}. We compare \methodName with MatFuse, MatGen, and Substance 3D Sampler on text-prompted generation, showing two in-domain (left column) and two out-domain (right column) renderings per model. \methodName better follows the input prompt and successfully models 
    out-domain materials. Visual quality is on par or better than models trained on larger datasets.}
    \label{fig:comparison_text}
\end{figure*}

\noindent\textbf{Latent Diffusion model} is trained supervisedly for $400,000$ iterations with a batch size of $16$ using an AdamW~\cite{adamw} optimizer, with a learning rate of $3.2\cdot10^{-5}$. 
The model is fine-tuned semi-supervisedly for $200,000$ iterations with a batch size of $8$. 
We fine-tune the refiner model for $50,000$ iterations using a batch size of $16$ and a learning rate of $2\cdot10^{-7}$. 
Both models use the original OpenAI U-Net architecture with 18 input and output channels. 

\noindent\textbf{Latent Consistency model} is fine-tuned for $10,000$ iterations, with a batch size of $16$, using an AdamW %
optimizer, with a learning rate value of $1\cdot10^{-6}$. We use a linear schedule for $\beta$ and denoise using the LCM %
sampling schedule at inference time with $T=4$ steps.

\paragraph{Inference}
\label{sec:inference}
We assess execution speed and memory usage. We generate with 4 denoising steps, followed by 2 refinement steps, using the LCM sampler with a fixed seed and processing up to 8 patches in parallel at half-precision.
Generation takes 0.6s at $512\times512$, 1.5s at $1024\times1024$ and 6.5GB of VRAM, 4.9s at $2048\times2048$ and 7.4GB VRAM, and 18.6s at $4096\times4096$ and 12GB of VRAM. 
In contrast, an LDM with a DDIM sampler (50 steps) plus 25 refinement steps requires 20.6s at $2048\times2048$, and 65.4s at $4096\times4096$.

\subsection{Results and comparison}
All results show both generated material maps and ambient-lit renderings. 
For evaluation, we carefully selected test prompts to include both in-domain concepts (present in training categories) and novel out-domain concepts, ensuring no direct overlap with training prompts.
\suppmat{Additional samples, material editing results, and a CLIP-based nearest-neighbor search are included in the Supplemental Materials.}

\paragraph{Generation results}
\label{sec:gen_results}
We show the generative capabilities of our model, for both image (Fig.~\ref{fig:generation_img}) and text (Fig.~\ref{fig:generation_text}) conditioning. %
We include both \emph{in-domain} samples (categories found in the annotated dataset) and \emph{out-domain} samples (from unannotated data). In all cases, \methodName produces realistic results that closely follow the prompts.

\paragraph{Qualitative comparison}
\label{sec:comparison}
We compare \methodName against MatFuse~\cite{vecchio2023matfuse}, MatGen~\cite{vecchio2024controlmat}, Material Palette~\cite{lopes2023material}, and Adobe Substance 3D Sampler~\cite{sampler}, as shown in Figures~\ref{fig:comparison_img} and \ref{fig:comparison_text} for image and text prompting, respectively.
MatFuse, Material Palette, and \methodName each use public datasets, while MatGen and Sampler rely on private data. %
MatFuse is limited to a $256\times256$ resolution, often generating blurry or simplistic outputs and struggling with complex textures. Material Palette and Sampler take a two-step approach (texture generation followed by SVBRDF estimation).%
This approach benefits from leveraging large image models, but suffers from biases inherent in SVBRDF predictions, which guess material properties from light-surface interactions, that are further exacerbated by the nonmaterial-specific training sometimes generating natural images (with perspective) instead of surfaces.
Moreover, MaterialPalette fine-tunes a LoRA for each prompt, introducing a time and computation overhead. 
MatGen, on the other hand, produces high-quality materials, but presents over-sharpening artifacts and struggles in following more complex promps.
In contrast, \methodName directly outputs PBR maps using a tailored latent representation and a two-stage generation process, mitigating resolution constraints and reducing artifacts.
\suppmat{We only show the generation renderings in Figures~\ref{fig:comparison_img} and~\ref{fig:comparison_text}, with the PBR maps and additional samples included in the Supplemental Materials.}

\paragraph{User Study}
\label{sec:user}
To validate the quality of our results, we conducted a user study with 20 domain experts. Participants compared material generations from the different methods across ten prompts. For each prompt, they selected their preferred sample and rated all four materials on a 1--5 scale for \textbf{prompt fidelity}, \textbf{realism}, and \textbf{material consistency}.
\textit{StableMaterials} was preferred 105 times, compared to 66 for \textit{Sampler}, 25 for \textit{MatGen}, and 4 for \textit{MatFuse}. It also achieved the highest average rating (3.50), outperforming \textit{Sampler} (2.71), \textit{MatGen} (2.28), and \textit{MatFuse} (1.85). A one-way ANOVA confirmed the differences are statistically significant ($F = 60.5$, $p < 0.05$). A chi-square goodness-of-fit test on the preference counts $[4, 25, 66, 105]$ against a uniform distribution yielded $\chi^2 = 120.44$, $p < 0.001$, indicating strong deviation from chance. A correlation coefficient of $R = 0.99$ confirms alignment between average ratings and preference counts.

\paragraph{Quantitative Comparison} 
\label{sec:comparison_quantitative} 
Material quality is difficult to assess with standard image metrics (FID~\cite{heusel2017gans}, IS~\cite{salimans2016improved}) due to the different data distribution from that of natural images. Instead, we leverage CLIP-based metrics (CLIP Score~\cite{hessel2021clipscore} and CLIP-IQA~\cite{wang2023exploring}), which assess both semantic alignment with prompts and perceptual quality of outputs. 
Table~\ref{tab:quantitative} shows \methodName either outperforms or matches state-of-the-art methods—some trained on larger datasets—and significantly improves over MatFuse. Comparison is carried out on 80 text-conditioned generation.

\begin{table}
    \begin{tabular}{lcccc}
        \toprule
        & \small{MatFuse} & \small{MatGen} & \small{\makecell{Substance \\ 3D Sampler}} & \small{\makecell{\textbf{Stable} \\ \textbf{Materials}}} \\
        \hline
        \textbf{CLIP Score} $\uparrow$ & 26.2 & 28.8 & 24.9 & \textbf{29.6} \\
        \textbf{CLIP-IQA} $\uparrow$ & 0.52 & 0.66 & \textbf{0.71} & \textbf{0.70} \\
        \bottomrule
    \end{tabular}
    \caption{\textbf{Quantitative comparison}. We compare the generation quality of \methodName to MatFuse, MatGen and Substance 3D Sampler. The CLIP-IQA score is computed using the ``high-quality/low-quality'' contrastive pair.}
    \label{tab:quantitative}
\end{table}

\subsection{Ablation Study}
\label{sec:ablation}

We evaluate our different design choices by comparing the performance of our model against the baseline solutions.
\suppmat{We provide high-resolution ablation results in the Supplemental Materials.}

\paragraph{VAE Architecture}
\label{sec:ablation_encoder}

\begin{table}    
    \begin{tabular}{lccccc}
        \toprule
         & Albedo & Normal & Height & Rough. & Metal. \\
        \hline
        \makecell{Multi-$\mathcal{E}$ \\ \footnotesize{(271M par.)}} & 0.030 & \textbf{0.035} & \textbf{0.030} & \textbf{0.032} & 0.016\\ 
        \makecell{Single-$\mathcal{E}$ \\ \footnotesize{(101M par.)}} & \textbf{0.028} & 0.037 & \textbf{0.030} & \textbf{0.032} & \textbf{0.015}\\ 
        \bottomrule
    \end{tabular}
    \caption{\textbf{Analysis of the VAE architecture}. We report the RMSE$\downarrow$ between reconstructed and ground-truth maps. The single-encoder model, having less parameters, achieves performances on par with the multi-encoder VAE.}
    \label{tab:ablation_vae}
\end{table}

We report the reconstruction performance, in terms of RMSE, for both the multi-encoder and single-encoder models in Tab.~\ref{tab:ablation_vae}. Results show that the adopted transfer-learning strategy allows us to obtain a smaller network while achieving performances comparable to those of the larger multiencoder VAE and retaining the same dientangled latent space.

\begin{figure}
    \centering
    \setlength{\tabcolsep}{.5pt}
    \begin{tabular}{cccc}
        \multicolumn{2}{c}{\textbf{In-domain samples}} & \multicolumn{2}{c}{\textbf{Out-domain samples}}\\
        
        \hline
        
        \hspace{-1mm}\small{W/o distillation} & \small{W/ distillation} & \small{W/o distillation} & \small{W/ distillation}\\
        
        \vspace{-1mm}\includegraphics[width=0.25\linewidth, height=0.25\linewidth]{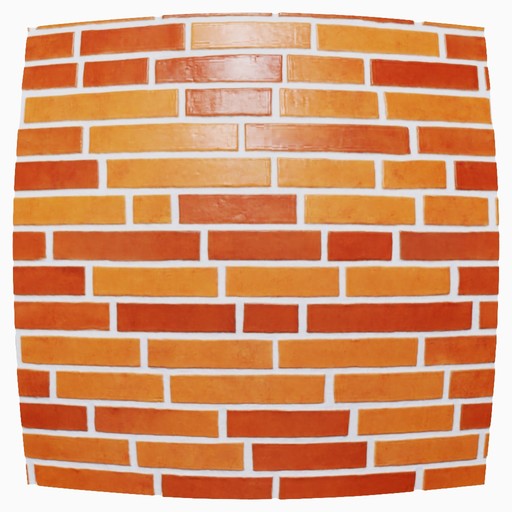} &
        \includegraphics[width=0.25\linewidth, height=0.25\linewidth]{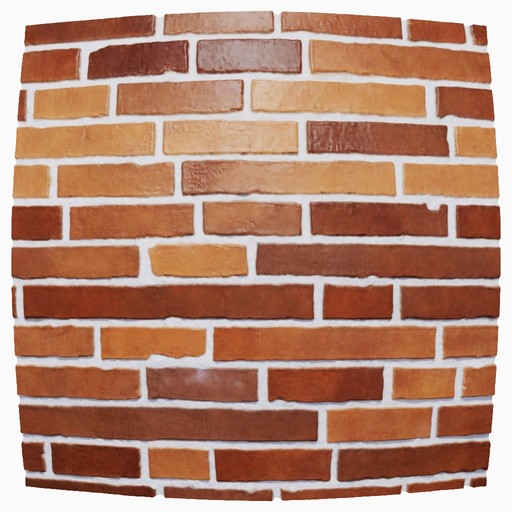} &
        \includegraphics[width=0.25\linewidth, height=0.25\linewidth]{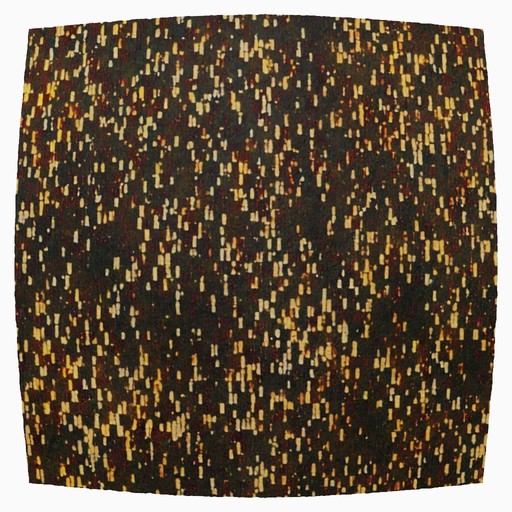} &
        \includegraphics[width=0.25\linewidth, height=0.25\linewidth]{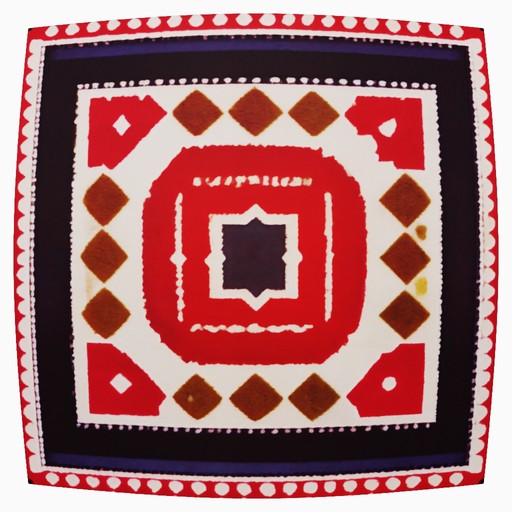} \\
        \multicolumn{2}{c}{\footnotesize{\makecell{`Terracotta brick wall with \\ white grout.'}}} & \multicolumn{2}{c}{\footnotesize{\makecell{`Kilim Rugs carpets with complex \\ geometric patterns.'}}}\\
               
    \end{tabular}
    
    \caption{\textbf{Ablation study of the training stategies}. The model effectively generates high-quality material with image or text prompts but struggles with unrepresented materials. Semi-supervised learning improves generation quality and diversity, including new materials. }
    \label{fig:ablation_train}
\end{figure}

\paragraph{Training strategies}
\label{sec:ablation_training}
We evaluate the effect of the semi-supervised training on model's generation capabilities in Fig.~\ref{fig:ablation_train}. 
Our semi-supervised training improves generation diversity, especially on out-of-domain materials that the purely supervised baseline cannot produce reliably.

\begin{figure}
    \centering
    \setlength{\tabcolsep}{.5pt}
    \begin{tabular}{cccc}
        \hspace{-1mm}\small{Base} & \small{Refined} & \small{\makecell{Patch \\ diffusion}} & \small{\makecell{Two-stages \\ pipeline}}\\
        
        \vspace{-1mm}\includegraphics[width=0.25\linewidth]{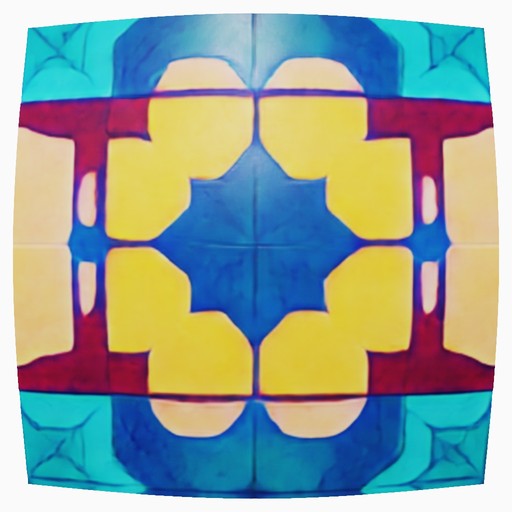} &
        \includegraphics[width=0.25\linewidth]{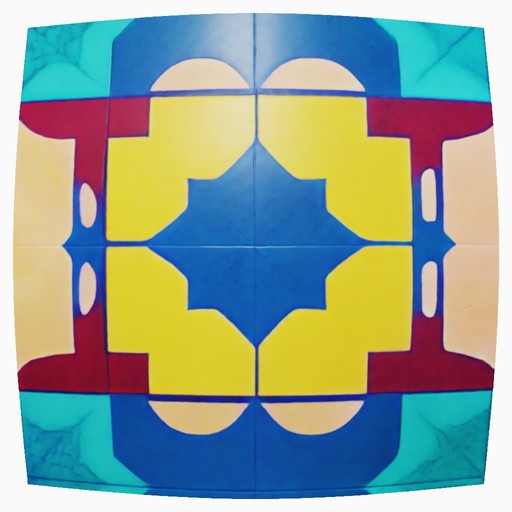} &
        \includegraphics[width=0.25\linewidth, height=0.25\linewidth]{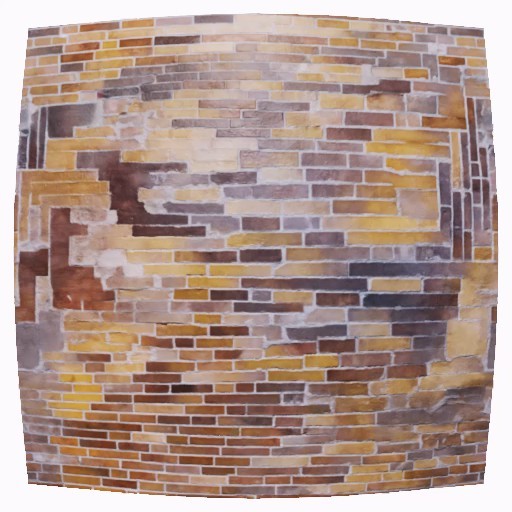} &
        \includegraphics[width=0.25\linewidth, height=0.25\linewidth]{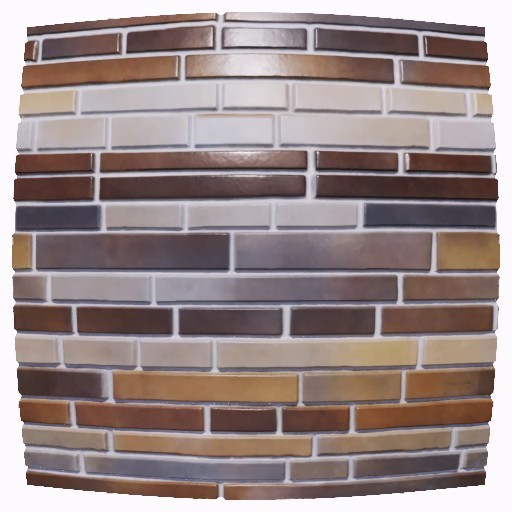} \\ 
    \end{tabular}

    \begin{subfigure}{0.495\linewidth}
        \caption{\textbf{Refinement ablation}.}
        \label{fig:ablation_refine}
    \end{subfigure}
    \hfill
    \begin{subfigure}{0.495\linewidth}
        \caption{\textbf{High-resolution ablation}.}
        \label{fig:ablation_stages}
    \end{subfigure}
    
    \caption{Ablation studies comparing (a) the effect of refinement on quality and (b) different approaches to achieving high-resolution. Results show that the use of a diffusion refiner significantly enhances generation quality and sharpness, and two-stage approach generates at the model's native resolution avoids the inconsistency in scale of the patched diffusion.}
    \label{fig:ablations_hd}
\end{figure}

\paragraph{High-Resolution Generation} 
\label{sec:two_stage} 
Our two-stage approach to high-resolution sharpens and improves details (Fig.\ref{fig:ablation_refine}). Additionally, using only patched diffusion can introduce scale and consistency artifacts across patches, whereas our two-stage approach (base generation + refinement) produces more coherent large-format outputs by consolidating patches at $512\times512$ before upscaling (Fig.\ref{fig:ablation_stages}).

\section{Limitations and Future Work}
\label{sec:limitation}

\begin{figure}
    \centering
    \setlength{\tabcolsep}{.5pt}
    \begin{tabular}{ccc}
        (1) & (2) & (3)\\
        
        \vspace{-1mm}\includegraphics[width=0.3\linewidth]{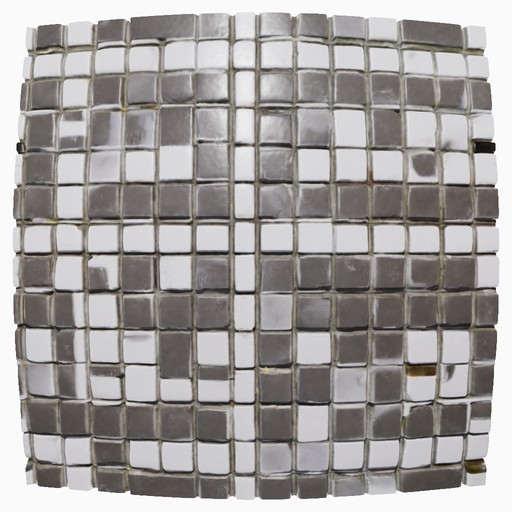} &
        \includegraphics[width=0.3\linewidth]{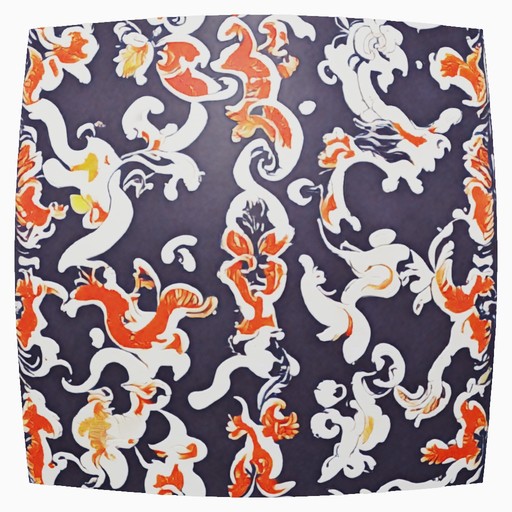} &
        \includegraphics[width=0.3\linewidth]{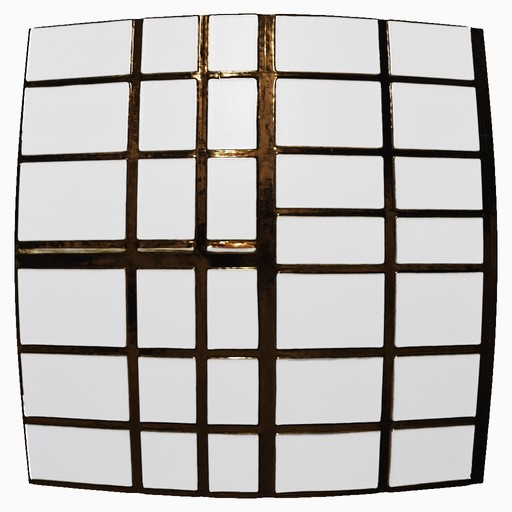} \\

        \textit{\footnotesize{\makecell{`Squared tiles enclosed \\ by rectangular tiles.'}}} & 
        \textit{\footnotesize{\makecell{`Yukata fabric with \\ patterns of dragons.'}}} & 
        \textit{\footnotesize{\makecell{`Shoji Screens made \\ of paper and wood.'}}} \\
    \end{tabular}
    
    \caption{\textbf{Limitations}. Left to right: (1) Struggles with complex prompts describing spatial relations. (2) Unable to represent complex figures or patterns. (3) Can hallucinate reflectance properties.}
    \label{fig:limitation}
\end{figure}

\methodName presents some limitations as shown in Fig.~\ref{fig:limitation}. First, it struggles with natural prompts describing spatial relations and is unable to accurately represent complex concepts or figures.
Introducing more variety in training prompts could help mitigate the problem. 
Additionally, it can occasionally generate incorrect reflectance properties (e.g.: material misclassified as metal) for material classes that are only present in the unannotated dataset. The use of text prompts describing surface properties, at training time, could mitigate the problem. 
Finally, despite being able to represent a wide variety of materials outside the annotated dataset, it is still limited to the classes represented in the unannotated data.

\section{Conclusion}
\label{sec:conclusion}

We introduce \methodName, a novel diffusion-based model for fast, tileable, high-resolution material generation. By integrating semi-supervised training and knowledge distillation from large-scale pretrained models, it overcomes the lack of annotated data and delivers greater realism and variety in PBR materials. Moreover, features rolling enables tileable generation in few-steps settings, making our approach practical for real-world applications.

We believe that \methodName can serve as a blueprint for future research, demonstrating how domain-specific generators can effectively leverage large-scale pretrained models and unsupervised data to improve diversity while maintaining physical plausibility and generation quality.

{
    \small
    \bibliographystyle{ieeenat_fullname}
    \bibliography{bibliography}
}

\begin{figure*}
    \centering
    \setlength{\tabcolsep}{.5pt}
    \begin{tabular}{ccccccc}
        \small{Prompt} & \small{Basecolor} & \small{Normal} & \small{Height} & \small{Roughness} & \small{Metallic} & \small{Rendering}\\
        
        \vspace{-1mm}\includegraphics[width=0.14\linewidth, height=0.14\linewidth]{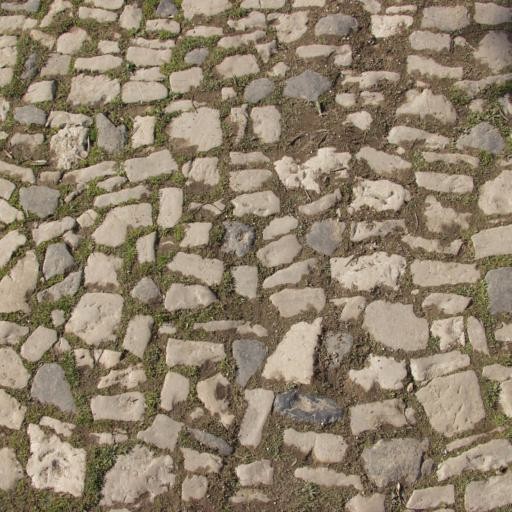} &
        \includegraphics[width=0.14\linewidth, height=0.14\linewidth]{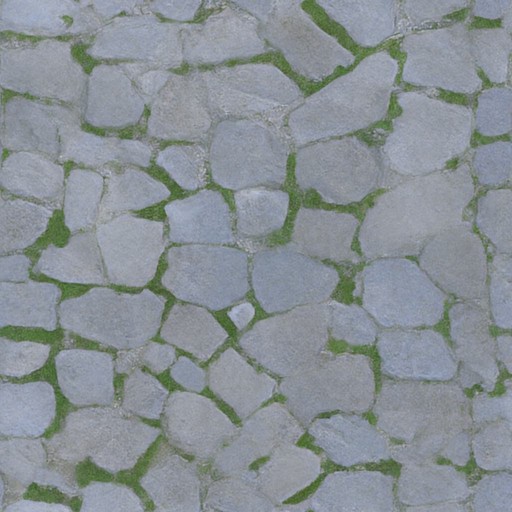} &
        \includegraphics[width=0.14\linewidth, height=0.14\linewidth]{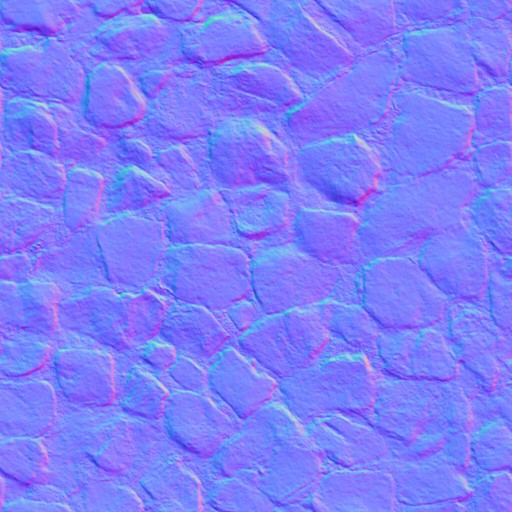} &
        \includegraphics[width=0.14\linewidth, height=0.14\linewidth]{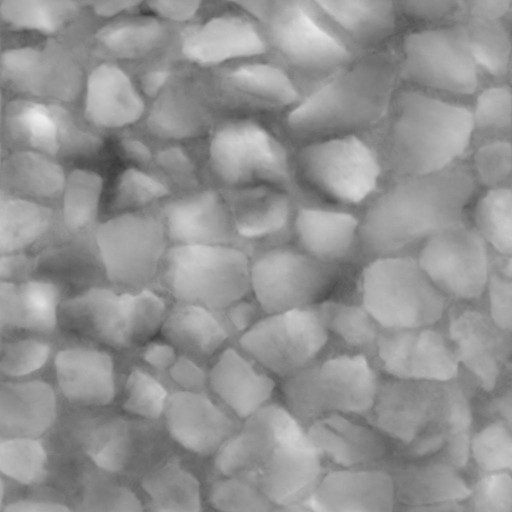} &
        \includegraphics[width=0.14\linewidth, height=0.14\linewidth]{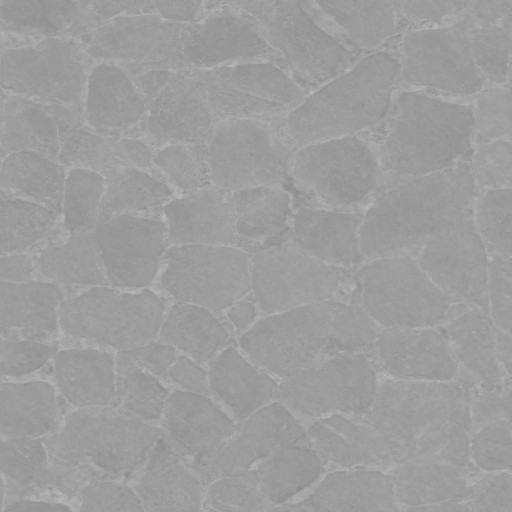} &
        \includegraphics[width=0.14\linewidth, height=0.14\linewidth]{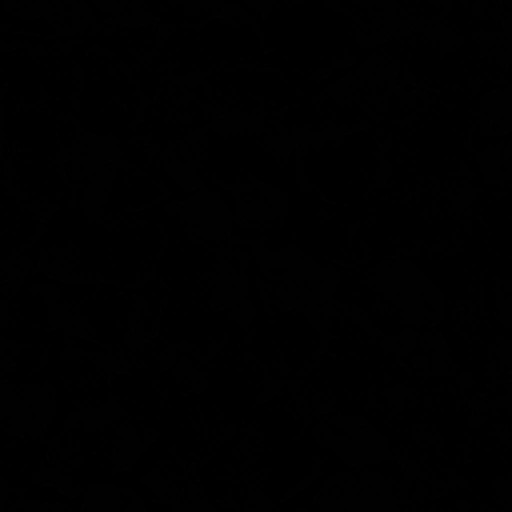} &
        \includegraphics[width=0.14\linewidth, height=0.14\linewidth]{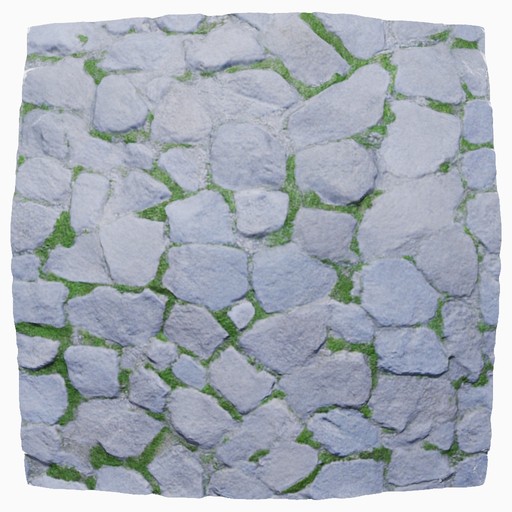} \\
    
        \vspace{-1mm}\includegraphics[width=0.14\linewidth, height=0.14\linewidth]{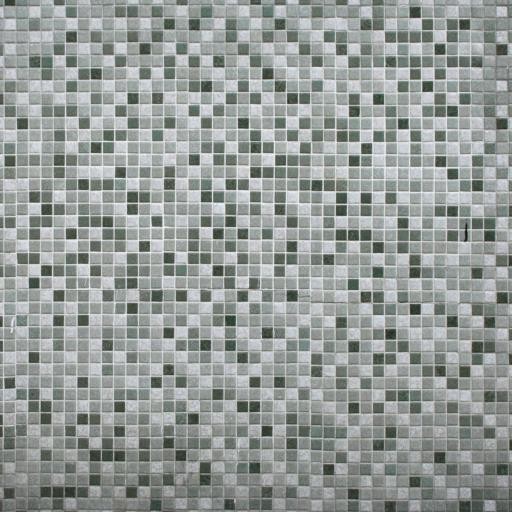} &
        \includegraphics[width=0.14\linewidth, height=0.14\linewidth]{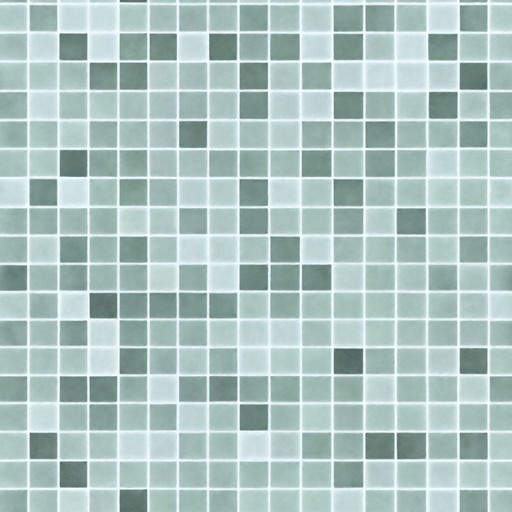} &
        \includegraphics[width=0.14\linewidth, height=0.14\linewidth]{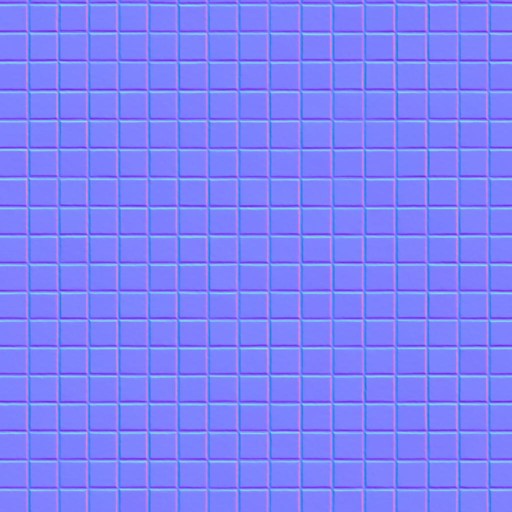} &
        \includegraphics[width=0.14\linewidth, height=0.14\linewidth]{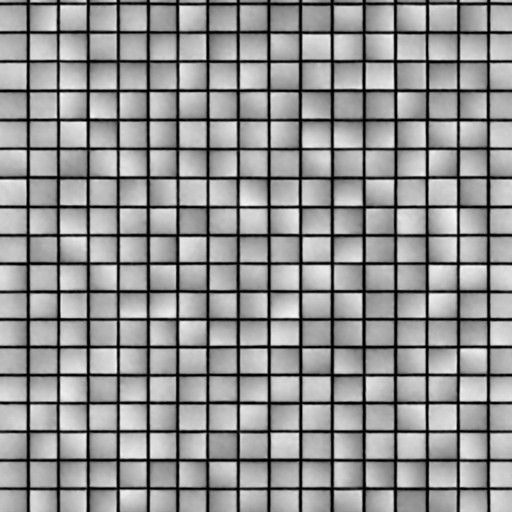} &
        \includegraphics[width=0.14\linewidth, height=0.14\linewidth]{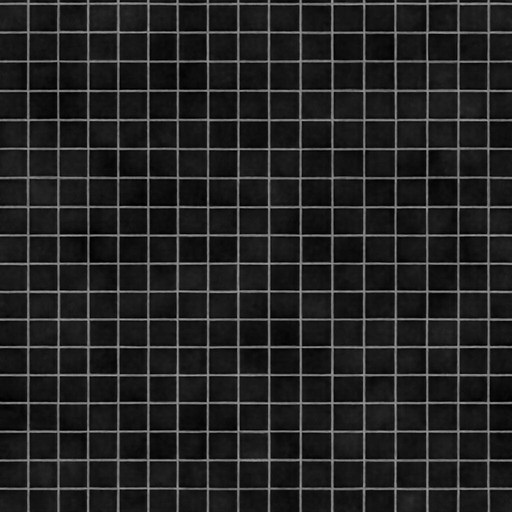} &
        \includegraphics[width=0.14\linewidth, height=0.14\linewidth]{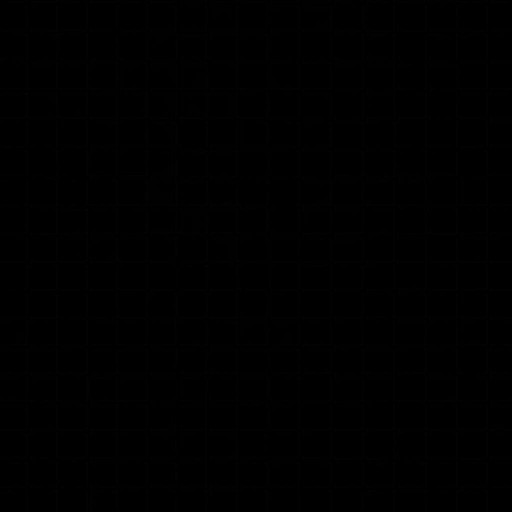} &
        \includegraphics[width=0.14\linewidth, height=0.14\linewidth]{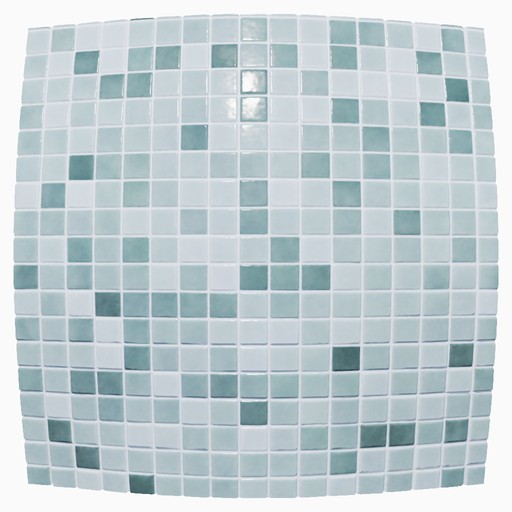} \\
    
        \vspace{-1mm}\includegraphics[width=0.14\linewidth, height=0.14\linewidth]{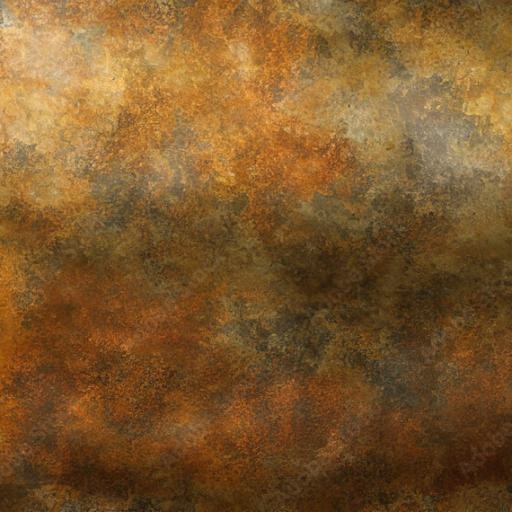} &
        \includegraphics[width=0.14\linewidth, height=0.14\linewidth]{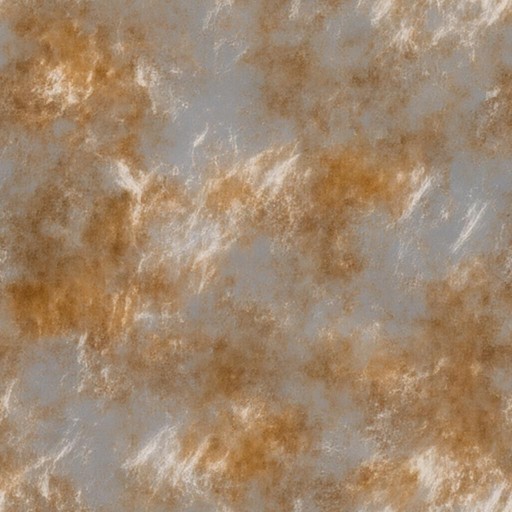} &
        \includegraphics[width=0.14\linewidth, height=0.14\linewidth]{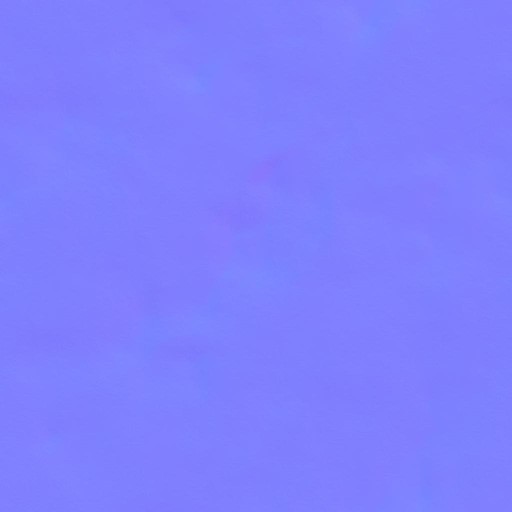} &
        \includegraphics[width=0.14\linewidth, height=0.14\linewidth]{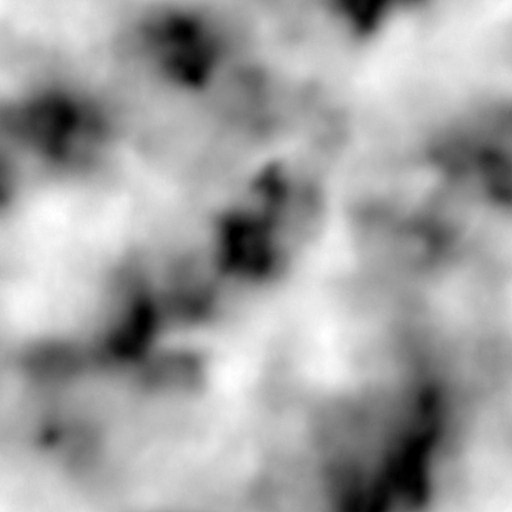} &
        \includegraphics[width=0.14\linewidth, height=0.14\linewidth]{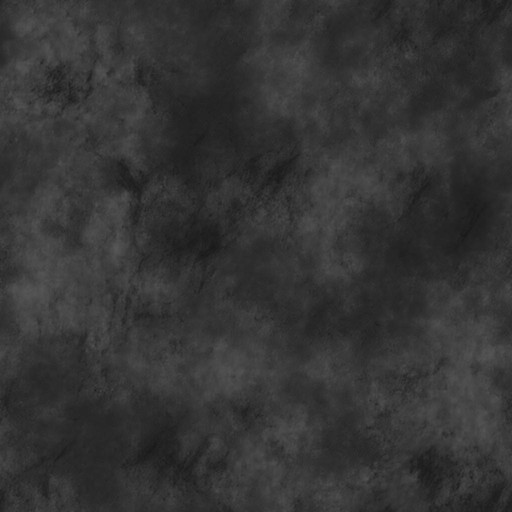} &
        \includegraphics[width=0.14\linewidth, height=0.14\linewidth]{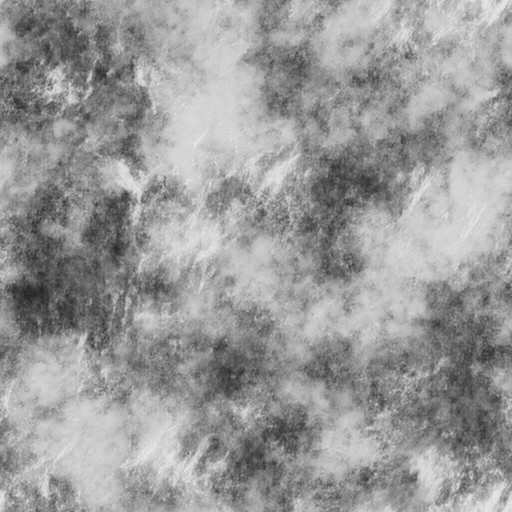} &
        \includegraphics[width=0.14\linewidth, height=0.14\linewidth]{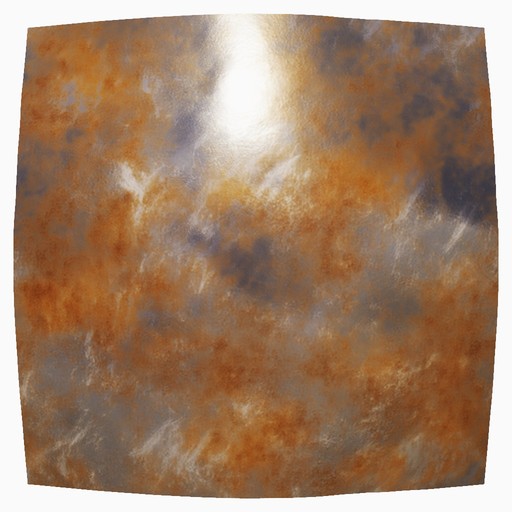} \\

        \vspace{-1mm}\includegraphics[width=0.14\linewidth, height=0.14\linewidth]{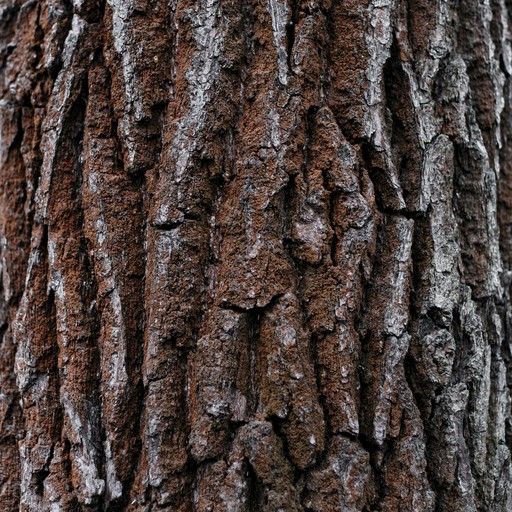} &
        \includegraphics[width=0.14\linewidth, height=0.14\linewidth]{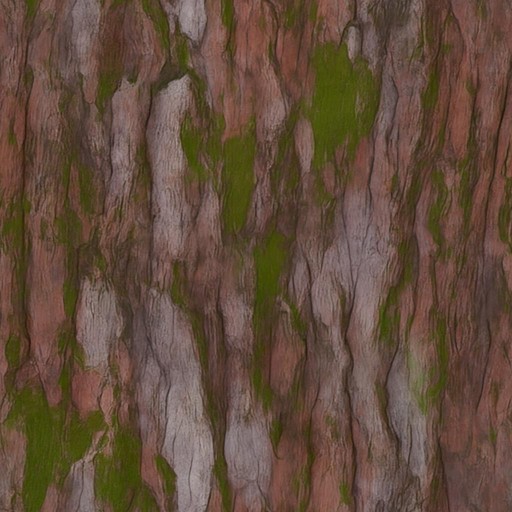} &
        \includegraphics[width=0.14\linewidth, height=0.14\linewidth]{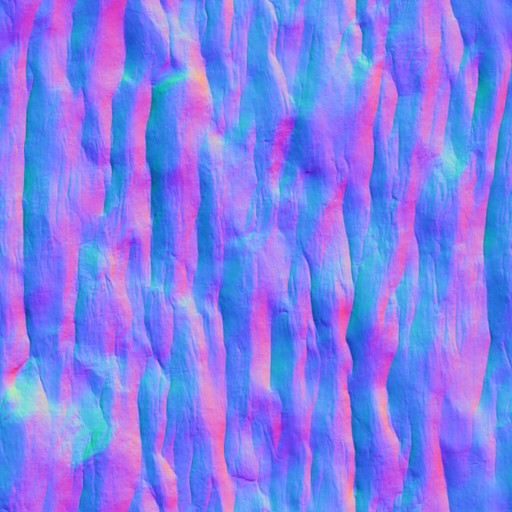} &
        \includegraphics[width=0.14\linewidth, height=0.14\linewidth]{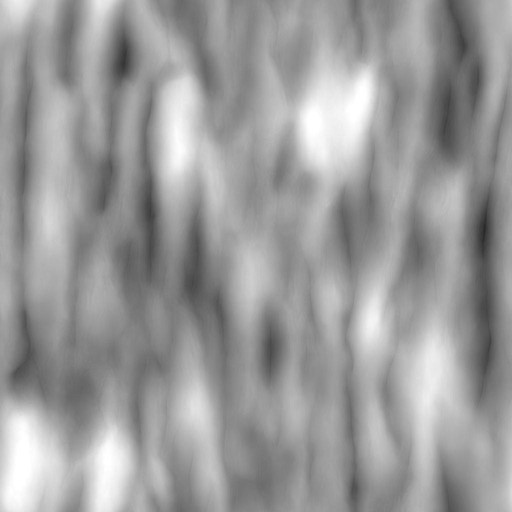} &
        \includegraphics[width=0.14\linewidth, height=0.14\linewidth]{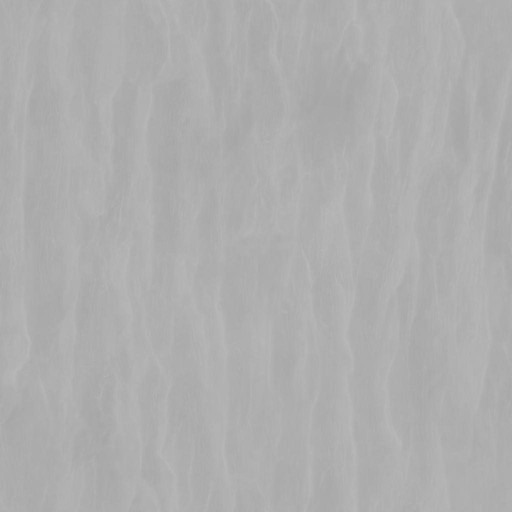} &
        \includegraphics[width=0.14\linewidth, height=0.14\linewidth]{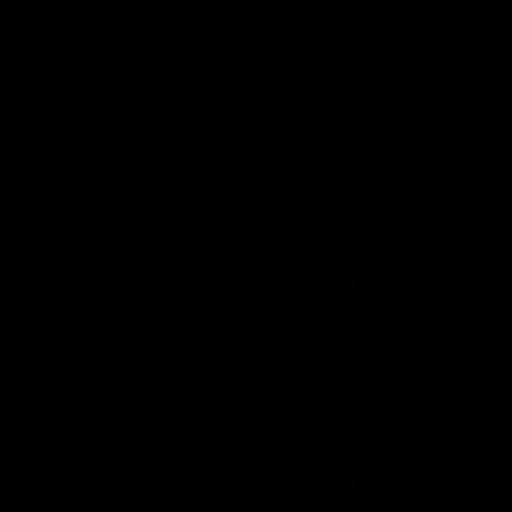} &
        \includegraphics[width=0.14\linewidth, height=0.14\linewidth]{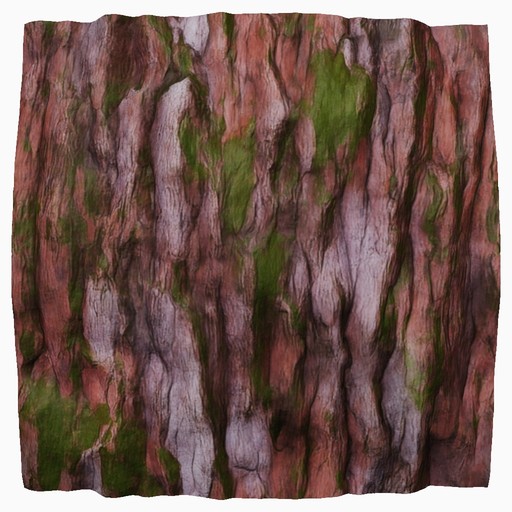} \\

        \vspace{-1mm}\includegraphics[width=0.14\linewidth, height=0.14\linewidth]{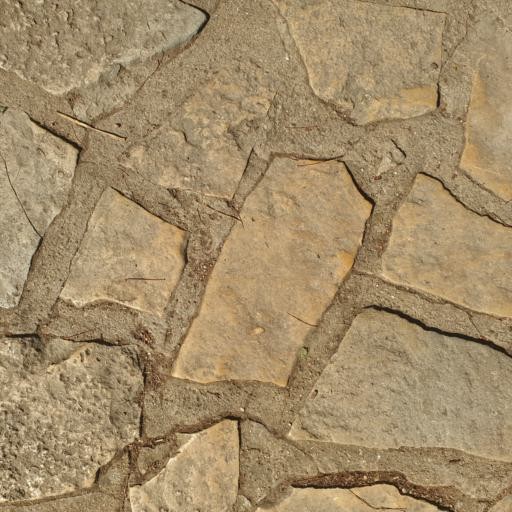} &
        \includegraphics[width=0.14\linewidth, height=0.14\linewidth]{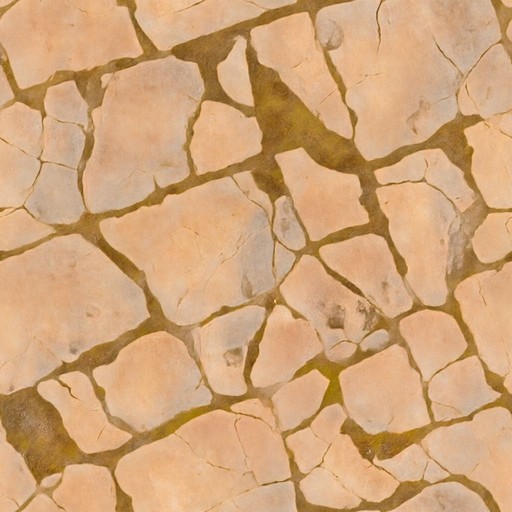} &
        \includegraphics[width=0.14\linewidth, height=0.14\linewidth]{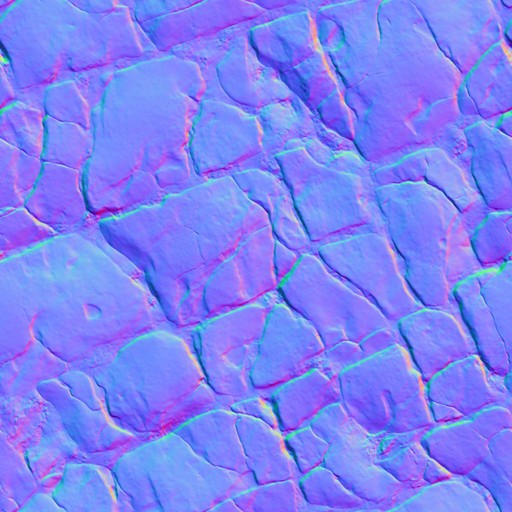} &
        \includegraphics[width=0.14\linewidth, height=0.14\linewidth]{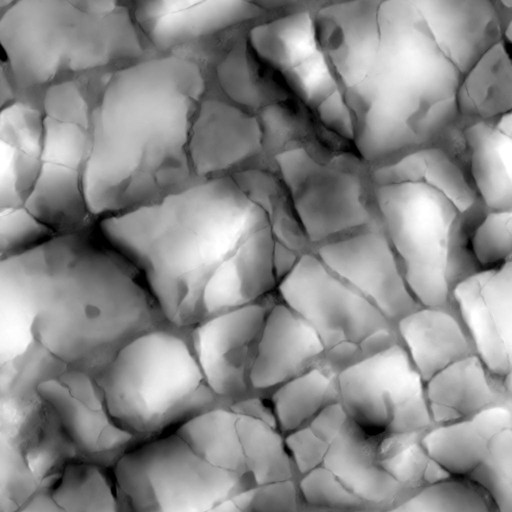} &
        \includegraphics[width=0.14\linewidth, height=0.14\linewidth]{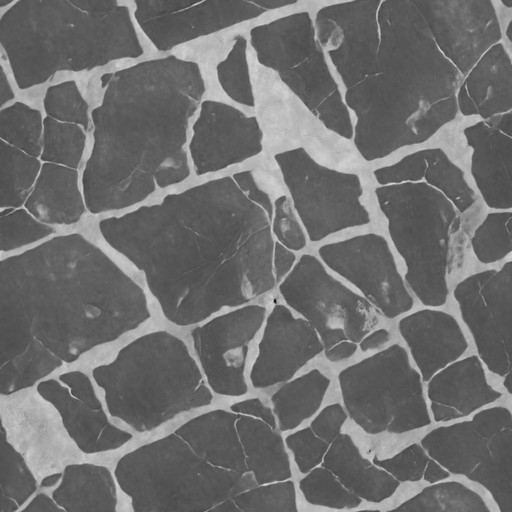} &
        \includegraphics[width=0.14\linewidth, height=0.14\linewidth]{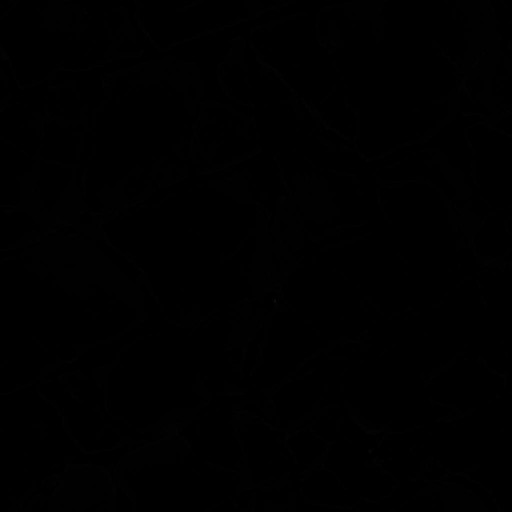} &
        \includegraphics[width=0.14\linewidth, height=0.14\linewidth]{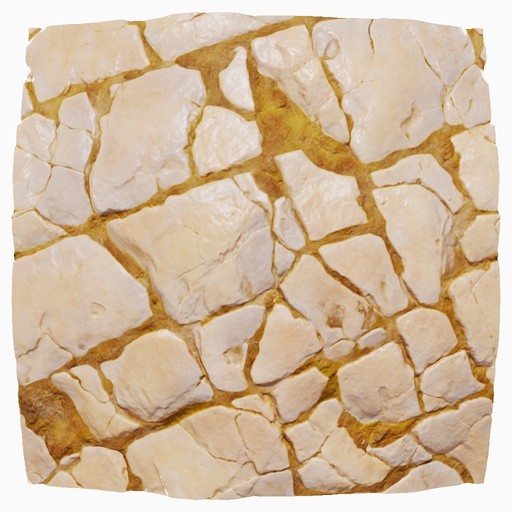} \\
        
        \vspace{-1mm}\includegraphics[width=0.14\linewidth, height=0.14\linewidth]{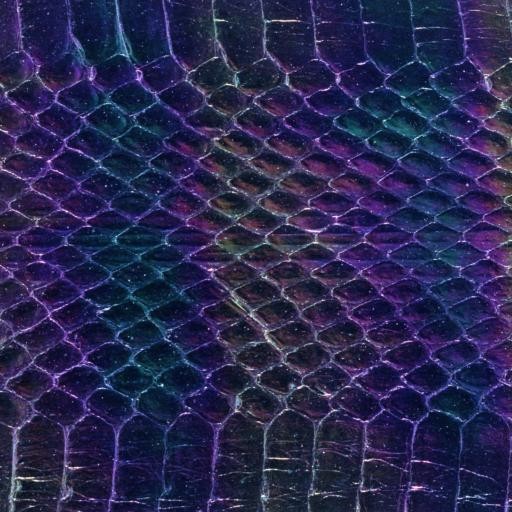} &
        \includegraphics[width=0.14\linewidth, height=0.14\linewidth]{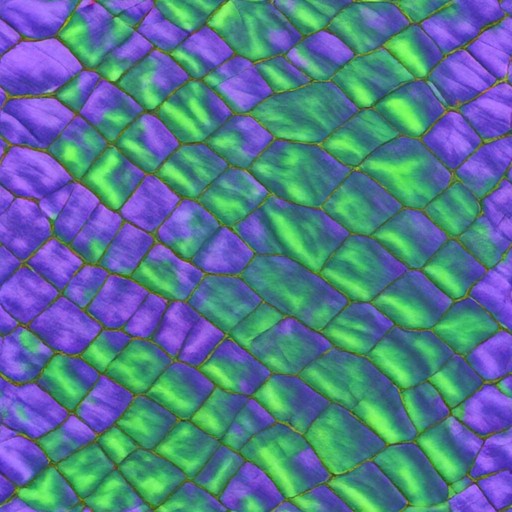} &
        \includegraphics[width=0.14\linewidth, height=0.14\linewidth]{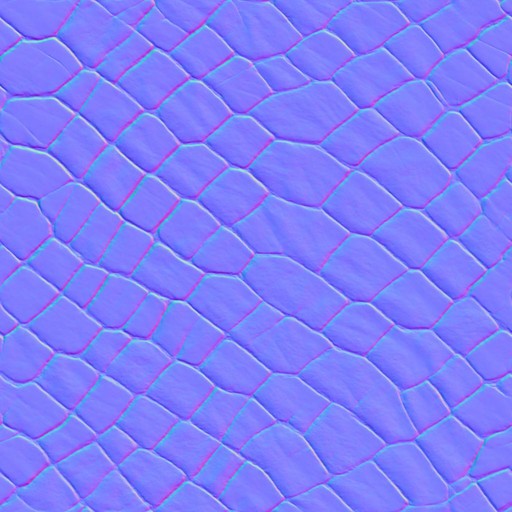} &
        \includegraphics[width=0.14\linewidth, height=0.14\linewidth]{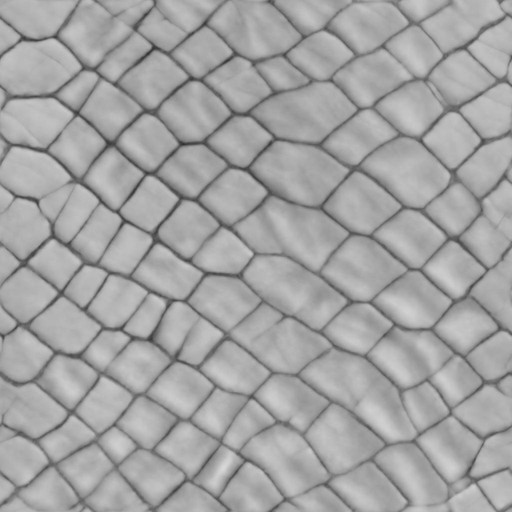} &
        \includegraphics[width=0.14\linewidth, height=0.14\linewidth]{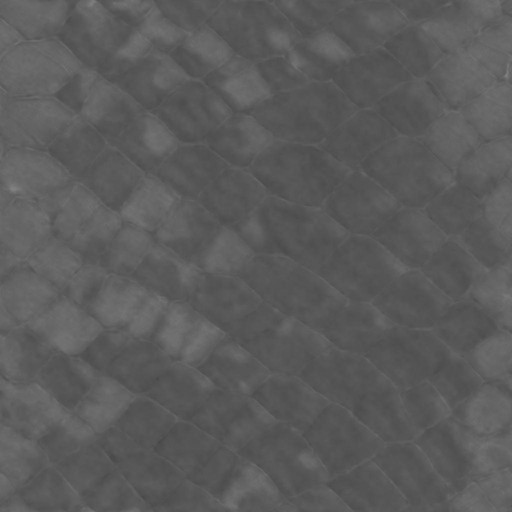} &
        \includegraphics[width=0.14\linewidth, height=0.14\linewidth]{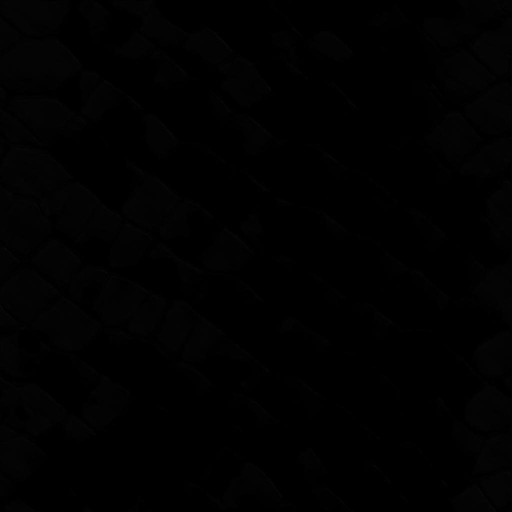} &
        \includegraphics[width=0.14\linewidth, height=0.14\linewidth]{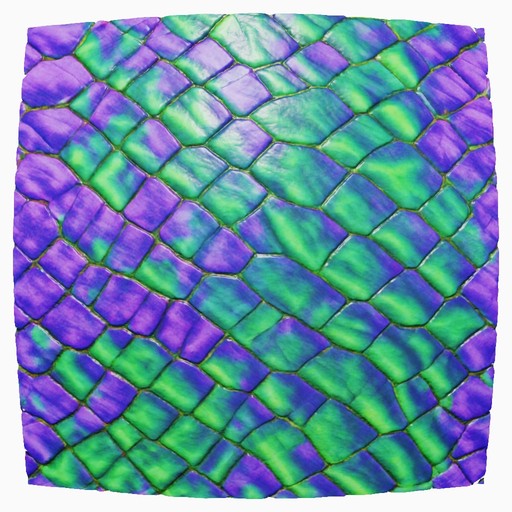} \\

        \vspace{-1mm}\includegraphics[width=0.14\linewidth, height=0.14\linewidth]{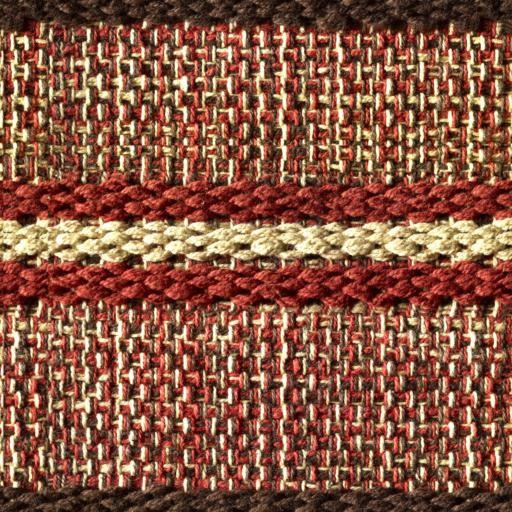} &
        \includegraphics[width=0.14\linewidth, height=0.14\linewidth]{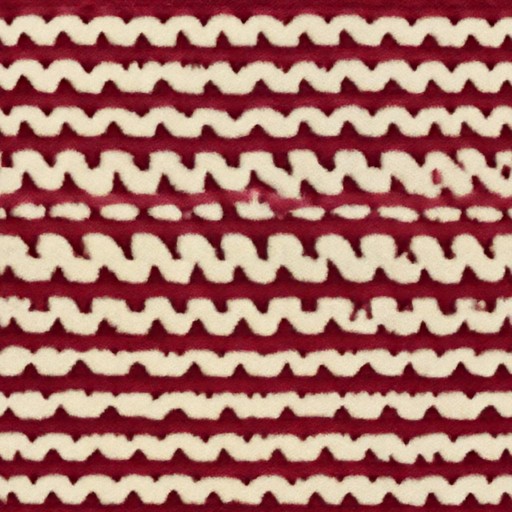} &
        \includegraphics[width=0.14\linewidth, height=0.14\linewidth]{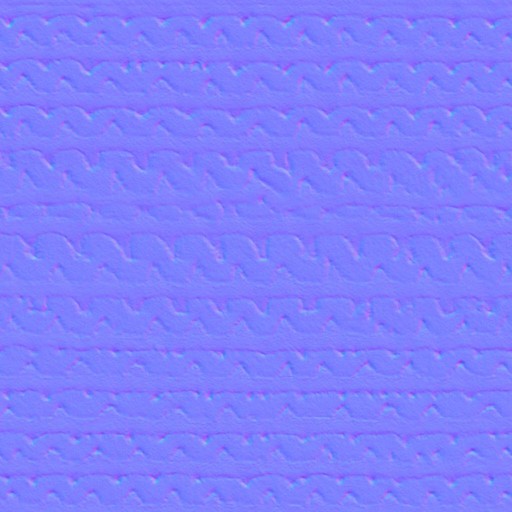} &
        \includegraphics[width=0.14\linewidth, height=0.14\linewidth]{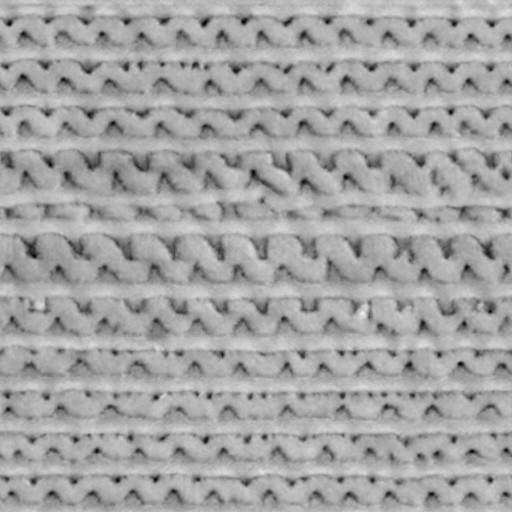} &
        \includegraphics[width=0.14\linewidth, height=0.14\linewidth]{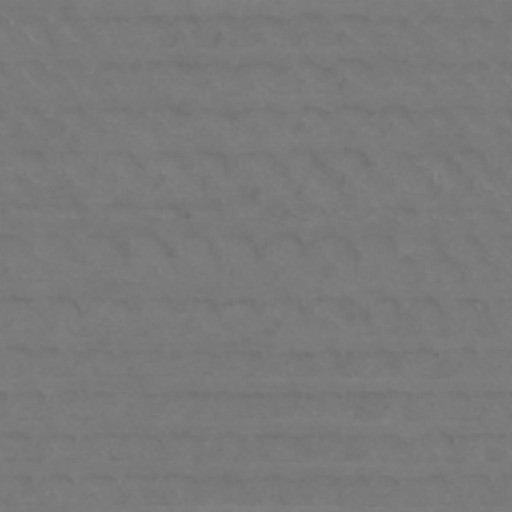} &
        \includegraphics[width=0.14\linewidth, height=0.14\linewidth]{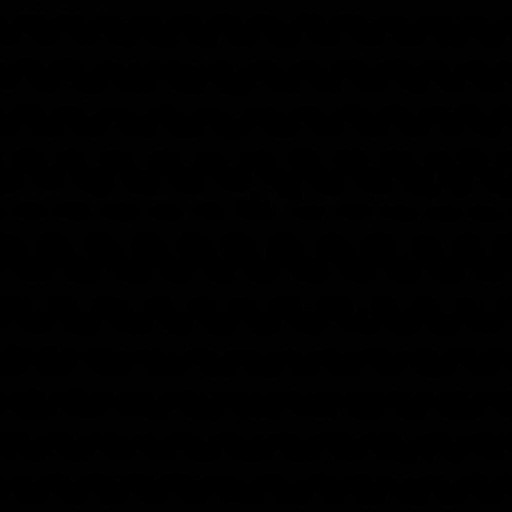} &
        \includegraphics[width=0.14\linewidth, height=0.14\linewidth]{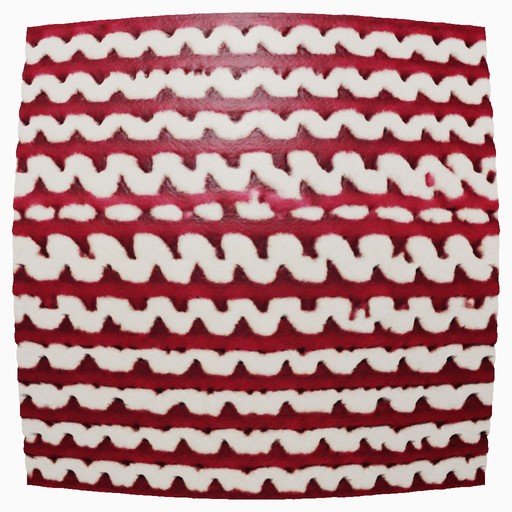} \\

        \vspace{-1mm}\includegraphics[width=0.14\linewidth, height=0.14\linewidth]{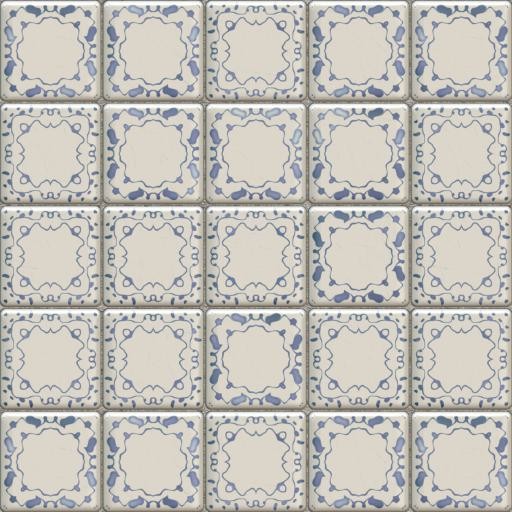} &
        \includegraphics[width=0.14\linewidth, height=0.14\linewidth]{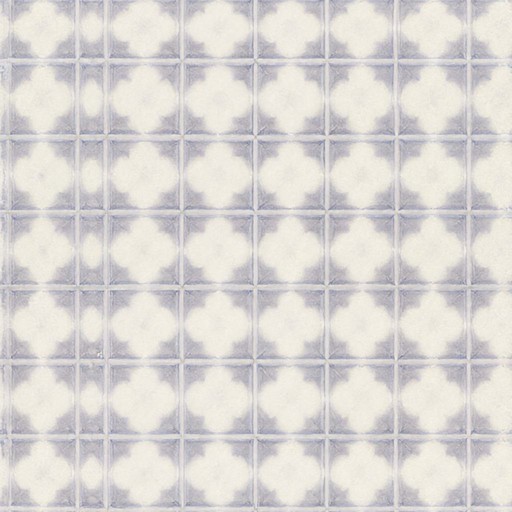} &
        \includegraphics[width=0.14\linewidth, height=0.14\linewidth]{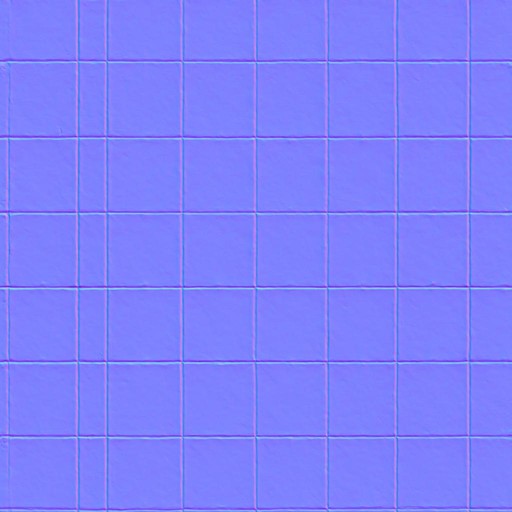} &
        \includegraphics[width=0.14\linewidth, height=0.14\linewidth]{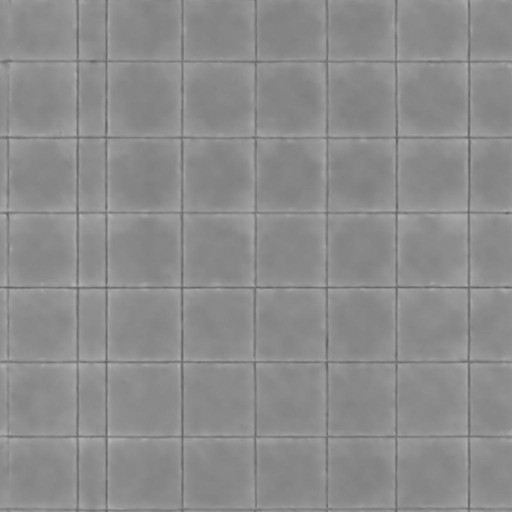} &
        \includegraphics[width=0.14\linewidth, height=0.14\linewidth]{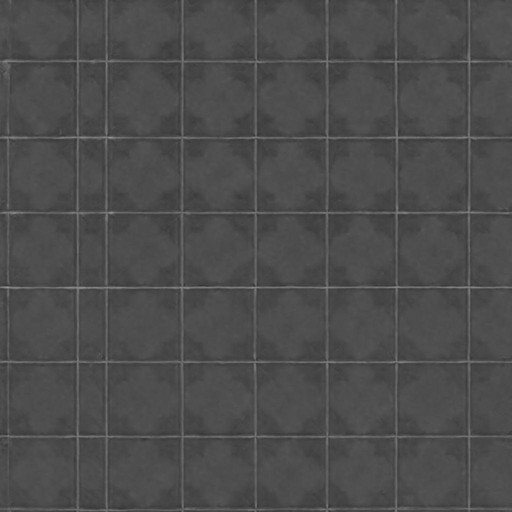} &
        \includegraphics[width=0.14\linewidth, height=0.14\linewidth]{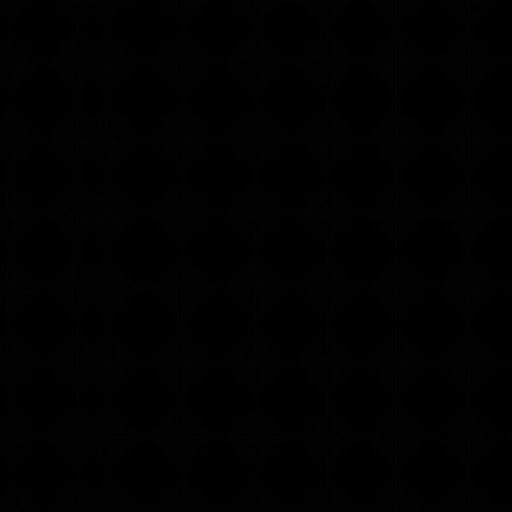} &
        \includegraphics[width=0.14\linewidth, height=0.14\linewidth]{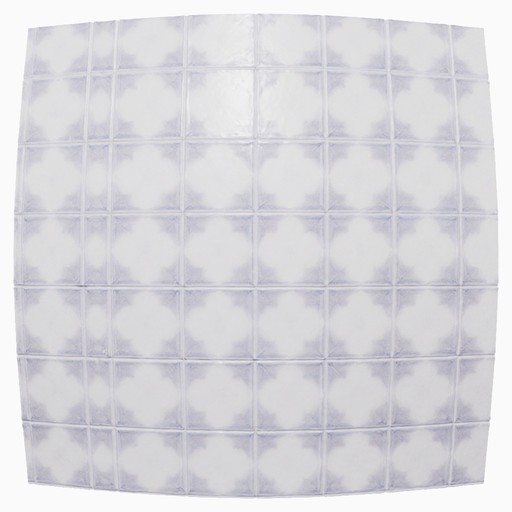} \\
    \end{tabular}
    
    \caption{\textbf{Image-prompting}. We show here a variety of materials generate using image prompts. \methodName is able to capture the visual feature of each input condition and generate a new, visually similar, material. \suppmat{Additional results are included in the Supplemental material.}}
    \label{fig:fp_generation_img}
\end{figure*}

\begin{figure*}
    \centering
    \setlength{\tabcolsep}{.5pt}
    \begin{tabular}{ccccccc}
        \small{Prompt} & \small{Basecolor} & \small{Normal} & \small{Height} & \small{Roughness} & \small{Metallic} & \small{Rendering}\\
        
        \vspace{-0.5mm}\vspace{-0.5mm}\hspace{2.5mm} \small{\makecell[b]{`Old rugged \\ concrete with \\ visible rusty \\ metal bars.'\vspace{5.5mm}}} \hspace{1mm} &
        \includegraphics[width=0.14\linewidth]{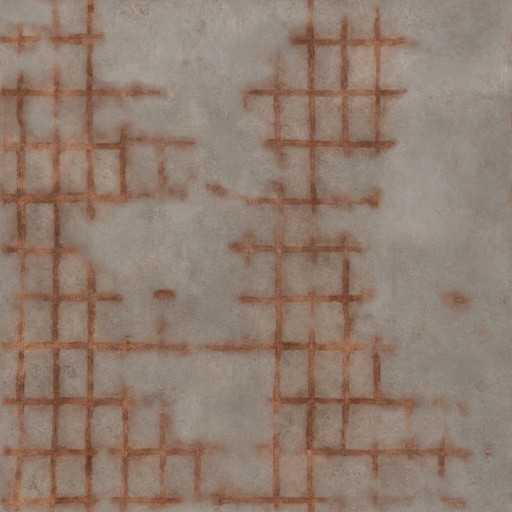} &
        \includegraphics[width=0.14\linewidth]{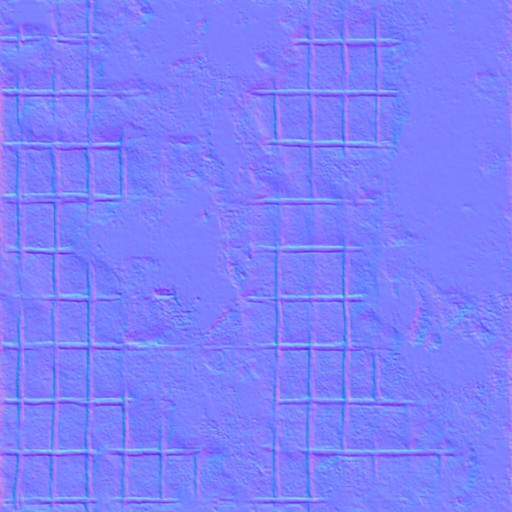} &
        \includegraphics[width=0.14\linewidth]{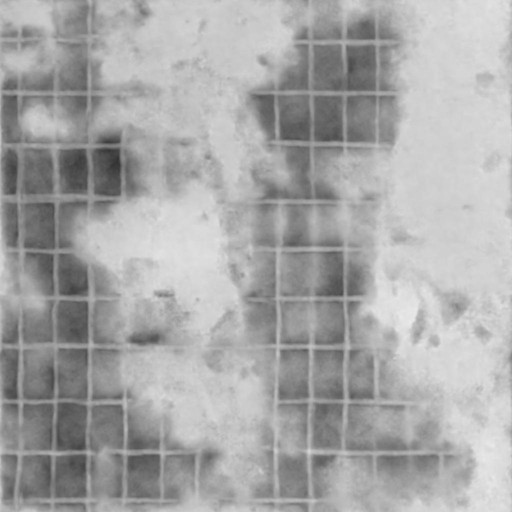} &
        \includegraphics[width=0.14\linewidth]{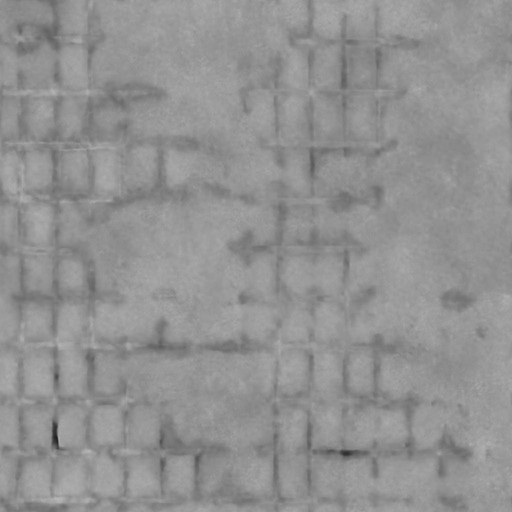} &
        \includegraphics[width=0.14\linewidth]{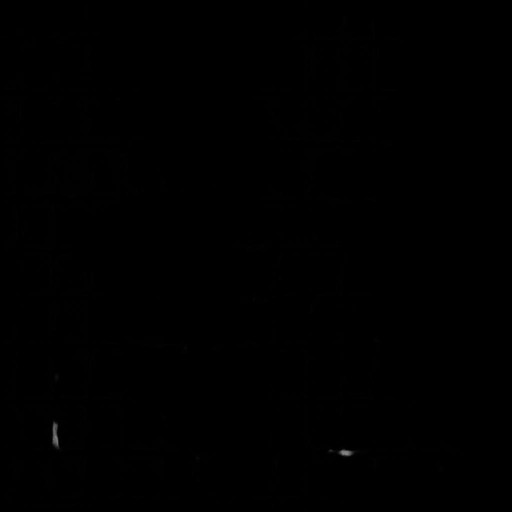} &
        \includegraphics[width=0.14\linewidth]{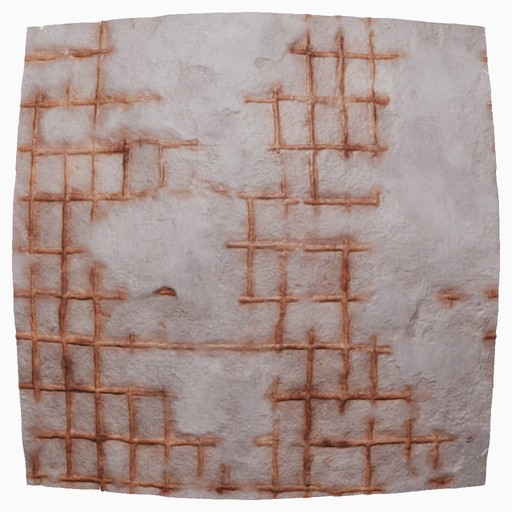} \\
    
        \vspace{-0.5mm}\vspace{-0.5mm}\hspace{2.5mm} \small{\makecell[b]{`Old wooden \\ parquet floor.'\vspace{9.5mm}}} \hspace{1mm} &
        \includegraphics[width=0.14\linewidth]{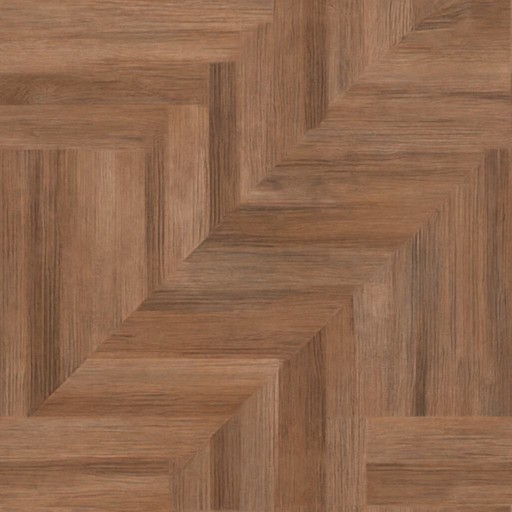} &
        \includegraphics[width=0.14\linewidth]{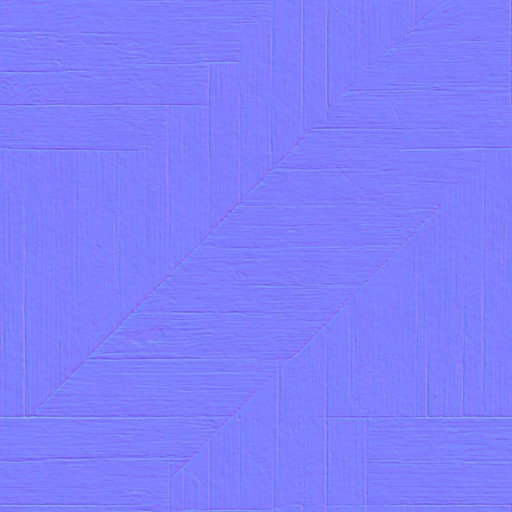} &
        \includegraphics[width=0.14\linewidth]{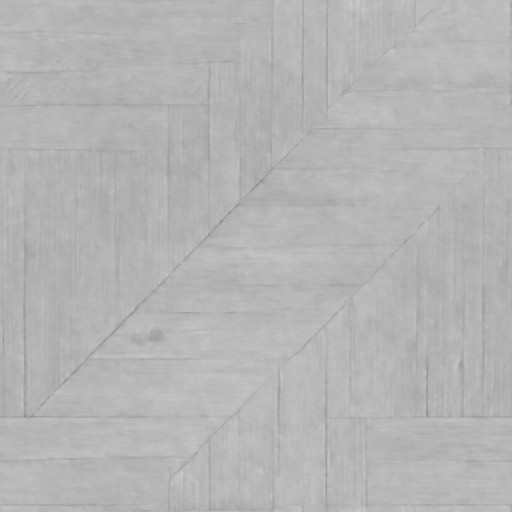} &
        \includegraphics[width=0.14\linewidth]{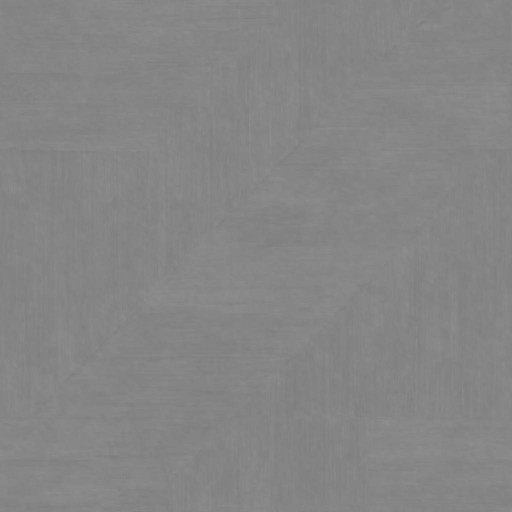} &
        \includegraphics[width=0.14\linewidth]{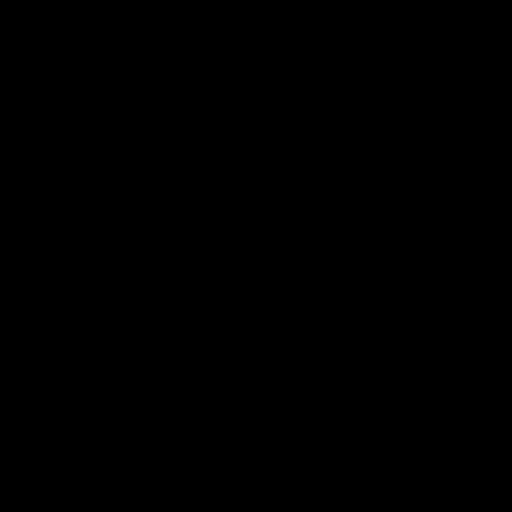} &
        \includegraphics[width=0.14\linewidth]{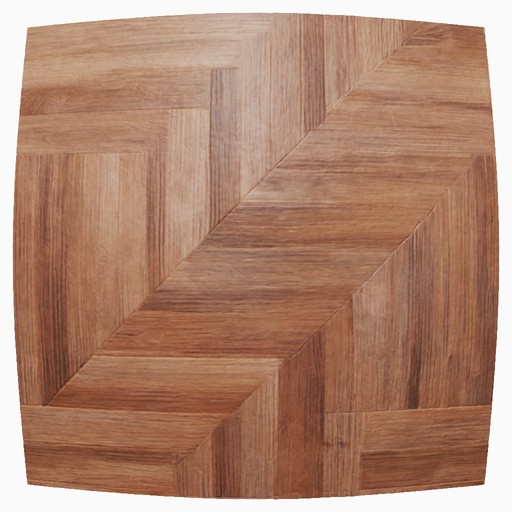} \\
    
        \vspace{-0.5mm}\vspace{-0.5mm}\hspace{2.5mm} \small{\makecell[b]{`Scottish tartan \\ wool with \\ intersecting \\ horizontal and \\ vertical  bands .'\vspace{3.5mm}}} \hspace{1mm} &
        \includegraphics[width=0.14\linewidth]{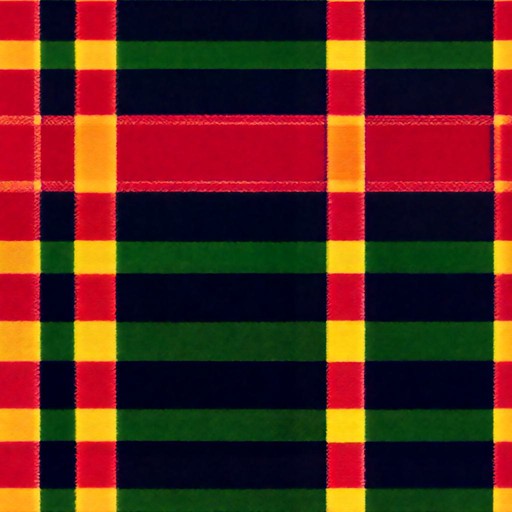} &
        \includegraphics[width=0.14\linewidth]{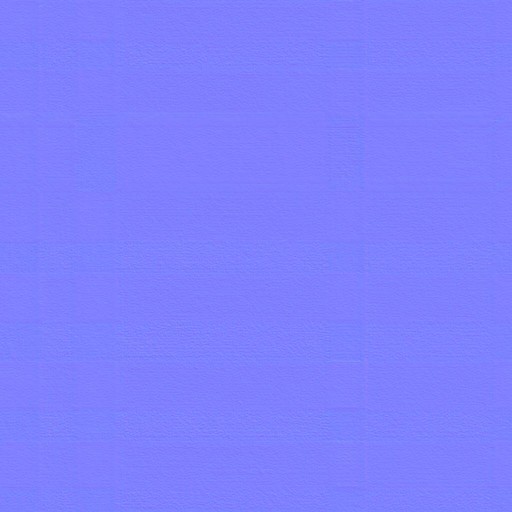} &
        \includegraphics[width=0.14\linewidth]{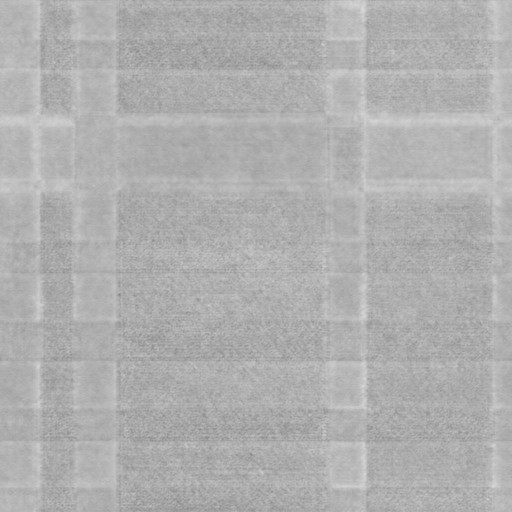} &
        \includegraphics[width=0.14\linewidth]{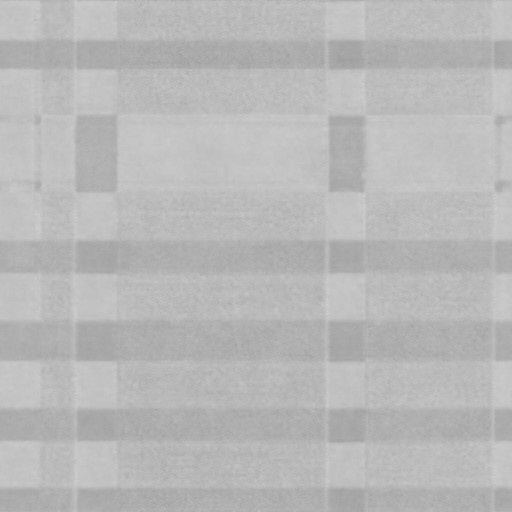} &
        \includegraphics[width=0.14\linewidth]{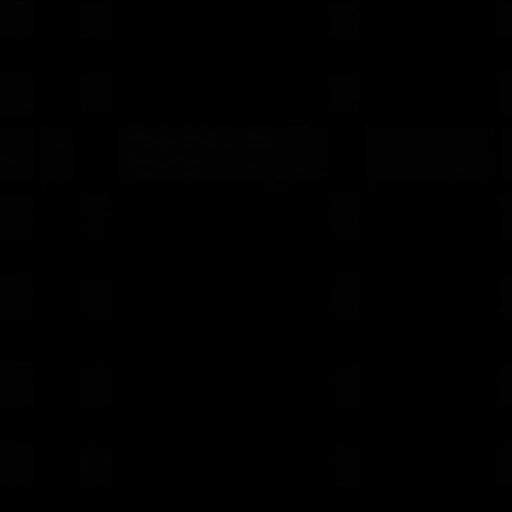} &
        \includegraphics[width=0.14\linewidth]{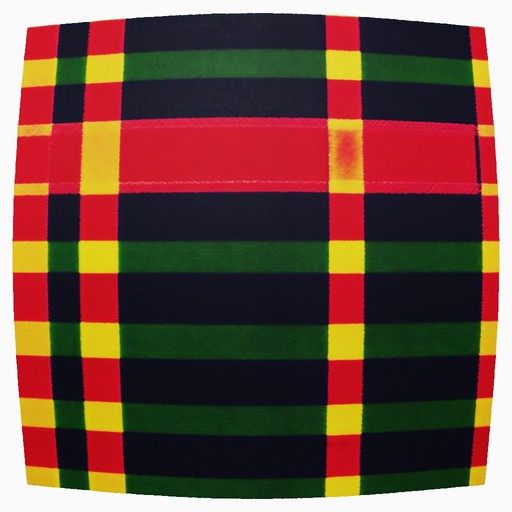} \\

        \vspace{-0.5mm}\vspace{-0.5mm}\hspace{2.5mm} \small{\makecell[b]{`Woven Bamboo \\ strips interlaced \\ into a tight \\ pattern.'\vspace{5.5mm}}} \hspace{1mm} &
        \includegraphics[width=0.14\linewidth]{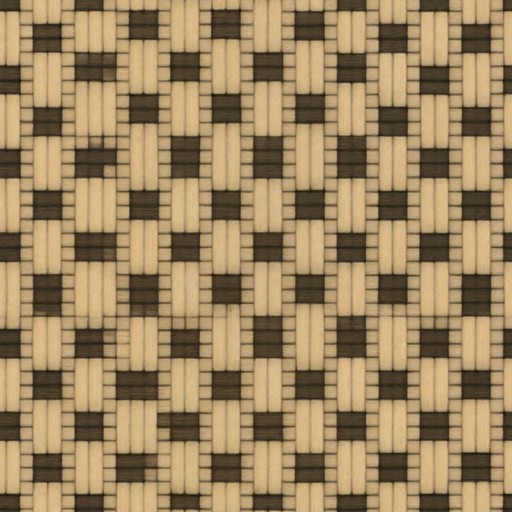} &
        \includegraphics[width=0.14\linewidth]{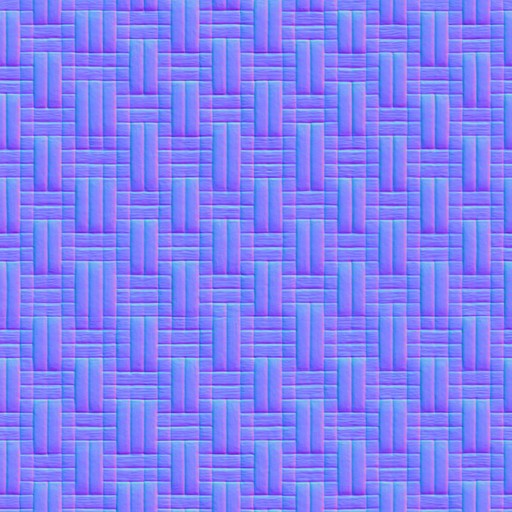} &
        \includegraphics[width=0.14\linewidth]{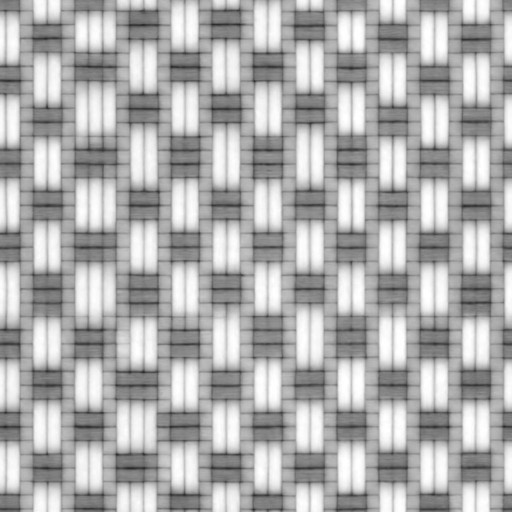} &
        \includegraphics[width=0.14\linewidth]{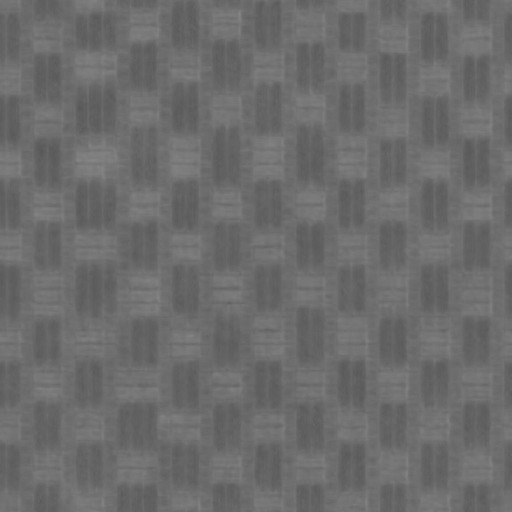} &
        \includegraphics[width=0.14\linewidth]{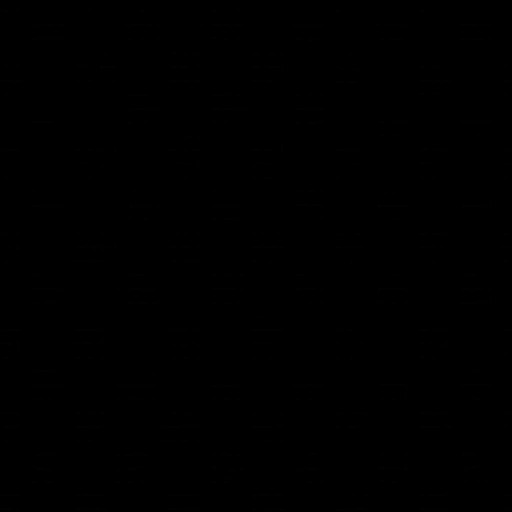} &
        \includegraphics[width=0.14\linewidth]{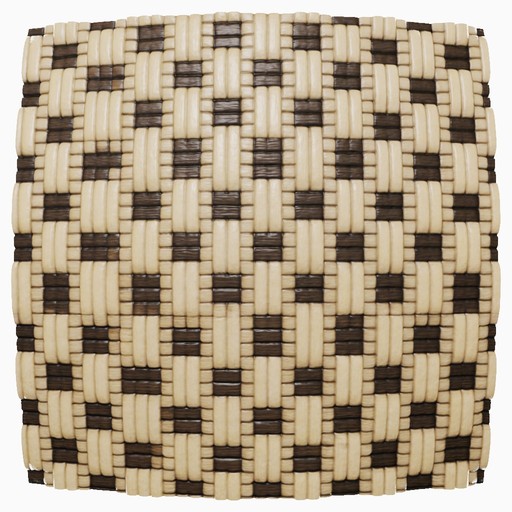} \\
        
        \vspace{-0.5mm}\vspace{-0.5mm}\hspace{2.5mm} \small{\makecell[b]{`Ancient wood \\ turned into \\ stone through \\ fossilization.'\vspace{5.5mm}}} \hspace{1mm} &
        \includegraphics[width=0.14\linewidth]{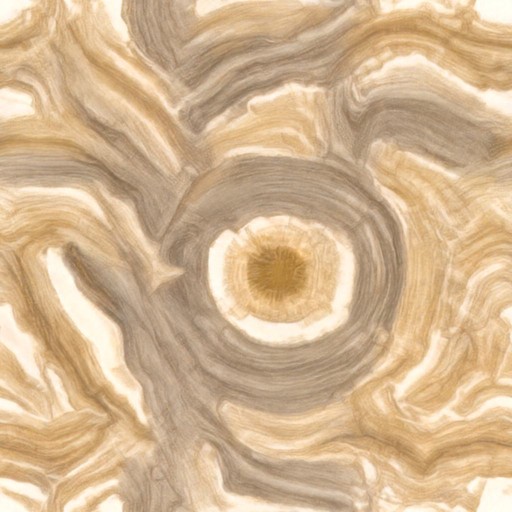} &
        \includegraphics[width=0.14\linewidth]{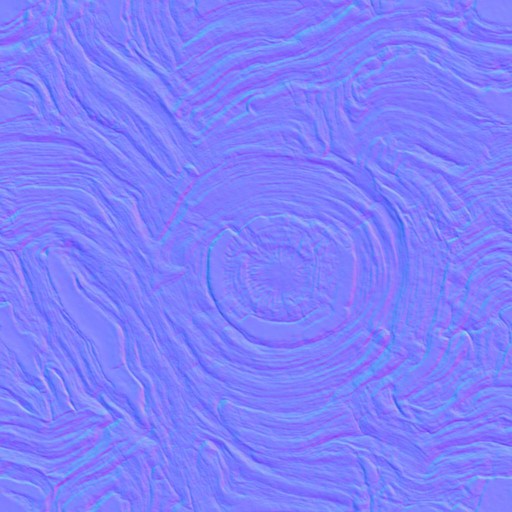} &
        \includegraphics[width=0.14\linewidth]{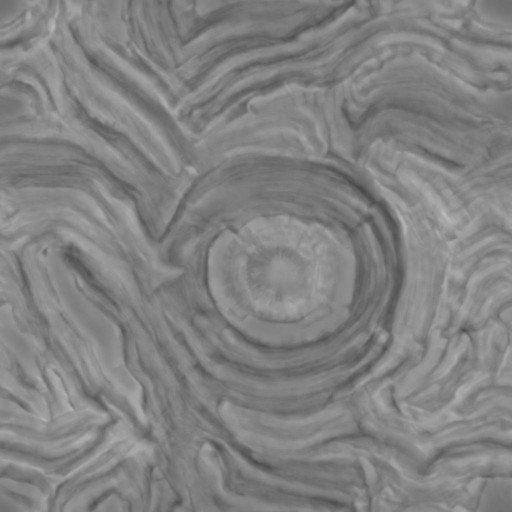} &
        \includegraphics[width=0.14\linewidth]{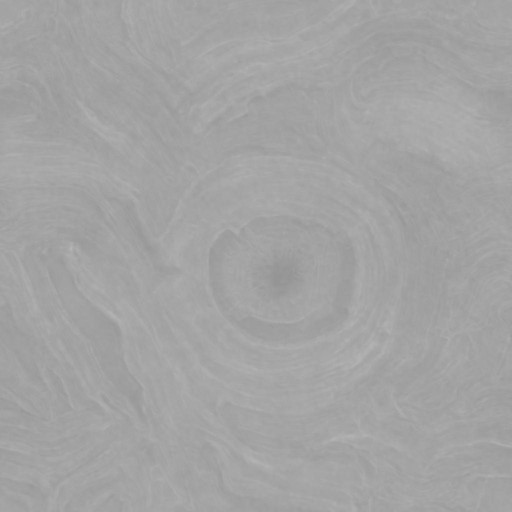} &
        \includegraphics[width=0.14\linewidth]{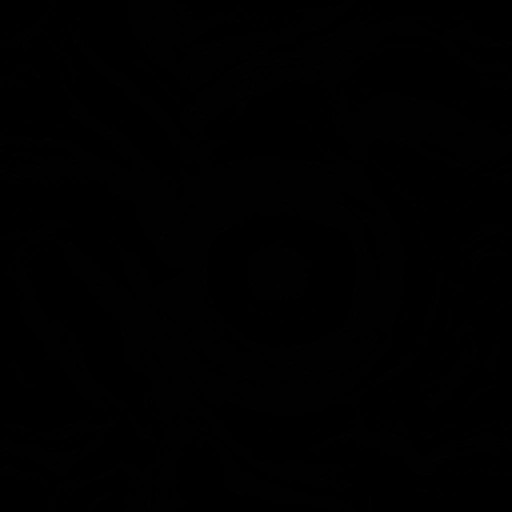} &
        \includegraphics[width=0.14\linewidth]{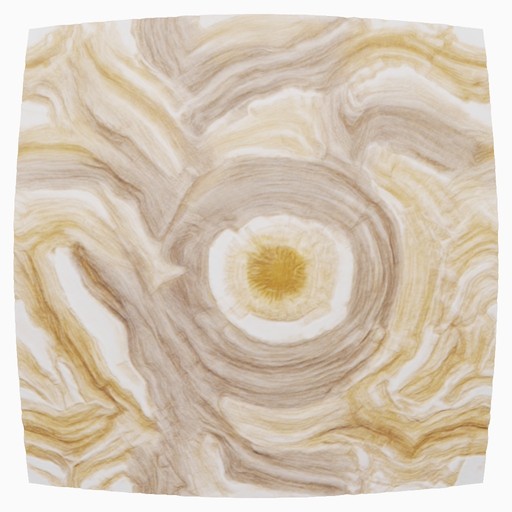} \\

        \vspace{-0.5mm}\vspace{-0.5mm}\hspace{2.5mm} \small{\makecell[b]{`Egyptian \\ papyrus with \\ hieroglyphic \\ inscriptions.'\vspace{5.5mm}}} \hspace{1mm} &
        \includegraphics[width=0.14\linewidth]{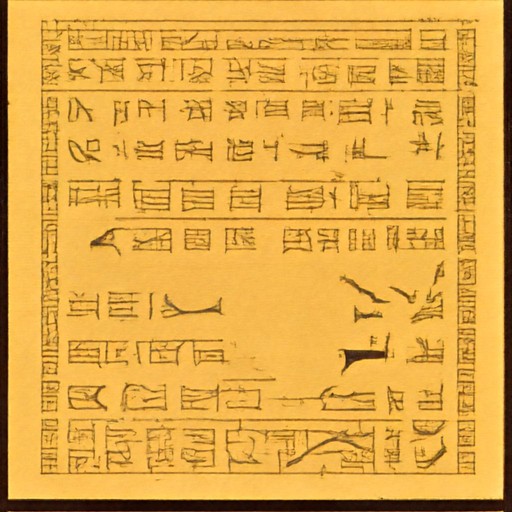} &
        \includegraphics[width=0.14\linewidth]{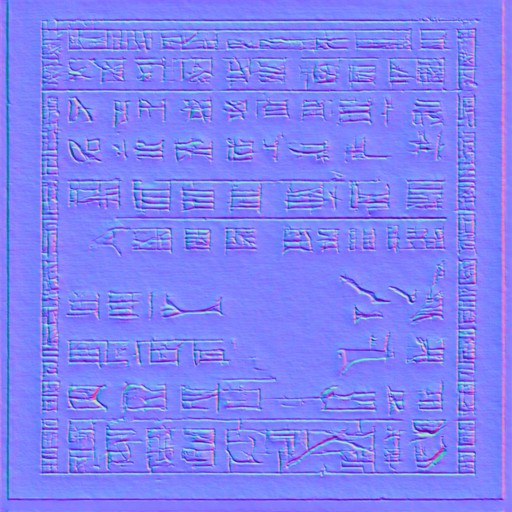} &
        \includegraphics[width=0.14\linewidth]{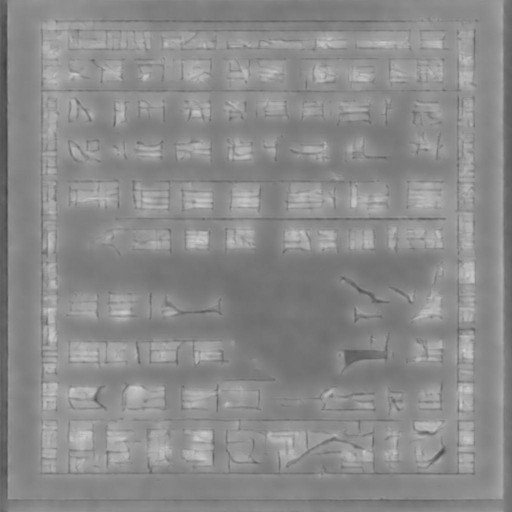} &
        \includegraphics[width=0.14\linewidth]{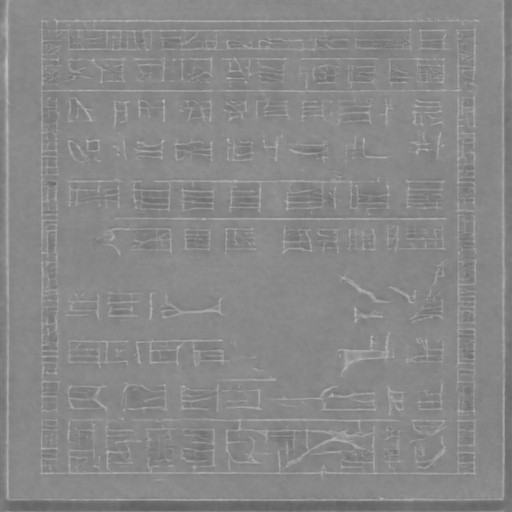} &
        \includegraphics[width=0.14\linewidth]{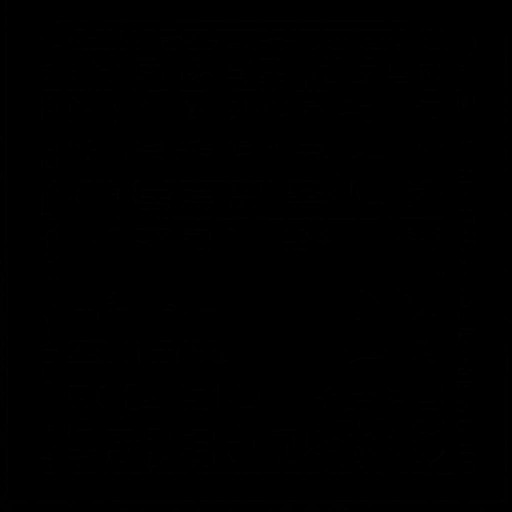} &
        \includegraphics[width=0.14\linewidth]{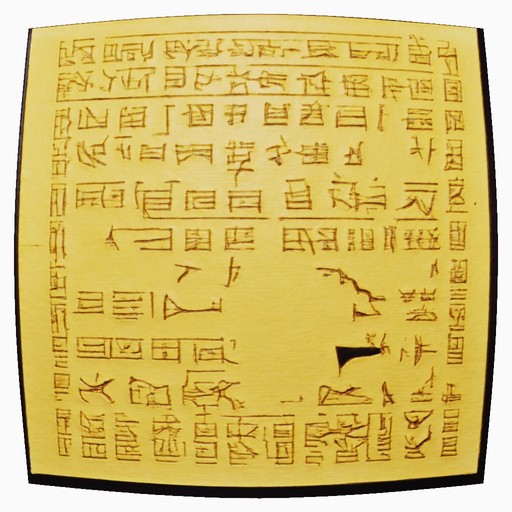} \\

        \vspace{-0.5mm}\vspace{-0.5mm}\hspace{2.5mm} \small{\makecell[b]{`Electronic \\ circuits used \\ in computers.'\vspace{7.5mm}}} \hspace{1mm} &
        \includegraphics[width=0.14\linewidth]{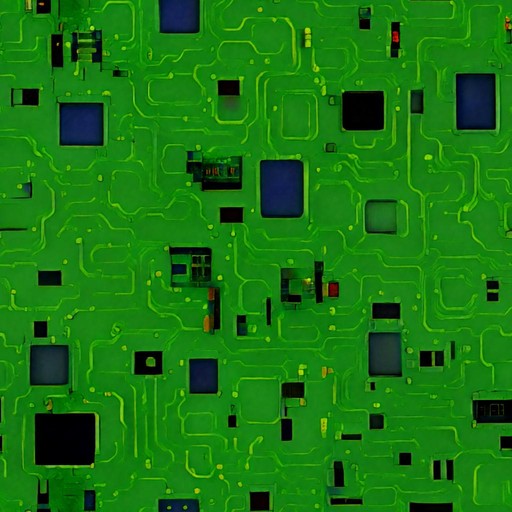} &
        \includegraphics[width=0.14\linewidth]{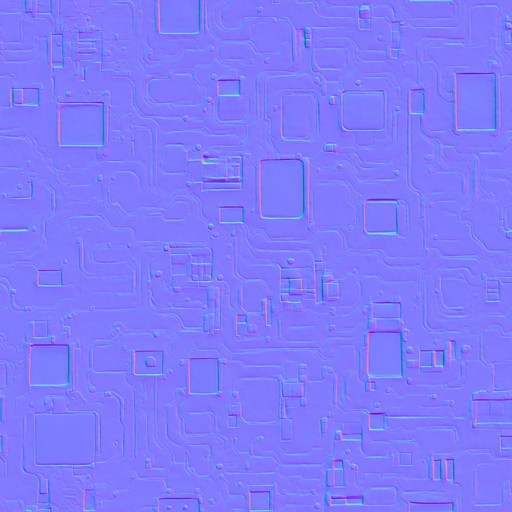} &
        \includegraphics[width=0.14\linewidth]{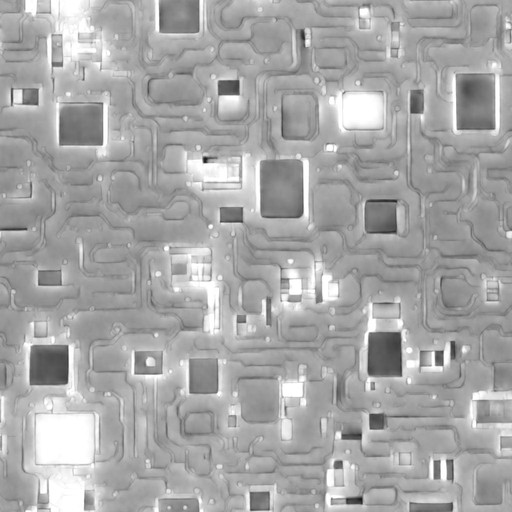} &
        \includegraphics[width=0.14\linewidth]{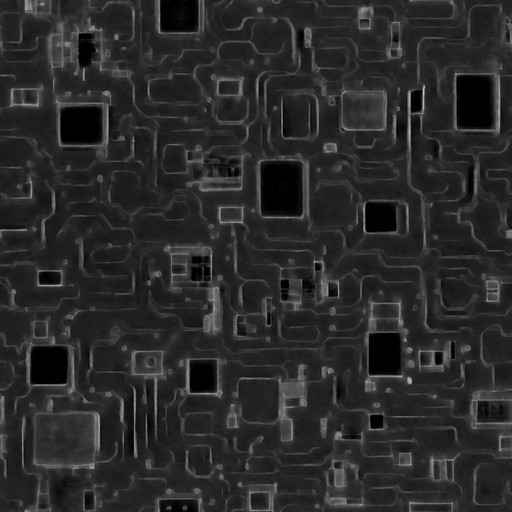} &
        \includegraphics[width=0.14\linewidth]{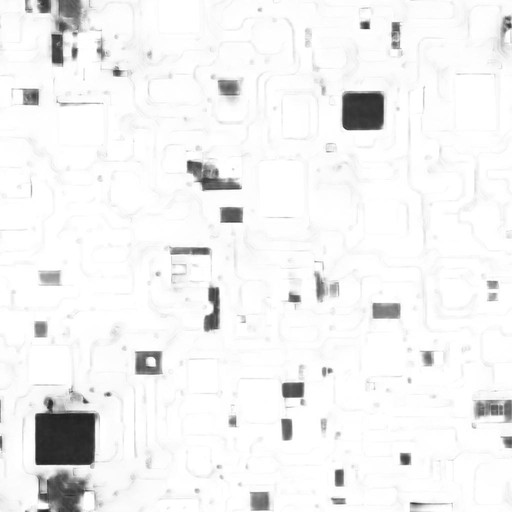} &
        \includegraphics[width=0.14\linewidth]{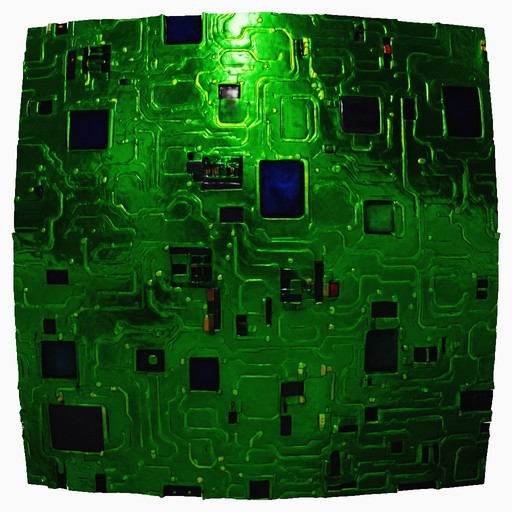} \\

        \vspace{-0.5mm}\vspace{-0.5mm}\hspace{2.5mm} \small{\makecell[b]{`Crocodile skin \\ with \\ armored scales.'\vspace{7.5mm}}} \hspace{1mm} &
        \includegraphics[width=0.14\linewidth]{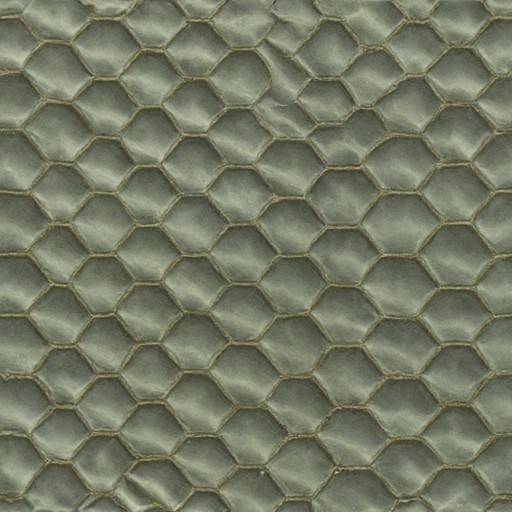} &
        \includegraphics[width=0.14\linewidth]{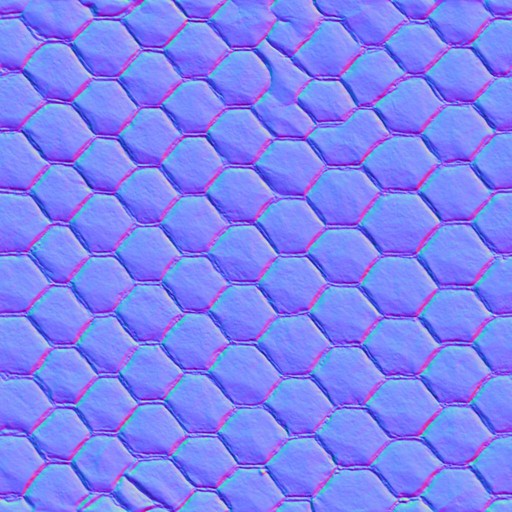} &
        \includegraphics[width=0.14\linewidth]{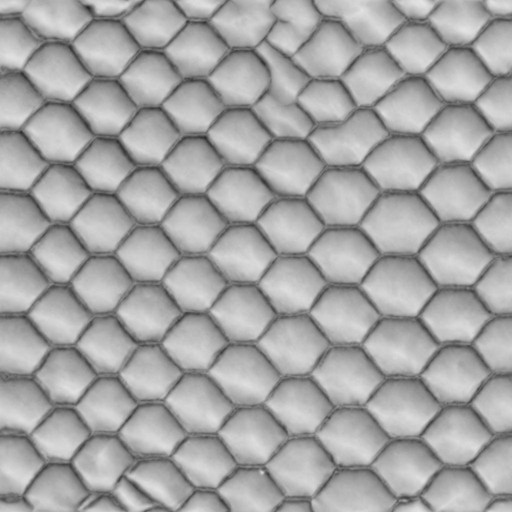} &
        \includegraphics[width=0.14\linewidth]{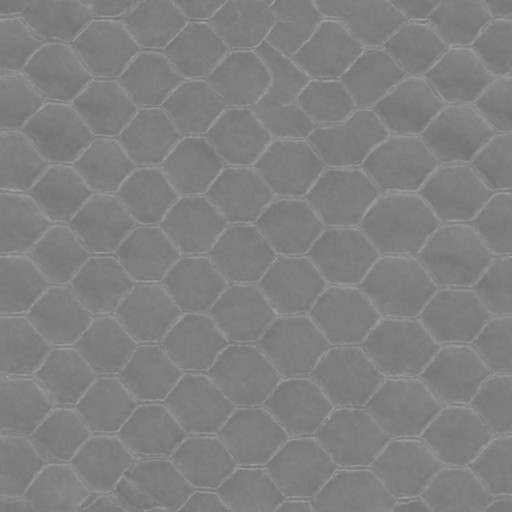} &
        \includegraphics[width=0.14\linewidth]{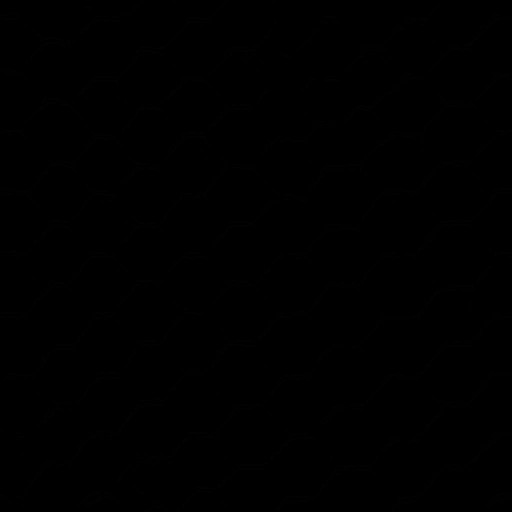} &
        \includegraphics[width=0.14\linewidth]{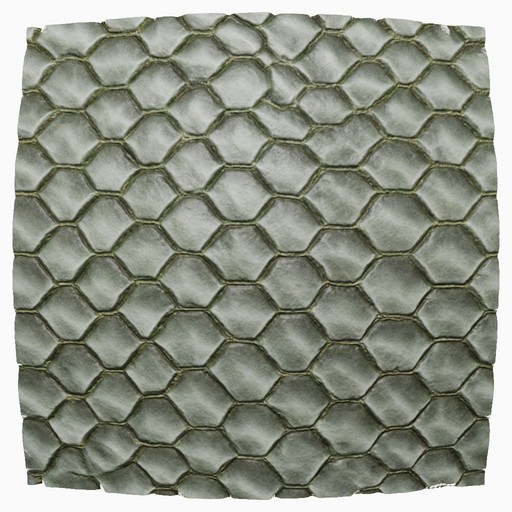} \\
    \end{tabular}
    
    \caption{\textbf{Text-prompting}. We show here a variety of materials generate using text prompts. \methodName is able to generate a new material representing the features described in the input prompt. \suppmat{Additional results are included in the Supplemental material.}}
    \label{fig:fp_generation_text}
\end{figure*}

\end{document}